\documentclass{article}

     \PassOptionsToPackage{numbers, compress}{natbib}

\usepackage[final]{neurips_2021}

\usepackage[utf8]{inputenc} 
\usepackage[T1]{fontenc}    
\usepackage{hyperref}       
\usepackage{url}            
\usepackage{booktabs}       
\usepackage{amsfonts}       
\usepackage{nicefrac}       
\usepackage{microtype}      
\usepackage{xcolor}         

\usepackage{amsthm}
\usepackage{amssymb}
\usepackage{amsmath}
\usepackage{thmtools}
\usepackage{thm-restate}
\usepackage{cleveref}
\usepackage{graphicx}
\usepackage{floatrow}
\usepackage{wrapfig}
\usepackage{comment}
\newcommand{\rulesep}{\unskip\ \vrule\ }

\usepackage{paralist}
\usepackage{enumitem}
\setlist[itemize]{leftmargin=*}
\usepackage[normalem]{ulem}
\usepackage{subcaption}
\usepackage{caption}
\usepackage[linesnumbered,ruled,vlined]{algorithm2e}

\def\R{\mathbb{R}}

\def\given{\mid}

\def\R{\mathbb{R}}

\newtheorem{theorem}{Theorem}
\newtheorem{prop}[theorem]{Proposition}

\newcommand{\abs}[1]{\left|#1\right|}
\newcommand{\size}[1]{\left|#1\right|}
\newcommand{\norm}[1]{\left\|#1\right\|}

\newcommand{\sign}[1]{\operatorname{sign}\left(#1\right)}
\newcommand{\prob}[1]{\operatorname{Pr}\left(#1\right)}

\title{Can we have it all? On the Trade-off between Spatial and Adversarial Robustness of Neural Networks}

\author{%
  Sandesh Kamath\\
  Indian Institute of Technology, Hyderabad\\
   \And
   Amit Deshpande \\
   Microsoft Research India \\
   \AND
   K V Subrahmanyam \\
   Chennai Mathematical Institue, Chennai \\
   \And
   Vineeth N. Balasubramanian \\
   Indian Institute of Technology, Hyderabad\\
}

\begin{document}

\maketitle

\begin{abstract}
(Non-)robustness of neural networks to small, adversarial pixel-wise perturbations, and as more recently shown, to even random spatial transformations (e.g., translations, rotations) entreats both theoretical and empirical understanding. Spatial robustness to random translations and rotations is commonly attained via equivariant models (e.g., StdCNNs, GCNNs) and training augmentation, whereas adversarial robustness is typically achieved by adversarial training. In this paper, we prove a quantitative trade-off between spatial and adversarial robustness in a simple statistical setting. We complement this empirically by showing that: (a) as the spatial robustness of equivariant models improves by training augmentation with progressively larger transformations, their adversarial robustness worsens progressively, and (b) as the state-of-the-art robust models are adversarially trained with progressively larger pixel-wise perturbations, their spatial robustness drops progressively. Towards achieving Pareto-optimality in this trade-off, we propose a method based on curriculum learning that trains gradually on more difficult perturbations (both spatial and adversarial) to improve spatial and adversarial robustness simultaneously.

\end{abstract}

\section{Introduction}

Neural network (NN) models have achieved state-of-the-art performance on several image tasks over the last few years. However, they have been known to be vulnerable to adversarial attacks that make small, imperceptible pixel-wise perturbations to images \cite{Szegedy13,Biggio_2013}. Most pixel-wise adversarial perturbations studied in previous work are carefully constructed, model-dependent and, in most cases also input-dependent, perturbations of small $\ell_{\infty}$ norm (and sometimes $\ell_{1}$ or $\ell_{2}$ norm). To achieve adversarial robustness, a popular, principled, and effective defense against $\ell_{\infty}$ adversarial attacks known at present is adversarial training \cite{Goodfellow15,Madry18,SinhaND2018,zhang2019theoretically}. 

From another perspective, recent work has also shown natural, physical-world attacks on neural network models using 2D and 3D spatial transformations such as translations, rotations and pose \cite{xiao2018spatially,Eykholt18robust,TramerBoneh2019,Alcorn19strike}. Even random spatial transformations such as random translations and random rotations degrade the accuracy of neural network models considerably \cite{Engstrom2019Exploring}. Recent benchmarks also measure the robustness of NN models to the average-case, model-agnostic perturbations (e.g., noise, blur, brightness, contrast) or natural distribution shifts, and not only to the worst-case adversarial perturbations \cite{hendrycks2018benchmarking,taori2020measuring}.
Spatial robustness or robustness to spatial transformations (e.g., translations, rotations) has been long studied as a problem of model invariance, and is generally addressed using equivariant models or training augmentation. While standard Convolutional Neural Networks (StdCNNs) are translation-equivariant, recent efforts have resulted in equivariant NN models for other transformations such as rotation, flip \cite{Gens2014deep,Cohen16,Dieleman16,Worrall16,Marcos17,Laptev16,Li17,Esteves18}, and scaling \cite{Marcos2018scale,Worrall2019deepscale,Sosnovik2020scale}. Such group-equivariant NN models (GCNNs) are designed to be invariant to a specific group of transformations, so they need training augmentation to attain spatial robustness to random or average-case transformations. However, aided by clever weight-sharing, these recent works have achieved good spatial robustness performance with lesser training augmentation than their non-equivariant counterparts. 

In this paper, we address an inevitable trade-off between the average-case spatial robustness and the worst-case $\ell_{\infty}$ adversarial robustness in neural network models, and show results theoretically as well as empirically. On the one hand, as we attain better spatial robustness via equivariant models and larger training augmentation, the adversarial robustness worsens. On the other hand, as we attain better adversarial robustness via adversarial training against larger pixel-wise perturbations, the spatial robustness worsens. We then provide a curriculum learning-based approach to mitigate this issue and achieve a certain degree of Pareto-optimality in this trade-off.

Our work extends recent progress on two important questions, namely, the robustness to multiple attacks and the robustness vs accuracy trade-off. The robustness to translations (up to $\pm 3$ pixels) and rotations (up to $\pm 30^{\circ}$) has been shown to have a fundamentally different loss landscape than the robustness to adversarial $\ell_{p}$ perturbations \cite{Engstrom2019Exploring}. A trade-off between robustness to adversarial $\ell_{\infty}$ perturbations and adversarial translations and rotations up to $\pm 3$ pixels and $\pm 40^{\circ}$, respectively, was also recently noted \cite{TramerBoneh2019}. Adversarial attacks on $\ell_{p}$ adversarially robust models by exploiting invariance at intermediate layers have also been of interest in recent work \cite{jacobsen2018excessive,tramer2020fundamental}. A key difference of our work with these efforts is that: (i) we study spatial robustness to \emph{random} translations and rotations (which is more closely related to practice when training NN models and has not been studied in these earlier efforts), instead of adversarial translations or rotations; (ii) we study spatial robustness w.r.t transformations over their entire range, and do not restrict ourselves to a $\pm 30-40^{\circ}$ range, for instance; and (iii) we consider the use of group-equivariant models with training augmentation (which are designed for spatial robustness) in our studies, beyond just StdCNN models. 
Our results show a \emph{progressive decline} in spatial robustness as adversarial robustness improves, and vice versa -- a trend that was not explicitly noted in previous results. 
Our result can also be interpreted as a progressive trade-off between adversarial robustness and generalization accuracy itself \cite{Tsipras2019odds,zhang2019theoretically,YangRZSC20}, where generalization performance measures accuracy on a test distribution that includes natural data augmentation such as random translations, rotations, etc.

Given the nature of this trade-off, we then ask the question of how one can obtain a model close to the Pareto-optimal frontier for adversarial and random spatial robustness. Are there training strategies that can help achieve a balance between both? We observe that the obvious solutions of training in two stages -- first, adversarial training and then training augmentation by spatial transformations, or vice versa -- does not help address this issue. This results in a version of catastrophic forgetting, where as the model learns robustness on one of these fronts, it loses robustness on the other. We instead show that a curriculum learning-based strategy, where the model is trained by applying an adversarial perturbation to a randomly transformed input with gradually increased difficulty levels, gives solutions consistently closer to the Pareto-optimal frontier across well-known datasets and models. This strategy reflects the progressive nature noted of the trade-off itself, and highlights the importance of the observation of this trend.

\noindent Our key contributions can be summarized as follows:
\vspace{-10pt}
\begin{itemize}
\setlength\itemsep{-0.3em}
\item We study a trade-off between adversarial robustness and spatial robustness (with a focus on \emph{random} spatial transformations) both theoretically and empirically, on a setting closer to natural neural network training, which to the best of our knowledge, has not been considered before. We explain the trade-off between spatial and adversarial robustness with a well-motivated theoretical construction.
\item We conduct a comprehensive suite of experiments across popularly used models and datasets, and find that models trained for spatial robustness (StdCNNs and GCNNs) progressively improve spatial robustness when their training is augmented with larger random spatial transformations but while doing so, their $\ell_{\infty}$ adversarial robustness drops progressively. 
Similarly, models adversarially trained using progressively larger $\ell_{\infty}$-norm attacks improve their adversarial robustness but while doing so, their spatial robustness drops progressively.
\item We provide a simple yet effective approach based on curriculum learning that gives better-balanced, Pareto-optimal spatial and adversarial robustness compared to several natural baselines derived from combining adversarial training and training augmentation by random spatial transformations. 
\end{itemize}

\section{Problem Setting and Formulation}
\vspace{-2pt}
We begin by describing our problem formally. Let $(X,Y)$ denote a random input-label pair from a given data distribution over the input-label pairs $(x, y) \in \mathcal{X} \times \mathcal{Y}$. Let $f$ be a neural network classifier. The accuracy of $f$ is given by $\prob{f(X) = Y}$, the fraction of inputs where the prediction matches the true label. For $\epsilon > 0$, an $\ell_{\infty}$ adversarial attack $\mathcal{A}$ maps each input $x$ to $x + \mathcal{A}(x)$ such that $\norm{\mathcal{A}(x)}_{\infty} \leq \epsilon$, for all $x \in \mathcal{X}$. The \emph{fooling rate} of $\mathcal{A}$ is given by $\prob{f(X + \mathcal{A}(X)) \neq f(X)}$, the fraction of inputs where the prediction changes after adversarial perturbation. For a given spatial transformation $T$, the classifier $f$ is said to be $T$-invariant if the predicted label remains unchanged after the transformation $T$ for all inputs; in other words, $f(Tx) = f(x)$, for all $x \in \mathcal{X}$. The spatial transformations such as translation, rotation and flip preserve the $\ell_{\infty}$ norm, so $T\mathcal{A}(x)$ is an adversarial perturbation of small $\ell_{\infty}$ norm for the input $Tx$ for a $T$-invariant classifier. 
Therefore, for a given perturbation bound $\epsilon > 0$, the maximum fooling rate for any $T$-invariant classifier $f$ on the transformed data $\{Tx \;:\; x \in \mathcal{X}\}$ must be equal to the maximum fooling rate for $f$ on the original data $\mathcal{X}$. It is a bit more subtle when $f$ is not truly invariant, that is, $f(Tx) = f(x)$ for most inputs $x$ but not all $x$. We define the rate of invariance of a classifier $f$ to a transformation $T$ as $\prob{f(TX) = f(X)}$ or the fraction of test images whose predicted labels remain unchanged under the transformation $T$. For a class of transformations, e.g., random rotation $r$ from the range $[-\theta^{\circ}, +\theta^{\circ}]$, we define the \emph{rate of invariance} as the average rate of invariance over transformations $T$ in this class, i.e., $\prob{f(r(X)) = f(X)}$, where the probability is over the random input $X$ as well as the random transformation $r$. The rate of invariance is $100 \%$ if the model $f$ is truly invariant. When $f$ is not truly invariant, the interplay between the invariance under transformations and robustness under adversarial perturbations of small $\ell_{\infty}$-norm is subtle. \emph{This interplay is exactly what we investigate.}

In this paper, we study neural network models and the simultaneous interplay between their spatial robustness, viz, rate of invariance for spatial transformations (such as random rotations between $[- \theta^{\circ}, + \theta^{\circ}]$ for $\theta$ varying in the range $[0, 180]$), and their adversarial robustness to pixel-wise perturbations of $\ell_{\infty}$ norm at most $\epsilon$. 
Measuring the robustness of a model to adversarial perturbations of $\ell_{p}$ norm at most $\epsilon$ is NP-hard \cite{KatzBDJK2017,SinhaND2018}. Previous work on the comparison of different adversarial attacks has shown the Projected Gradient Descent (PGD) attack to be among the strongest \cite{Athalye2018obfuscated}. Hence, we use PGD-based adversarial training as the methodology for adversarial robustness in a model, as commonly done. We show that our conclusions also hold for other forms of adversarial training such as TRADES \cite{zhang2019theoretically}, which gives a trade-off between adversarial robustness and natural accuracy (on the original unperturbed data).

Theoretically speaking, we prove a quantitative trade-off between spatial and adversarial robustness in a simple, natural statistical setting proposed in previous work \cite{Tsipras2019odds}. Our theoretical results (described in Sec \ref{sec_theory}) show that there is a trade-off between spatial and adversarial robustness, and that it is not always possible to have both in a given model. 
Empirically, we study the following: \begin{inparaenum}[(a)] \item change in $\ell_{\infty}$ adversarial robustness as we improve only the rate of spatial robustness using training augmentation with progressively larger transformations; \item change in spatial invariance as we improve only adversarial robustness using PGD adversarial training with progressively larger $\ell_{\infty}$-norm of pixel-wise perturbations. \end{inparaenum} 
We study StdCNNs, group-equivariant GCNNs, as well as popular architectures used to assess adversarial robustness \cite{Madry18,zhang2019theoretically}. Our empirical studies are conducted on MNIST, CIFAR10, CIFAR100 as well as Tiny ImageNet datasets, thus showing the general nature of these results (described in Sec \ref{sec_expts}). Importantly, as stated earlier, we consider random spatial transformations that are more commonplace than adversarial rotations in previous work \cite{TramerBoneh2019}. We look at the entire possible range for spatial transformations. Similarly, we normalize the underlying dataset, and compute the accuracy of a given model on test inputs adversarially perturbed using PGD attack of $\ell_{\infty}$ norm at most $\epsilon$, for $\epsilon$ varying over $[0, 1]$. In contrast with the certifiable lower bounds on robustness \cite{balovic2019geo,fischer2020cert,ruoss2021efficient,li2020provable,mohapatra2020verify}, we study upper bounds on spatial and adversarial robustness, respectively.

\vspace{-4pt}
\section{Spatial-Adversarial Robustness Trade-off} \label{sec_theory}
\vspace{-2pt}
In this section, we prove the trade-off between spatial and adversarial robustness theoretically, and support this result with experiments in Sec \ref{sec:exp-results}. We use $\mathcal{A}(x)$ to denote an adversarial $\ell_{\infty}$ perturbation and $r(x)$ to denote a random spatial transformation. Equivariant model constructions often consider a group of transformations and construct a neural network model invariant to this group. For simplicity, we consider a cyclic group that can model a rotation group (e.g., integer multiples of $30^{\circ}$), or horizontal/vertical translations (e.g., horizontal translations by multiples of, say, $\pm 4$ pixels). 

We take a construction proposed in previous work for a robustness vs accuracy trade-off \cite{Tsipras2019odds}, and infuse it with a simple idea from the theory of error correcting codes \cite{roth2006intro}. Consider a binary cyclic code of length $d$, where each codeword is a binary string in $\{-1, 1\}^{d}$ and all the codewords can be obtained by applying successive cyclic shift of coordinates to a generator codeword; see Chapter 8 in \cite{roth2006intro}. The cyclic code is said to have relative distance $\delta$, if any two codewords differ in at least $\delta d$ coordinates. Let $c = (c_{1}, c_{2}, \dotsc, c_{d})$ be the generator codeword. Consider a random input-label pair $(X, Y)$, with $X = (X_{0}, X_{1}, \dotsc, X_{d})$ taking values in $\R^{d+1}$ and $Y$ taking values in $\{-1, 1\}$, generated as follows. The class label $Y$ takes value $\pm 1$ with probability $1/2$ each. $X_{0} \given Y=y$ takes value $y$ with probability $p$ and $-y$ with probability $1-p$, for some $p \geq 1/2$. The remaining coordinates are independent and normally distributed with $X_{t} \given Y=y$ as $N(2c_{t}y/\sqrt{d}, 1)$, for $1 \leq t \leq d$. Let $m$ be an integer divisor of $d$, and let $r_{j}(x)$ denote the cyclic shift of $mj$ coordinates applied to $(x_{1}, x_{2}, \dotsc, x_{d})$ while keeping $x_{0}$ unchanged, i.e., $r_{j}(x) = (x_{0}, x_{mj+1}, x_{mj+2}, \dotsc, x_{d}, x_{1}, \dotsc, x_{mj})$. For example, a $90^{\circ}$ rotation can be considered as a cyclic permutation of pixels with $m = d/4$ and the center pixel $x_{0}$ being fixed. Now let $r(x)$ denote a random permutation that takes value $r_{j}(x)$, uniformly at random over $j \in \{1, 2, \dotsc, d/m\}$. For example, using $m=d/4$ as above, $r(x)$ can model a random multiple of $90^{\circ}$ rotation.

First, we show that achieving a high degree of spatial robustness on the above distribution is non-trivial even if we have high accuracy on the original data without any transformations. So our subsequent trade-off between spatial and adversarial robustness cannot be derived from previously known trade-offs between accuracy and adversarial robustness \cite{Tsipras2019odds,zhang2019theoretically}. 
\begin{prop} 
\label{prop:accuracy}
There exists $p \geq 1/2$ and a cyclic code with relative distance $\delta \geq 3/8$ such that given the input distribution defined as above, the classifier of maximum accuracy on input $(X, Y)$ has accuracy at least $97\%$. Similarly, the classifier of maximum accuracy on the transformed input $(r_{j}(X), Y)$ also has accuracy at least $97\%$. However, when the classifier of maximum accuracy on $(X, Y)$ is applied to $(r_{j}(X), Y)$, for any $j$, it has accuracy at most $85\%$.
\end{prop}
Next we show that if we have a classifier of high adversarial robustness on the above data distribution, then it must have low spatial robustness and vice versa. The proof of this follows from a previously known accuracy vs robustness trade-off by Tsipras et al.\cite{Tsipras2019odds}. However, their distribution is invariant to any permutation of the coordinates $x_{1}, x_{2}, \dotsc, x_{d}$, so the accuracy and the spatial robustness are equal for their distribution.
\begin{theorem} \label{theorem:trade-off}
Given the input distribution defined as above, any $\eta > 0$ and any classifier $f: \mathbb{R}^{d+1} \rightarrow \{-1, 1\}$, if the adversarial accuracy of $f$ is at least $1-\eta$, then the spatial accuracy of $f$ is at most $\dfrac{\eta~ p}{(1-p)}$. Similarly, if the spatial accuracy of $f$ is at least $1-\eta$ then the adversarial accuracy $f$ is at most $1 - \dfrac{(1-p)(1-\eta)}{p}$.

\end{theorem}
The above results show that adversarial and spatial robustness can not just be high simultaneously.
We do not explicitly model the effect of adversarial training with larger $\ell_{\infty}$ perturbations but, as a reasonable proxy, consider the increase in the adversarial accuracy for a fixed perturbation bound. 
A tighter analysis of Theorem \ref{theorem:trade-off} towards a gradual trade-off between spatial and adversarial robustness is presented in the supplementary.
%
\vspace{-2pt}
\section{Empirical Analysis} 
\label{sec:exp-results}
\label{sec_expts}
\vspace{-2pt}
\noindent \textbf{Experimental Setup.} In order to study the effect of spatial transformations (e.g. rotations, translations) and adversarial perturbations (e.g. Projected Gradient Descent or PGD attacks), we use well-known popularly used NN architectures as well as state-of-the-art networks known to have strong invariance \cite{Cohen16}, as well as networks which are known to have strong robustness \cite{Madry18,zhang2019theoretically}. We study the spatial vs adversarial robustness trade-off using well-known architectures (described below) on MNIST, CIFAR10, CIFAR100 and Tiny ImageNet datasets. Our code is made available for reproducibility.

\uline{\textit{Spatially Robust Model Architectures:}} 
StdCNNs are known to be translation-equivariant by design, and GCNNs \cite{Cohen16} are rotation-equivariant by design through clever weight sharing \cite{Kondor18}. Equivariant models, especially GCNNs, when trained with random rotation augmentations have been observed to come very close to being truly rotation-invariant \cite{Cohen16,Cohen17,Cohen18} (or spatially robust in our context). We hence use both StdCNNs and equivalent GCNNs trained with suitable data augmentations for our studies with spatially robust architectures. In particular, 
for each StdCNN we use, the corresponding GCNN architecture is obtained by replacing the layer operations with equivalent GCNN operations as in \cite{Cohen16}\footnote{https://github.com/adambielski/pytorch-gconv-experiments}. For the StdCNN, we use a Conv-Conv-Pool-Conv-Conv-Pool-FC-FC architecture for MNIST (more details in the supplementary); VGG16 \cite{Simonyan14} and ResNet18 \cite{He16} for CIFAR10 and CIFAR100; and ResNet18 for the Tiny ImageNet dataset.

\uline{\textit{Adversarially Robust Model Architectures:}} 
For adversarial training, we use a LeNet-based architecture for MNIST\footnote{https://github.com/MadryLab/mnist\_challenge/} and a ResNet-based architecture for CIFAR10\footnote{https://github.com/MadryLab/cifar10\_challenge}. Both these models are exactly as given in \cite{Madry18}. For CIFAR100, we use the popularly used WideResNet-34\footnote{https://github.com/yaodongyu/TRADES/models/} architecture also used in \cite{zhang2019theoretically}. We use ResNet18 \cite{He16} for the Tiny ImageNet dataset.





\begin{figure*}[!h]
\begin{center}
\includegraphics[width=0.24\linewidth]{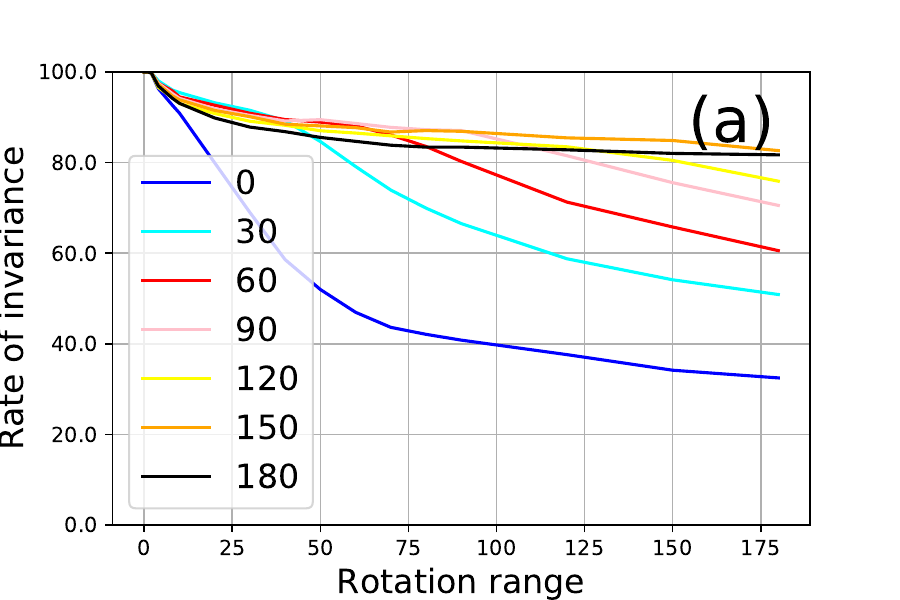} 
\includegraphics[width=0.24\linewidth]{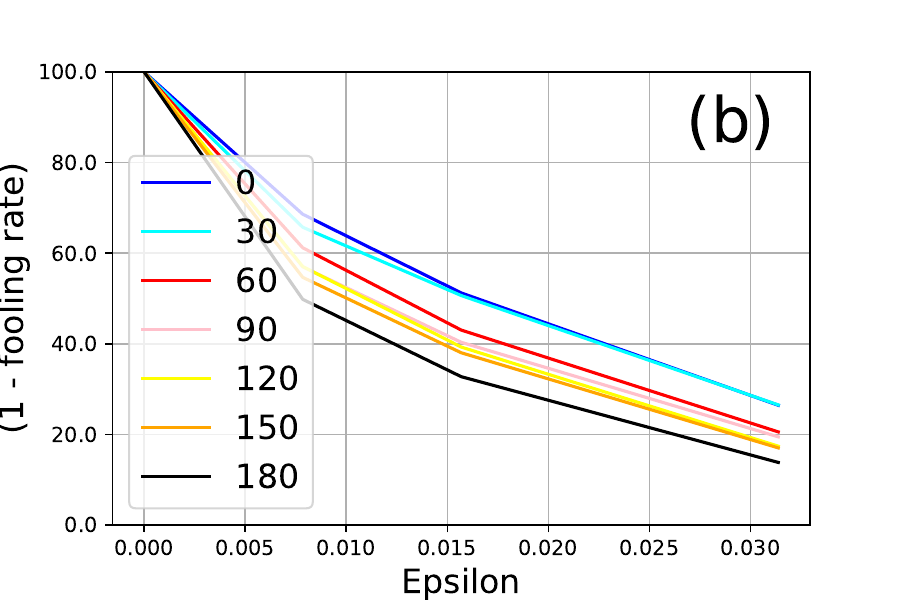}
\rulesep
\includegraphics[width=0.24\linewidth]{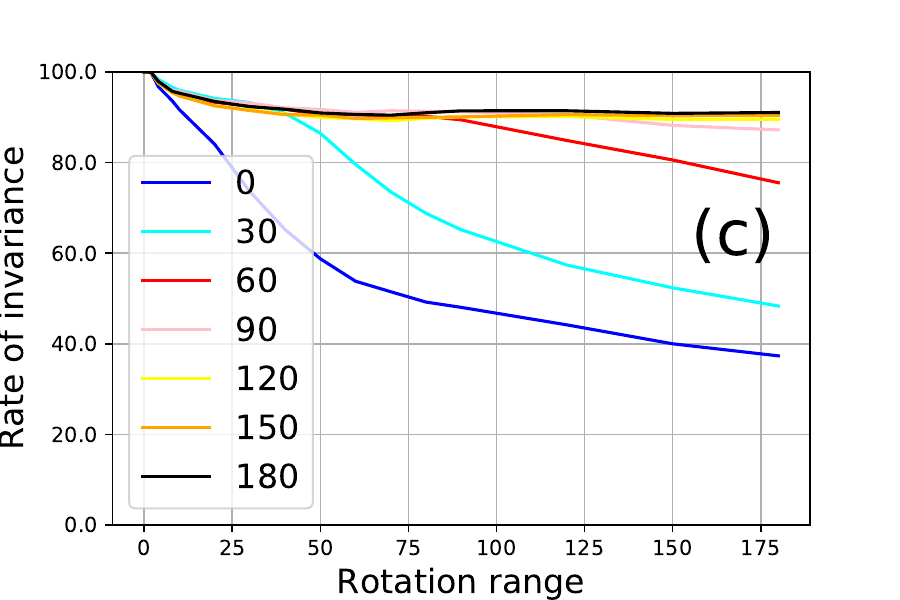} 
\includegraphics[width=0.24\linewidth]{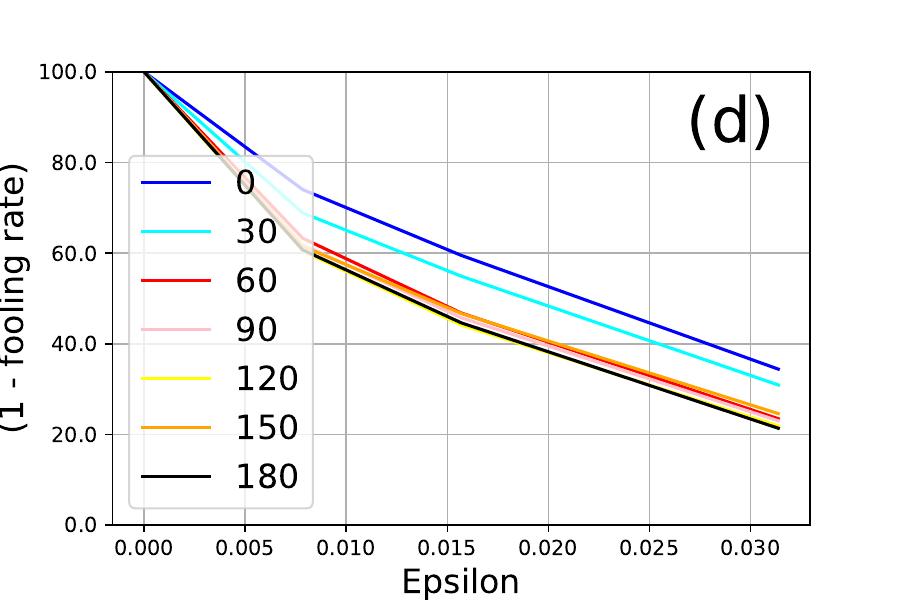}
\includegraphics[width=0.24\linewidth]{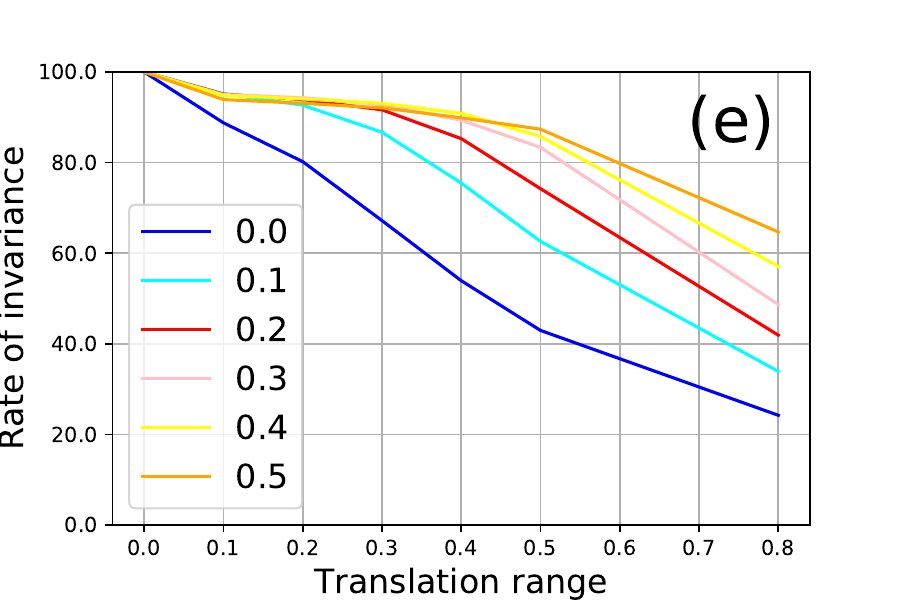}
\includegraphics[width=0.24\linewidth]{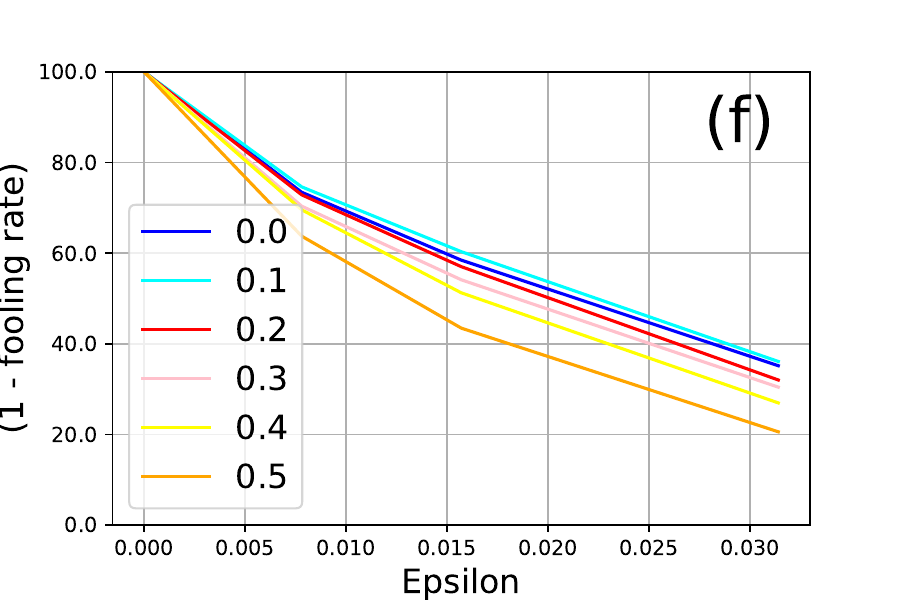}
\rulesep
\includegraphics[width=0.24\linewidth]{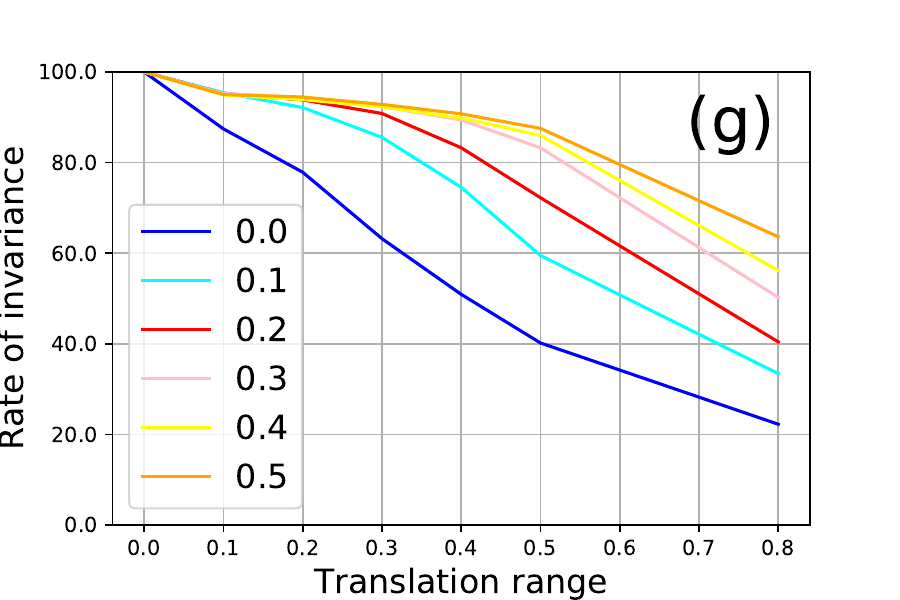}
\includegraphics[width=0.24\linewidth]{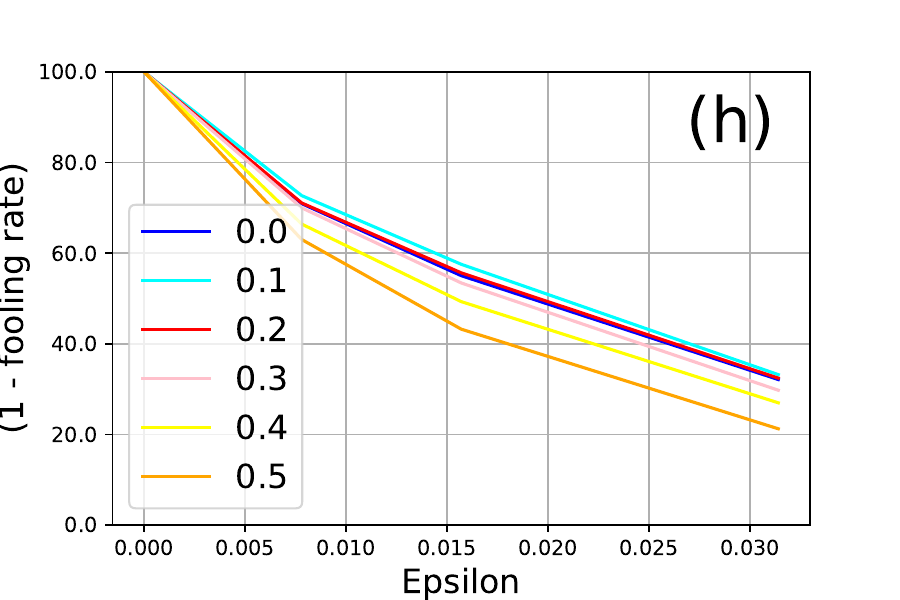}
\includegraphics[width=0.24\linewidth]{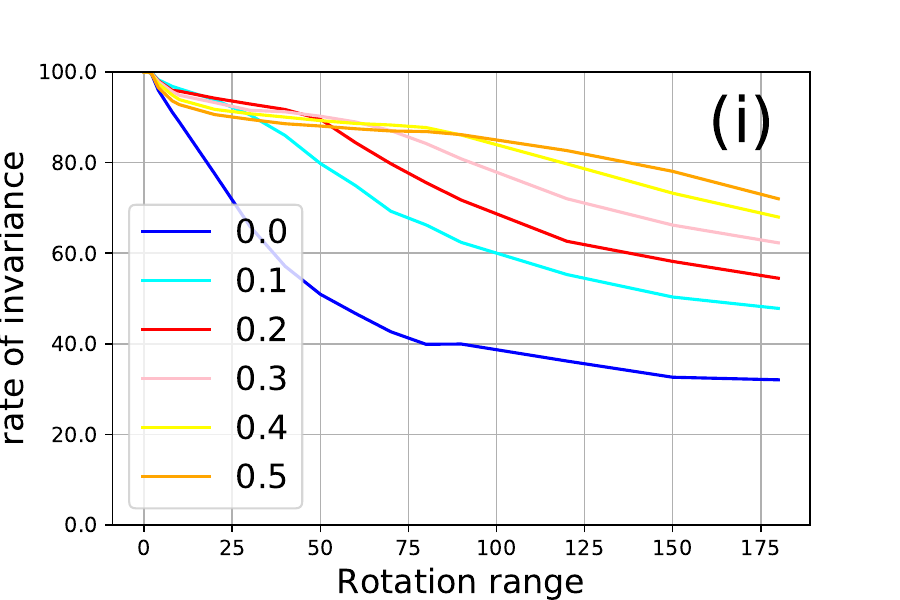}
\includegraphics[width=0.24\linewidth]{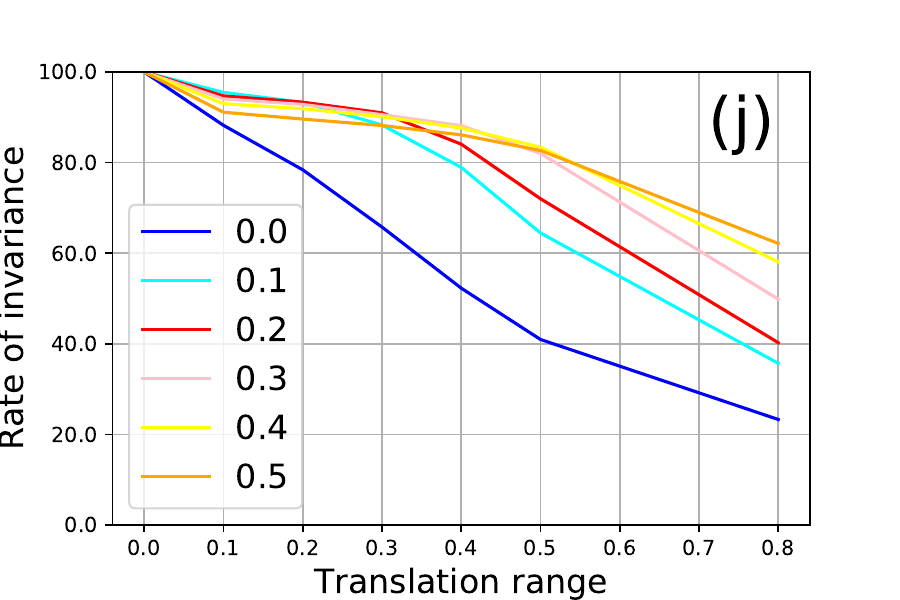}
\includegraphics[width=0.24\linewidth]{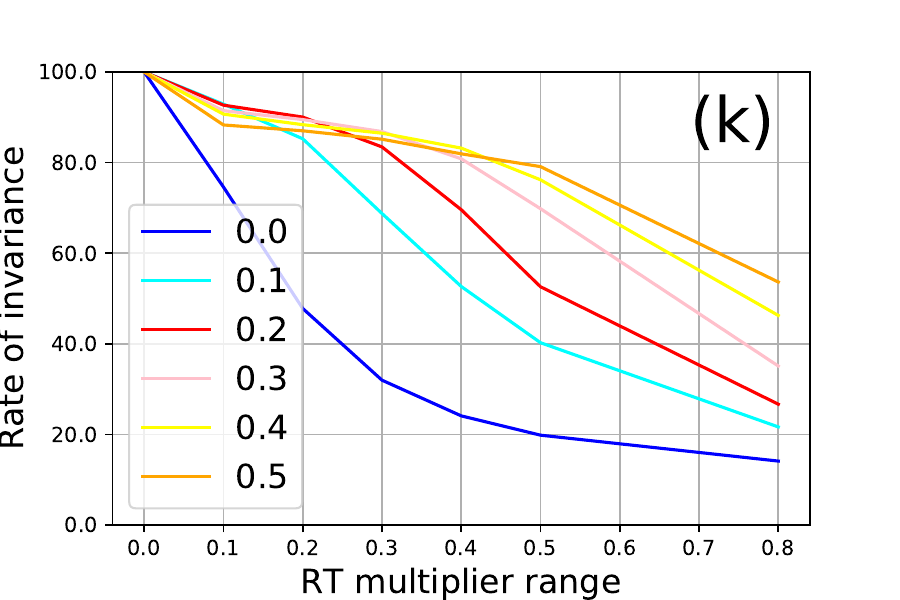}
\includegraphics[width=0.24\linewidth]{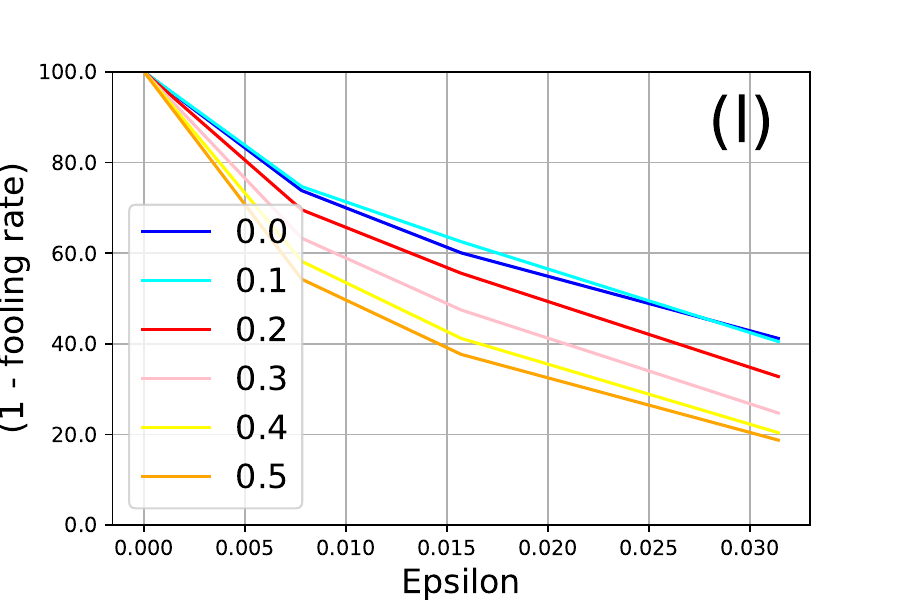}
\includegraphics[width=0.24\linewidth]{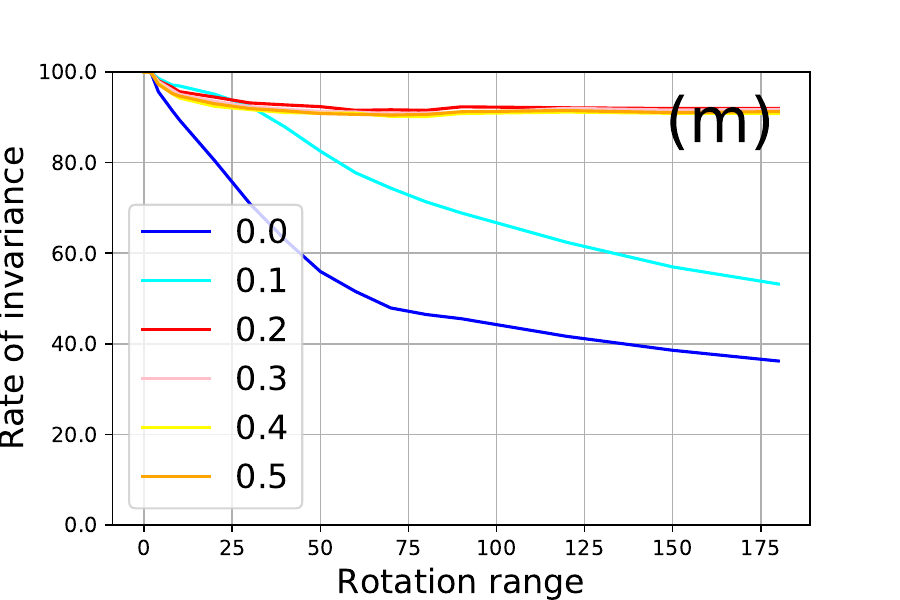}
\includegraphics[width=0.24\linewidth]{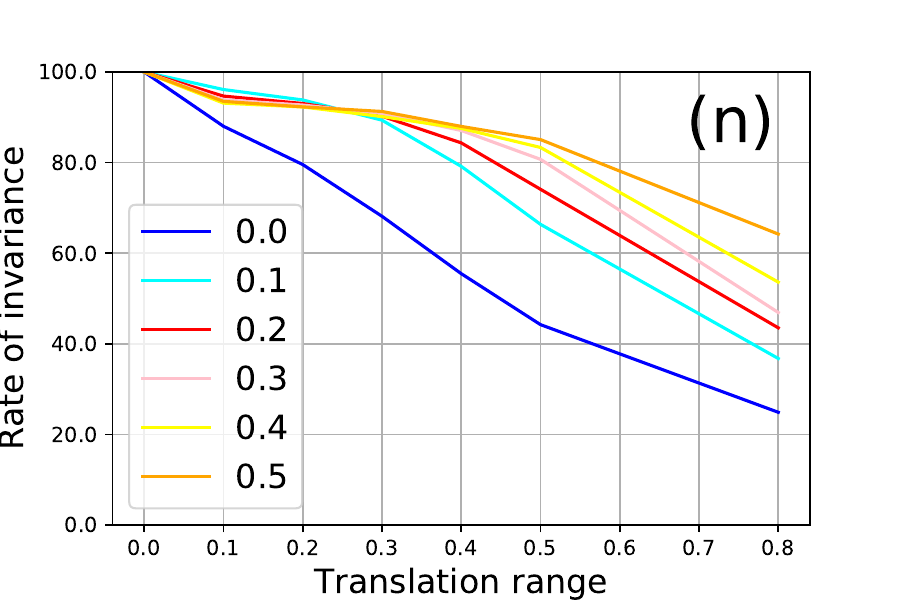}
\includegraphics[width=0.24\linewidth]{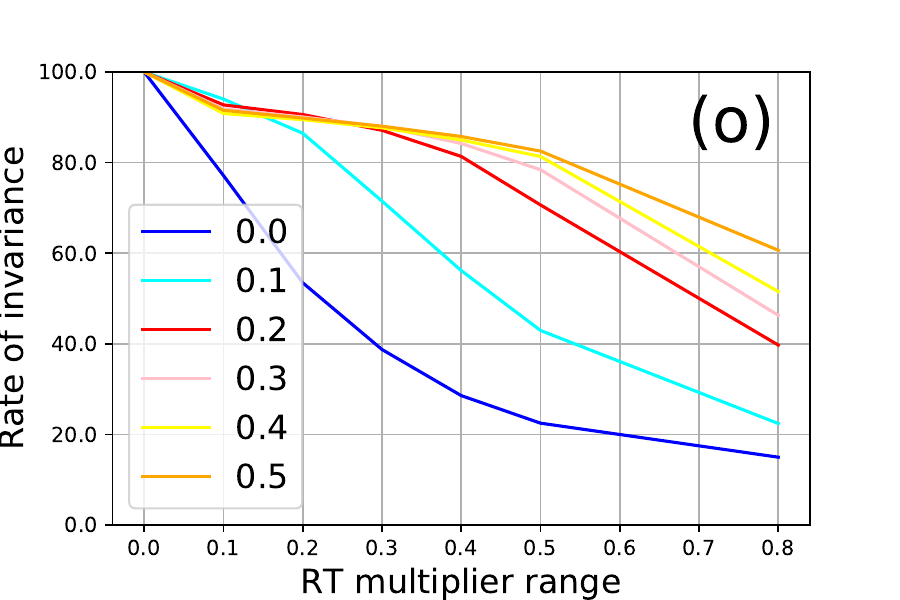}
\includegraphics[width=0.24\linewidth]{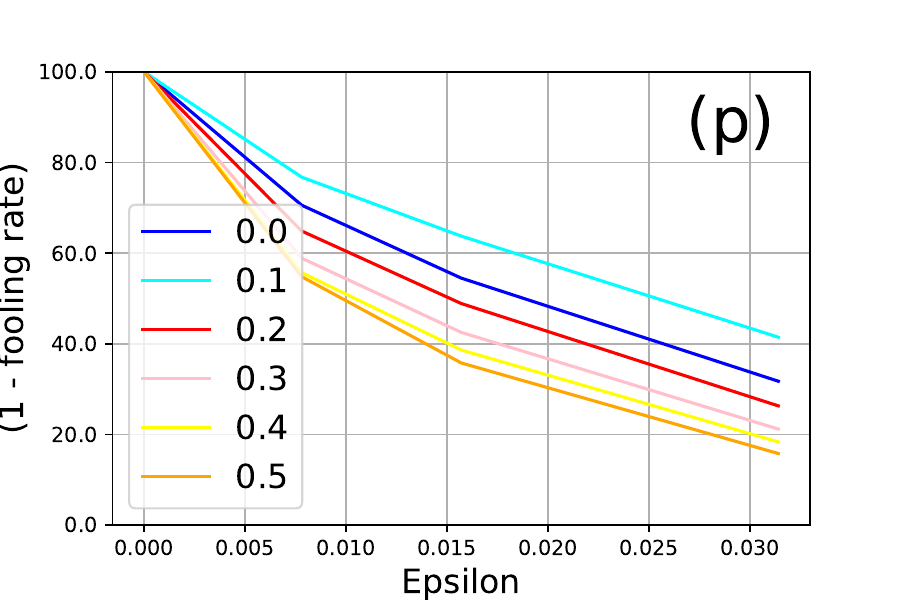}
\end{center}
\caption{On CIFAR10, For VGG16 model (a-b) \textbf{Aug - R} StdCNN, (c-d) \textbf{Aug - R} GCNN, (e-f) \textbf{Aug - T} StdCNN, (g-h) \textbf{Aug - T} GCNN, (i-l) \textbf{Aug - RT} StdCNN, (m-p) \textbf{Aug - RT}, invariance profiles of StdCNN/GCNN models and corresponding robustness profiles.}
\vspace{-8pt}
\label{cifar10-stdcnn-gcnn-vgg16-full}
\end{figure*}

\uline{\textit{Training Data Augmentation:}}
\emph{Spatial}: \begin{inparaenum}[(a)] \item \textbf{Aug - R} : Data is augmented with random rotations in the range $\pm \theta^{\circ}$ given $\theta$, along with random crops and random horizontal flips (for MNIST alone, we do not apply crop and horizontal flips); \item \textbf{Aug - T}: Data is augmented with random translations within $[-i, +i]$ range of pixels in the image, given $i$ (eg. for CIFAR10 with $i=0.1$ is $32*0.1 \approx \pm3 px$) in both horizontal and vertical directions; \item \textbf{Aug - RT}: Data is augmented with random rotations in $\pm i*180^{\circ}$ and random translations within $[-i, +i]$ range of pixels in the image (eg. for CIFAR10 with $i=0.1$ is $0.1*180^{\circ} = \pm 18^{\circ}$ rotation and $32*0.1 \approx \pm3 px$ translation), here no cropping and no horizontal flip is used. We use nearest neighbour interpolation and black filling to obtain the transformed image. \end{inparaenum} \emph{Adversarial} : \textbf{Adv - PGD}: Adversarial training using PGD-perturbed adversarial samples using an $\epsilon$-budget of given $\epsilon$. Our experiments with PGD use a random start, 40 iterations, and step size $0.01$ on MNIST, and a random start, 10 iterations, and step size $2/255$ on CIFAR10, CIFAR100 and Tiny ImageNet. While we study the trade-off under a fixed PGD setting our trend holds even with different PGD hyperparameter settings. Refer supplementary \ref{suppl:exp-pgd-hyperparameters} for details. Our results, presented in the next section, are best understood by noting the augmentation method mentioned in the figure caption. For example, in Fig \ref{cifar10-stdcnn-gcnn-vgg16-full}(a), the augmentation scheme used is \textbf{Aug-R}. The red line (annotated as 60 in the legend) corresponds to the model trained with random rotation augmentations in the range $\pm60^{\circ}$.

\uline{\textit{Hardware Configuration:}}
We used a server with 4 Nvidia GeForce GTX 1080i GPU to run all the experiments in the paper.

\uline{\textit{Evaluation Metrics:}}
We quantify performance using a spatial invariance profile and an adversarial robustness profile. 

\textit{Spatial invariance:} We quantify rotation invariance by measuring the rate of invariance or the fraction of test images whose predicted label remains the same after rotation by a random angle between $[-\theta^{\circ}, \theta^{\circ}]$. As $\theta$ varies from $0$ to $180$, we plot the rate of invariance. We call this the \emph{rotation invariance profile} of a given model. Similarly, by training a model with other augmentation schemes given above, we obtain a \emph{translation invariance profile} and a \emph{rotation-translation invariance profile} for a given model.

\textit{Adversarial robustness:} Similarly, we quantify the $\ell_{\infty}$ adversarial robustness of a given model to a fixed adversarial attack (e.g., PGD) and a fixed $\ell_{\infty}$ norm $\epsilon \in [0, 1]$ by (1 - fooling rate), i.e., the fraction of test inputs for which their predicted label does not change after adversarial perturbation. We plot this for $\epsilon$ varying from $0$ to $1$. We call this the \emph{robustness profile} of a given model. (One can also choose accuracy instead of 1 - fooling rate and we observe similar trends, as shown in the supplementary. We use 1 - fooling rate for clearer visual presentation of our main results.)

\noindent \textbf{Results.} We now present our results on spatially robust model architectures followed by adversarially robust ones.

\uline{\textit{Results on Spatially Robust Model Architectures:}}
Fig \ref{cifar10-stdcnn-gcnn-vgg16-full} present our results on the trade-off on adversarial robustness for VGG16, trained with spatial augmentations, on CIFAR10. (More results including ResNet18 on CIFAR10, VGG16/ResNet18 on CIFAR100 and the StdCNN/GCNN-based model on MNIST are deferred to the supplementary owing to space constraints.)


For any fixed $\theta \in [0, 180]$, we take an equivariant model, namely, StdCNN or GCNN, and augment its training data by random rotations from $[-\theta^{\circ}, +\theta^{\circ}]$. 
For example, Fig \ref{cifar10-stdcnn-gcnn-vgg16-full}(b) show how the robustness profile of StdCNN changes on CIFAR10, as we increase the degree $\theta$ used in training augmentation of the model. We use PGD attack provided in versions of AdverTorch \cite{ding2019advertorch} to obtain the robustness profiles.
Similarly, Figs \ref{cifar10-stdcnn-gcnn-vgg16-full}(a) show the rotation invariance profile of the same models on CIFAR10. 
The black line in Fig \ref{cifar10-stdcnn-gcnn-vgg16-full}(b) shows that the adversarial robustness of a StdCNN which is trained to handle rotations up to $\pm 180$ degrees on CIFAR10, drops by more than 50\%, even when the $\epsilon$ budget for PGD attack is only $2/255$. Similarly, the black line in Fig \ref{cifar10-stdcnn-gcnn-vgg16-full}(a) shows this model's rotation invariance profile - this model is invariant to larger rotations on test data. This can be contrasted with the model depicted by the red line - this StdCNN is trained to handle rotations up to $60$ degrees. The rotation invariance profile of this model is below that of the model depicted by the black line and hence is lesser invariant to large rotations. However, this model can handle adversarial $\ell_{\infty}$-perturbations up to $2/255$ on unrotated data, with an accuracy more than 10$\%$, as seen from the red line in Fig \ref{cifar10-stdcnn-gcnn-vgg16-full}(b). The remaining plots in Figs \ref{cifar10-stdcnn-gcnn-vgg16-full} show similar trends for translation, as well as a combination of rotation and translation (e.g. plots (e-p) in Figs \ref{cifar10-stdcnn-gcnn-vgg16-full} for CIFAR10) for both StdCNN and GCNN models (e.g. Figs \ref{cifar10-stdcnn-gcnn-vgg16-full}c,d).

The above results show that {\em the spatial robustness of these models improves by training augmentation but at the cost of their adversarial robustness, indicating a trade-off between spatial and adversarial robustness.} This trade-off exists in both StdCNN and GCNN models. 

\begin{figure*}[!h]
\begin{center}
\includegraphics[width=0.24\linewidth]{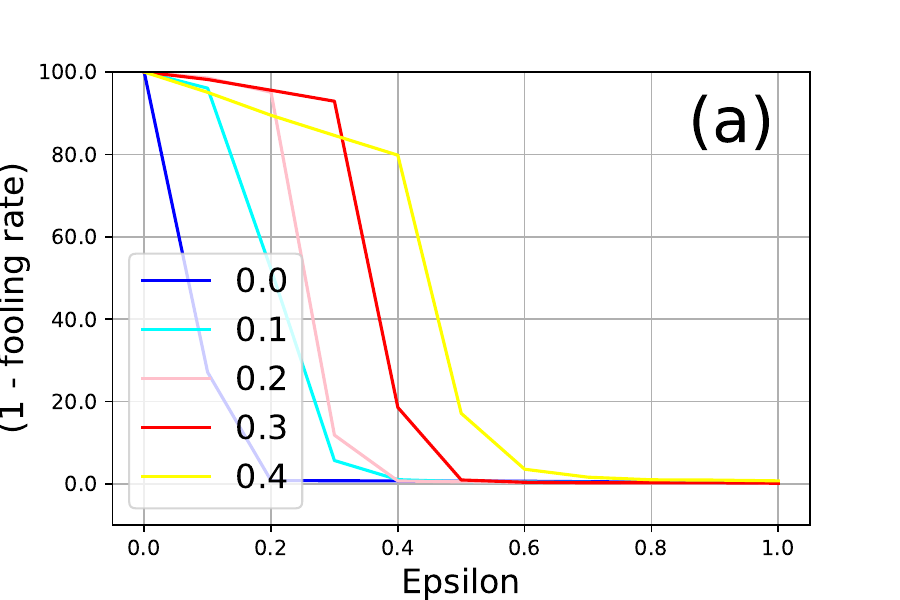}
\includegraphics[width=0.24\linewidth]{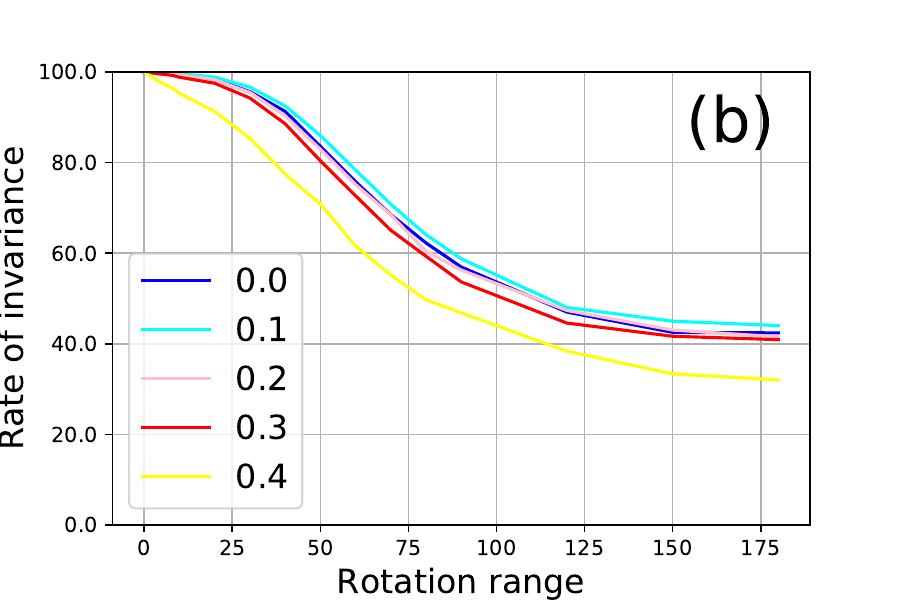}
\includegraphics[width=0.24\linewidth]{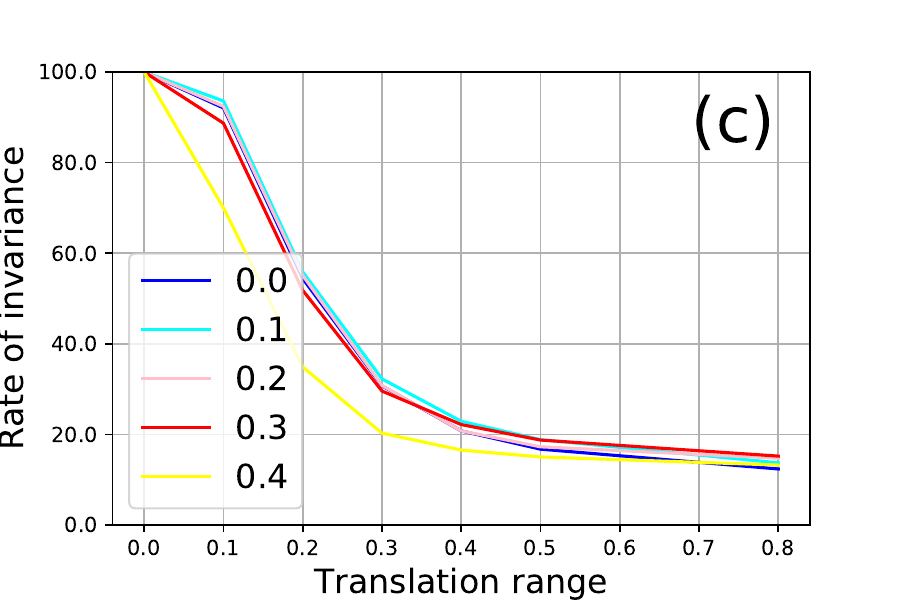}
\includegraphics[width=0.24\linewidth]{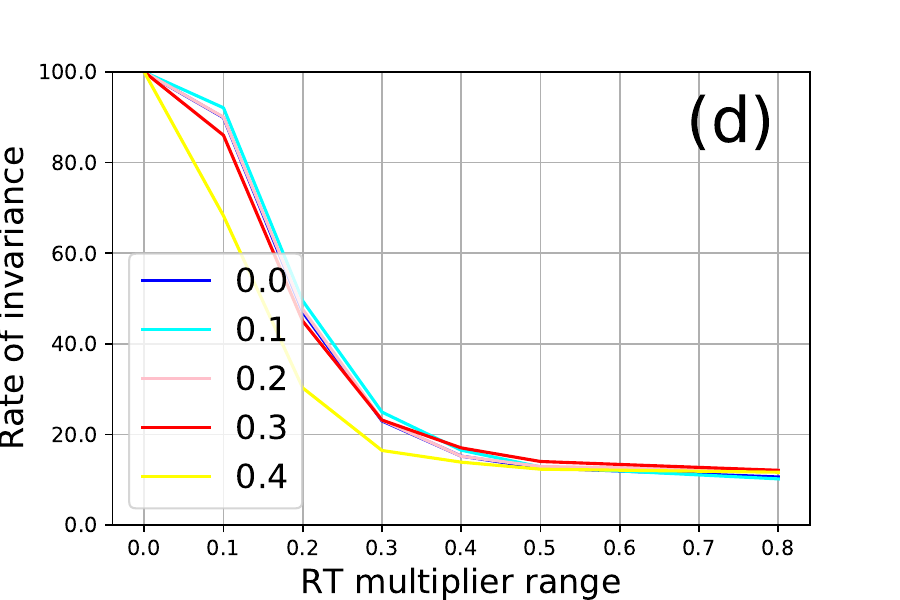}
\includegraphics[width=0.24\linewidth]{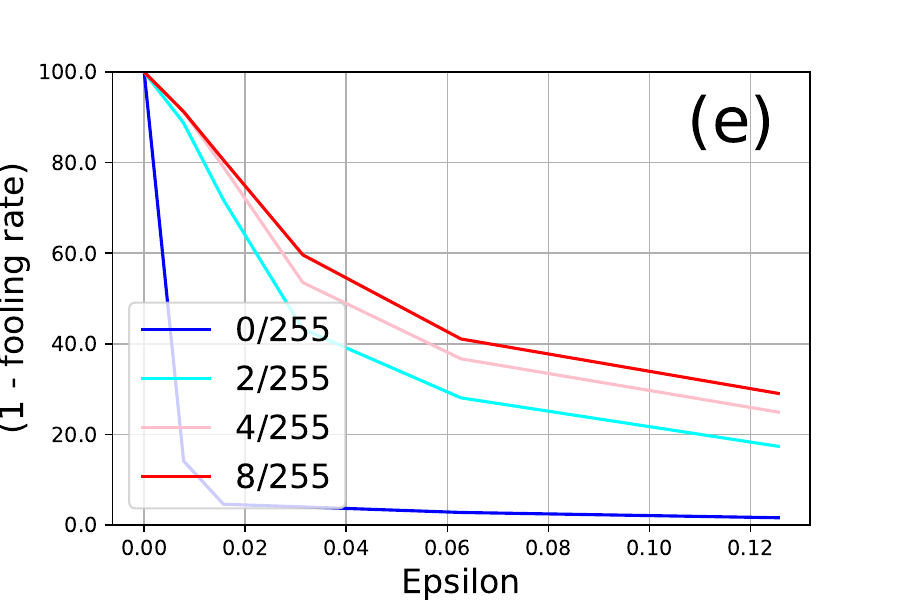}
\includegraphics[width=0.24\linewidth]{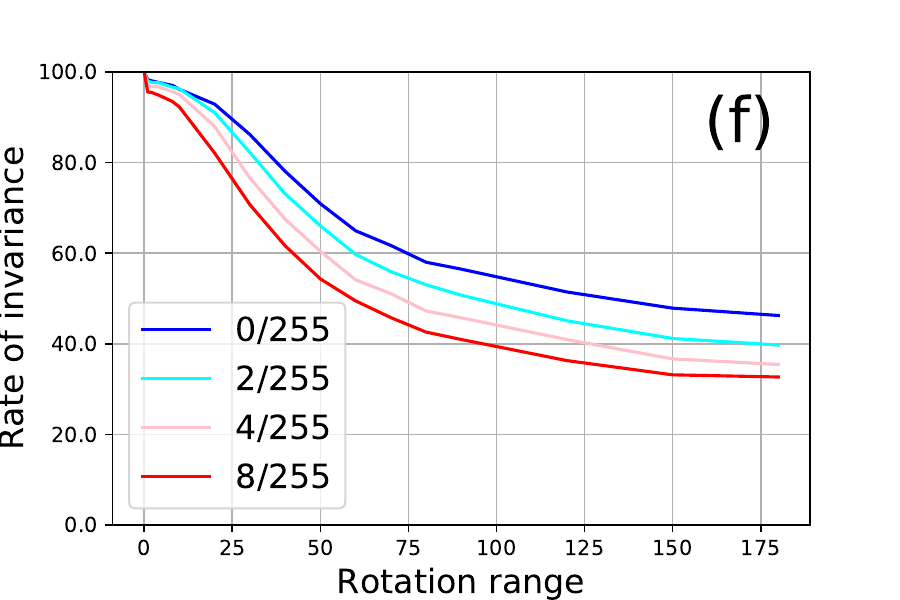}
\includegraphics[width=0.24\linewidth]{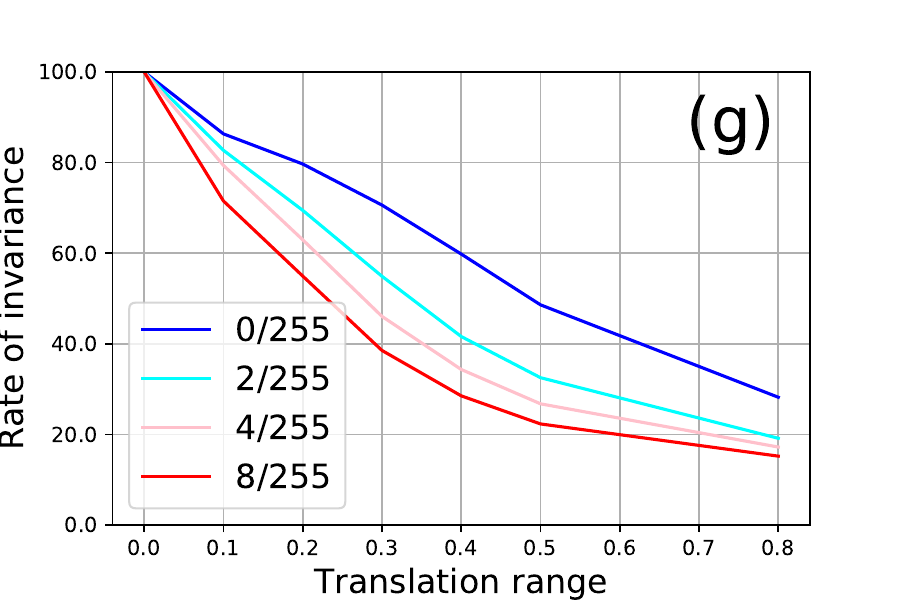}
\includegraphics[width=0.24\linewidth]{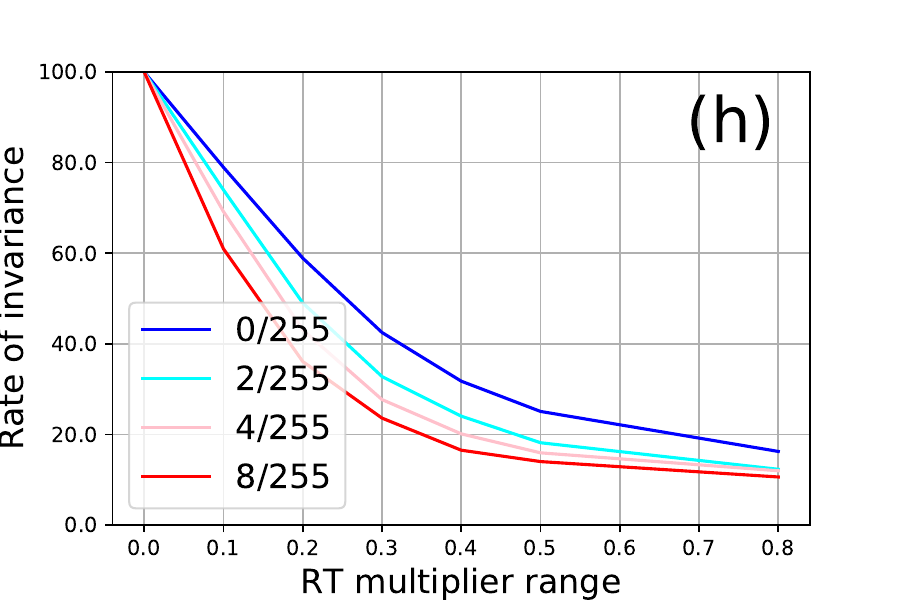}
\includegraphics[width=0.24\linewidth]{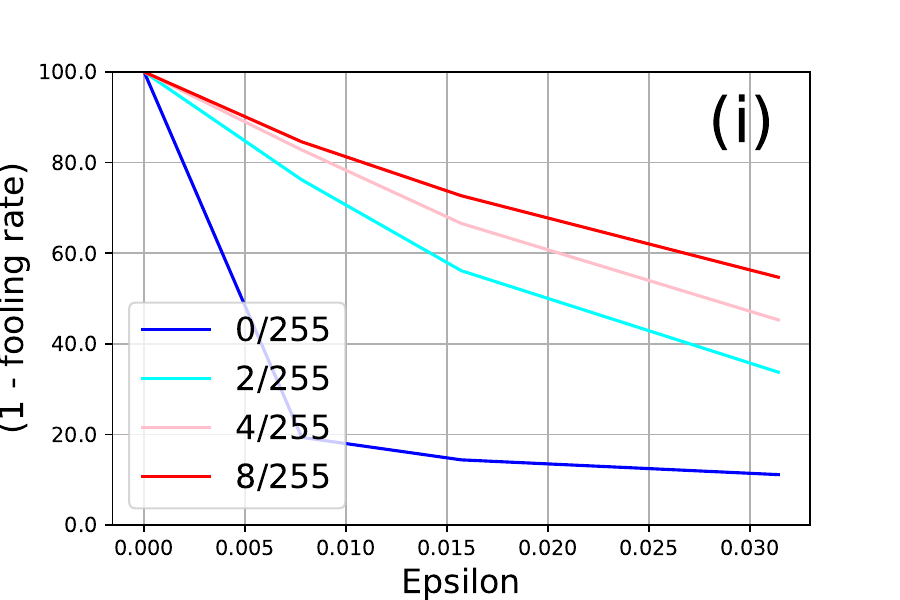}
\includegraphics[width=0.24\linewidth]{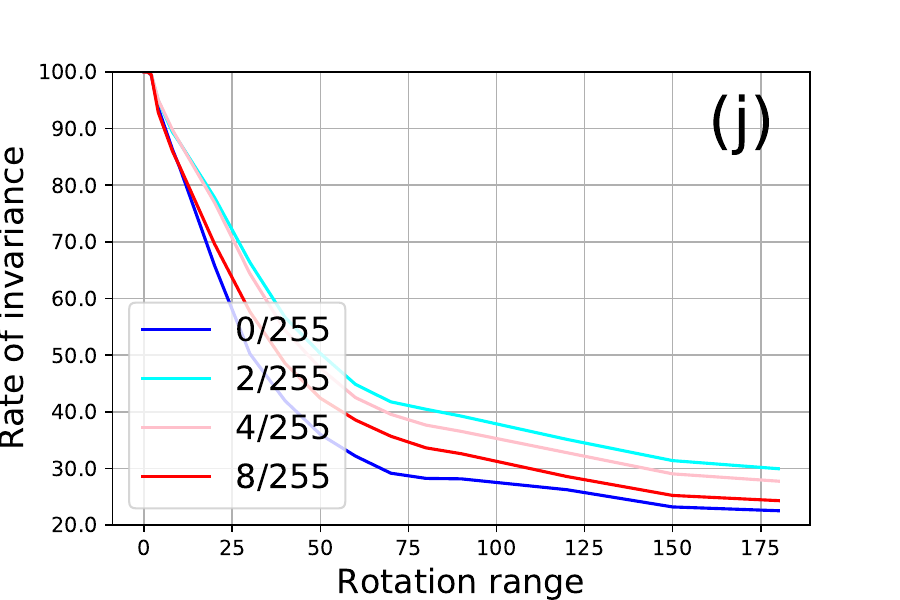}
\includegraphics[width=0.24\linewidth]{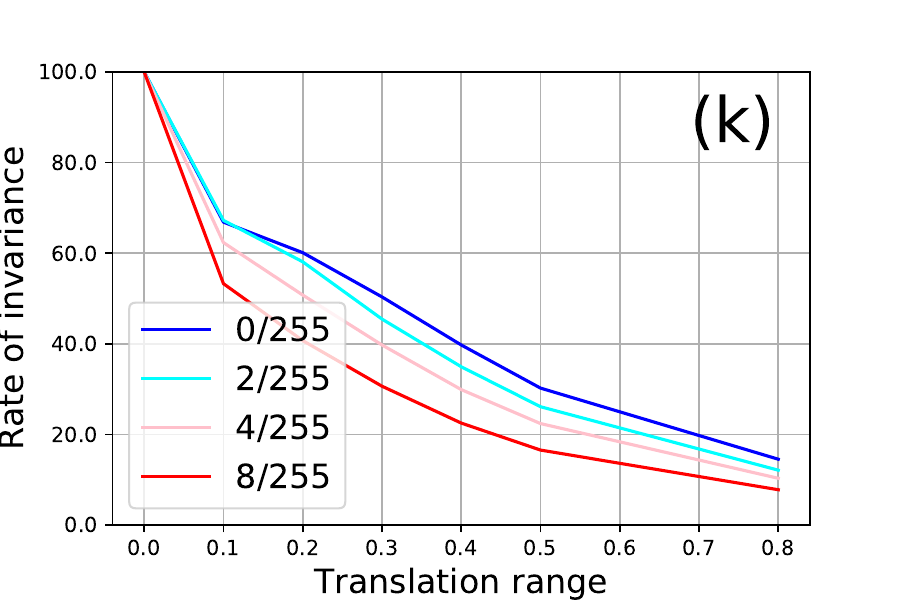}
\includegraphics[width=0.24\linewidth]{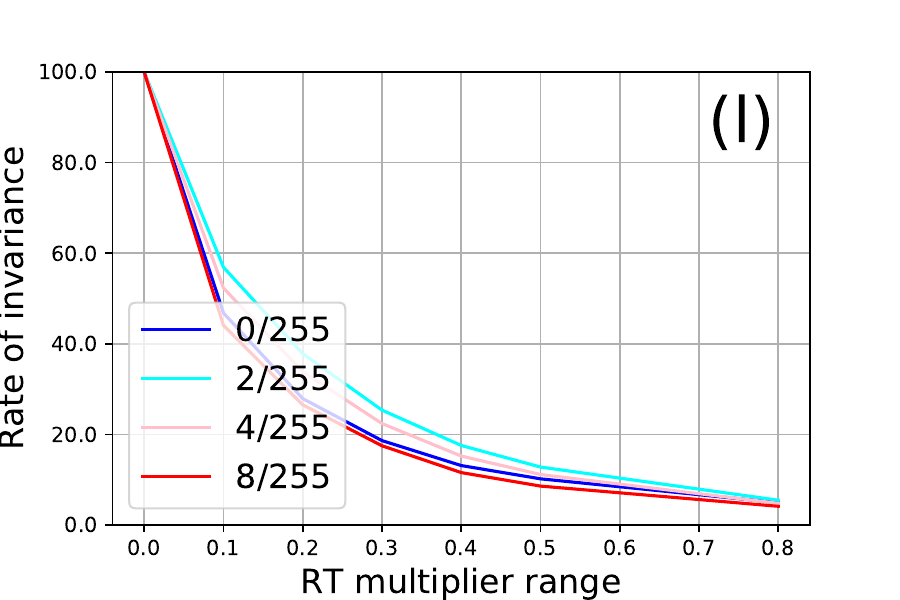}
\end{center}
\caption{\textbf{Adv - PGD}, (1) Robustness, (2) Rotation, (3) Translation, (4) Rotation-Translation invariance profiles for PGD adversarially trained models. (a-d) LeNet based model from \protect \cite{Madry18} on MNIST (e-h) ResNet based model from \protect \cite{Madry18} on CIFAR10, (i-l) WideResNet34 model on CIFAR100. Different colored lines represent models adversarially trained with different $\ell_{\infty}$ budgets $\epsilon \in [0, 1]$.}
\vspace{-8pt}
\label{robust-models-profiles}
\end{figure*}

\uline{\textit{Results on Adversarially Robust Model Architectures:}}
Fig \ref{robust-models-profiles} presents our results on the trade-off on spatial robustness for models trained adversarially using PGD \cite{Madry18} on MNIST, CIFAR10 and CIFAR100 datasets. Each row corresponds to experiments for a single dataset with a plot of robustness profile (for which they are trained) followed by the spatial invariance profiles. Each colored line in the plots corresponds to a model adversarially trained with a different value of an $\epsilon$-budget. 

On MNIST, adversarial training with PGD with larger $\epsilon$ results in a drop in the invariance profile of the LeNet-based model; in Fig \ref{robust-models-profiles}(b-d), the red line (PGD with $\epsilon=0.3$) is below the light blue line (PGD with $\epsilon=0.1$). Similar observations hold for the ResNet model on CIFAR10 (see Fig \ref{robust-models-profiles} (f-h)), as well as for WideResNet-34 on CIFAR100 (see Fig \ref{robust-models-profiles}(j-l)). To complete this picture, the robustness profile curves confirm that as these models are trained with PGD using larger $\epsilon$ budget, their adversarial robustness increases. The robustness profile curves of the LeNet model trained with a larger PGD budget dominates the robustness profile curve of the same model trained with a smaller PGD budget; the red line in Fig \ref{robust-models-profiles}(a) dominates the light blue line. This is true of the ResNet model too, as in Figs \ref{robust-models-profiles}(e) and \ref{robust-models-profiles}(i).

\begin{figure*}[!h]
\begin{center}
\includegraphics[width=0.24\linewidth]{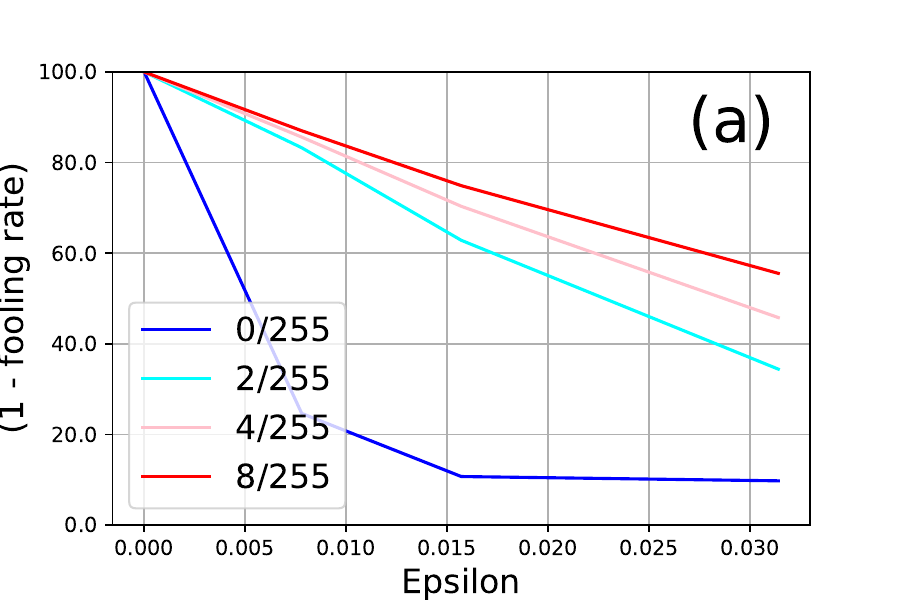}
\includegraphics[width=0.24\linewidth]{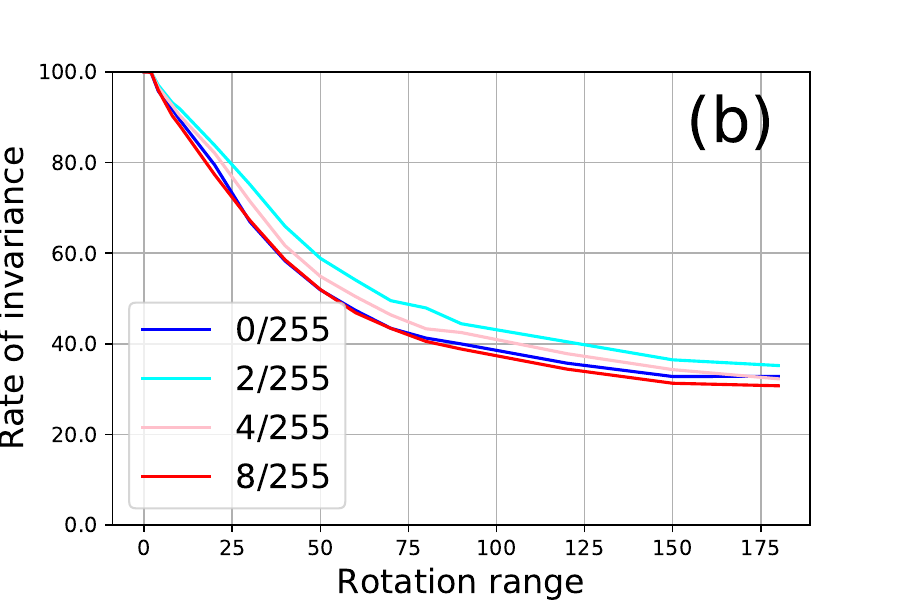}
\includegraphics[width=0.24\linewidth]{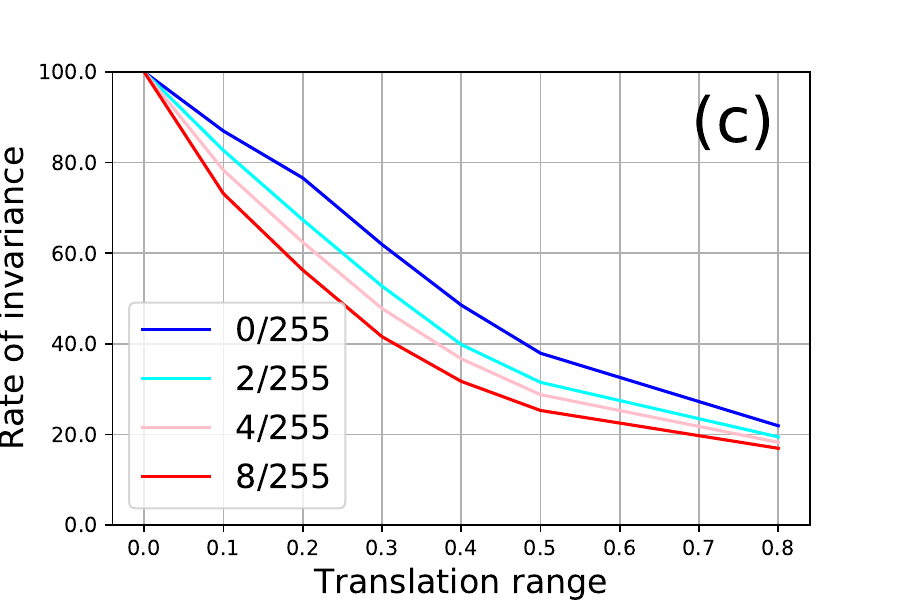}
\includegraphics[width=0.24\linewidth]{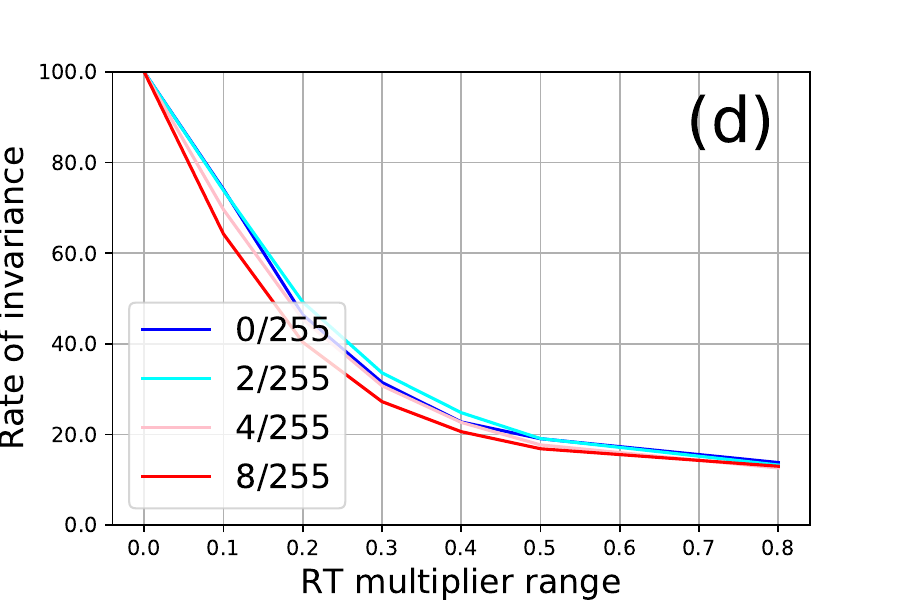}
\end{center}
\caption{On CIFAR10, \textbf{Adv - PGD}, For StdCNN/VGG16 model (a) Robustness profile, (b) Rotation invariance profile, (c) Translation invariance profile (d) Rotation-Translation invariance profile. Different colored lines represent models adversarially trained with different $\ell_{\infty}$ budgets $\epsilon \in [0, 1]$.}
\vspace{-8pt}
\label{cifar10-stdcnn-vgg16-adv-train}
\end{figure*}

In other words, {\em adversarial training with progressively larger $\epsilon$ leads to a drop in the rate of spatial invariance on test data.} Fig \ref{cifar10-stdcnn-vgg16-adv-train} shows an additional result with StdCNN (VGG-16) with PGD-based adversarial training on CIFAR10. These results too support the same findings.


\vspace{-18pt}

\uline{\textit{Results on Tiny ImageNet:}}
Appendix \ref{suppl:exp-results-tiny-imagenet} studies both adversarial robustness of spatially robust models, as well as spatial invariance of adversarially robust models on the Tiny ImageNet dataset. Our observations mentioned so far holds here too, corroborating our claim that this trade-off is not merely theoretical but noticed in real-world datasets too.

\begin{figure*}[!h]
\begin{center}
\includegraphics[width=0.24\linewidth]{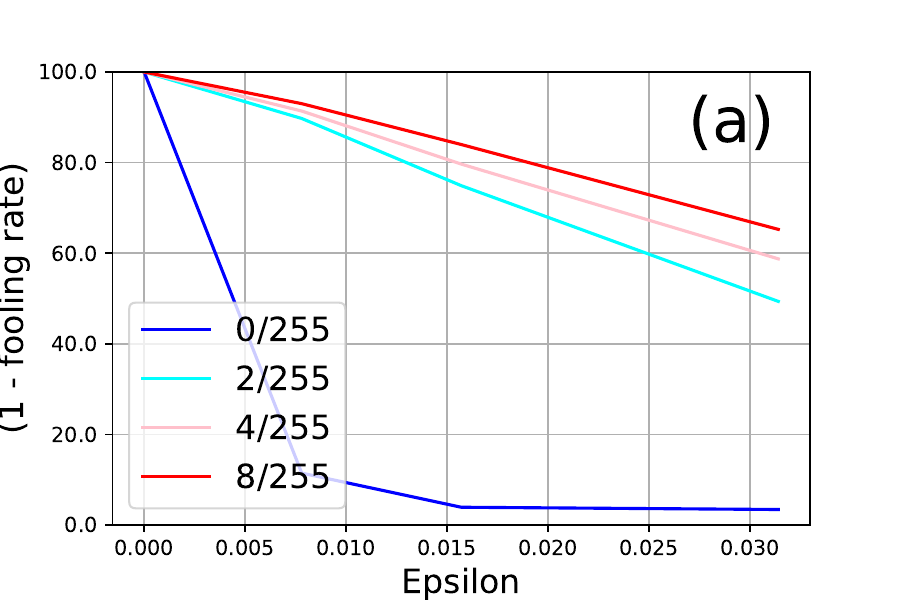}
\includegraphics[width=0.24\linewidth]{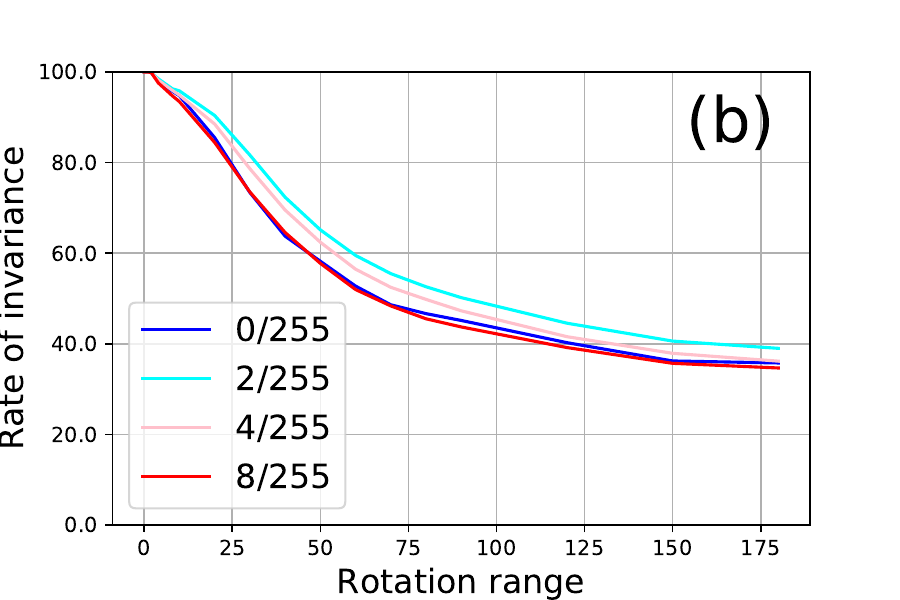}
\includegraphics[width=0.24\linewidth]{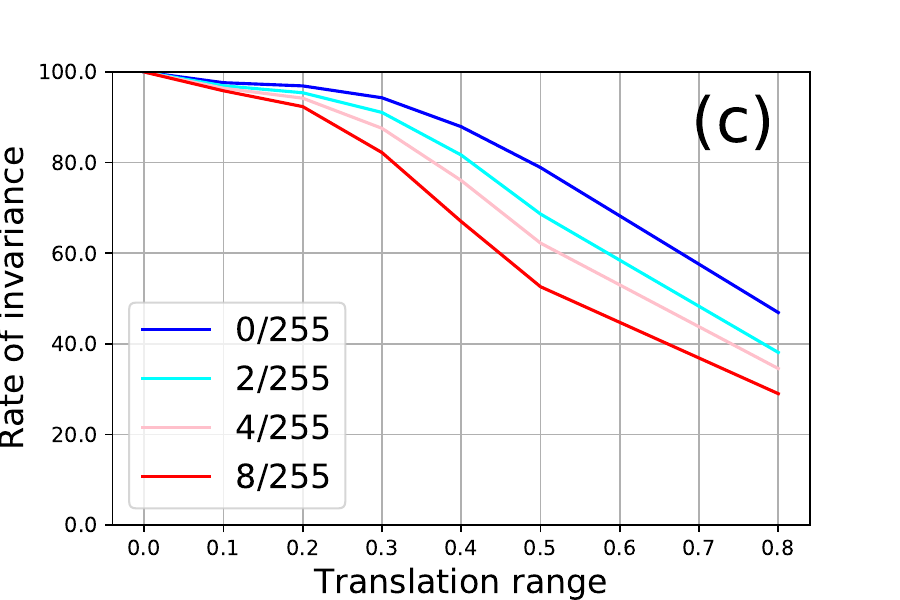}
\includegraphics[width=0.24\linewidth]{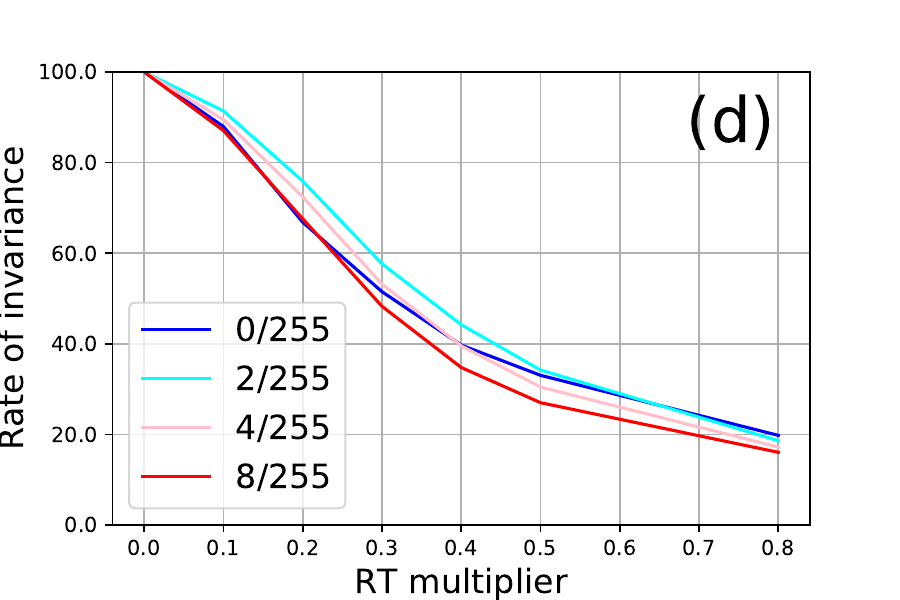}
\end{center}
\caption{On CIFAR10, \textbf{Adv - TRADES}, For StdCNN/WideResNet34 model (a) Robustness , (b) Rotation invariance , (c) Translation invariance (d) Rotation-Translation invariance profiles. Different colored lines represent models adversarially trained with different $\ell_{\infty}$ budgets $\epsilon \in [0, 1]$.}
\label{trades-cifar10-wres34}
\end{figure*}

\uline{\textit{Going beyond PGD-based Adversarial Training:}}
While we have used the popular PGD-based adversarial training for the study of robust models in all the results so far, we conduct experimental studies with another recent popular adversarial method, TRADES \cite{zhang2019theoretically}. Fig \ref{trades-cifar10-wres34} shows the results of these studies on CIFAR10, where we observe the same qualitative behavior of the models. 

\begin{figure*}[!h]
\begin{center}
\includegraphics[width=0.24\linewidth]{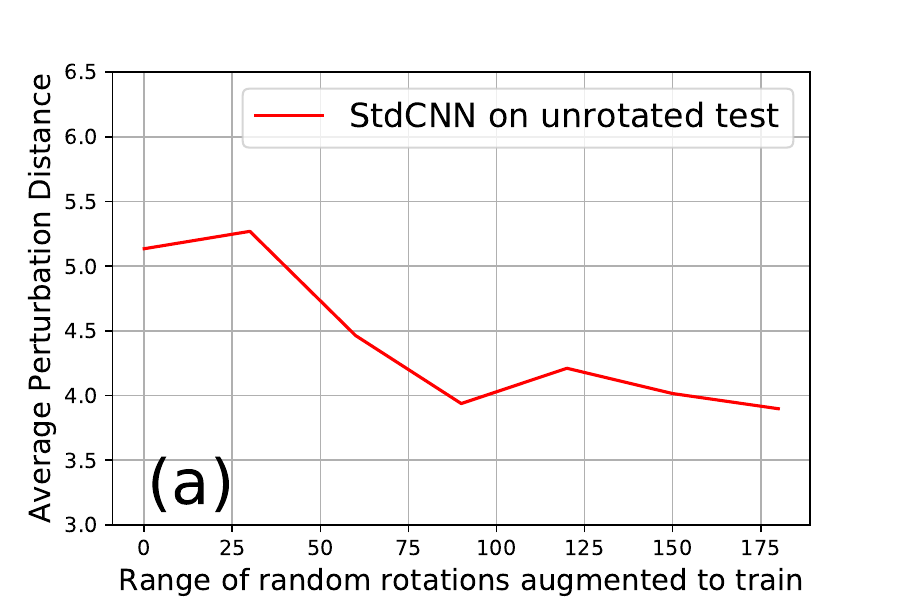}
\includegraphics[width=0.24\linewidth]{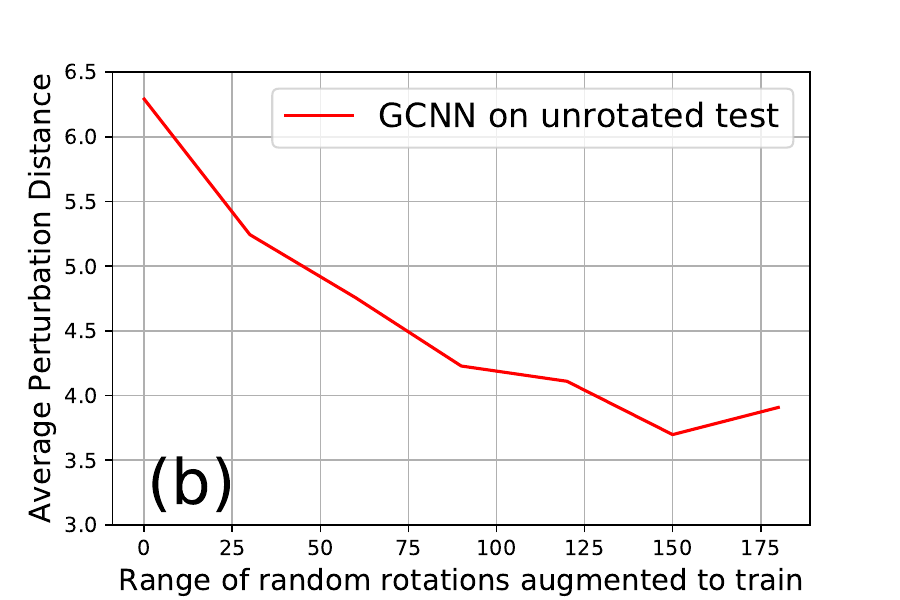}
\includegraphics[width=0.24\linewidth]{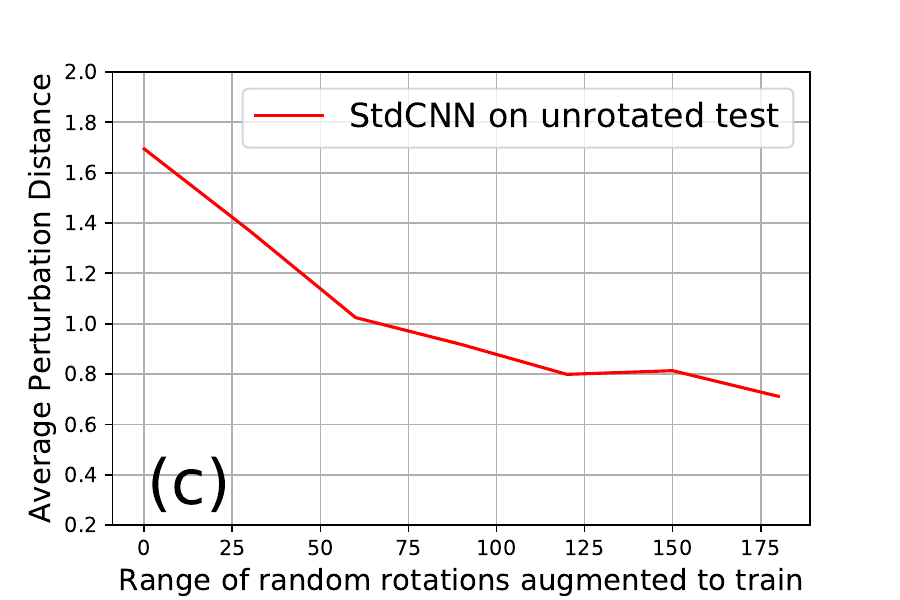}
\includegraphics[width=0.24\linewidth]{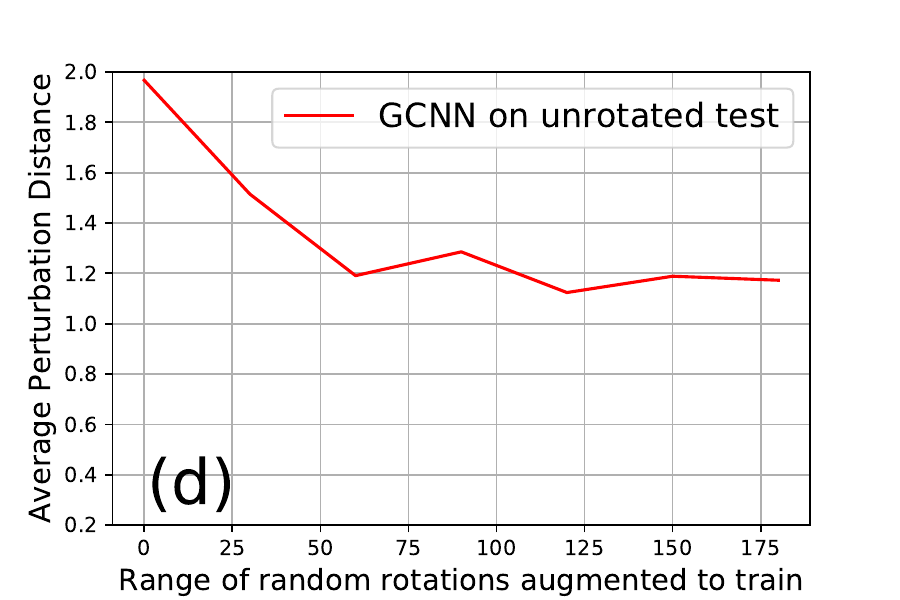}
\end{center}
\vspace{-3pt}
\caption{Average perturbation distance of StdCNN/GCNN on (a-b) MNIST(based on appendix Table \ref{gcnn-table}), (c-d) CIFAR10(VGG16) wrt unrotated test data.} 
\label{stdcnn-gcnn-pgd-norot-rot-dist-fig-1}
\end{figure*}

\uline{\textit{Additional Empirical Evidence (Average Perturbation Distance to Boundary) :}}
While more results and implementation details are provided in the supplementary section, we briefly provide an interesting evidence to our claims based on the average distance to the decision boundary (details in Appendix \ref{suppl:dist-to-bdry}). We notice that there is a distinct reduction in the distance to decision boundary as model achieves better spatial invariance (Figure \ref{stdcnn-gcnn-pgd-norot-rot-dist-fig-1}). This reduction in distance can be linked to the drop in adversarial robustness. Such a study also has the potential to be used to study trade-offs in the presence of other common corruptions \cite{hendrycks2018benchmarking,hendrycks2019nae,hendrycks2020many}.

\section{Curriculum Learning-based Approach for Pareto-Optimality}
The progressive nature of the trade-off between spatial invariance and adversarial robustness elicits the need to study training methods that can simultaneously achieve robustness on both of these fronts.
In this section, we propose a curriculum learning-based training strategy that consistently provides models close to the Pareto-optimal frontier of the trade-off.
An obvious strategy to obtain both kinds of robustness would be to sequentially combine adversarial training and augmentation by spatial transformations, i.e., take an adversarially trained network and re-train it with spatial (e.g., rotation) augmentation, or vice versa (Appendix \ref{suppl:other-exp}). We however observe that such a strategy presents a version of catastrophic forgetting where the robustness learned earlier is forgotten when the other robustness is enforced. A model thus trained has either good spatial robustness or good adversarial robustness but not both. We hence explore a curriculum learning strategy to this end.

Curriculum learning is inherently inspired by how humans learn; first learning simple tasks and then followed by gradually difficult tasks \cite{bengio2009curriculum}. Previous work has demonstrated its effectiveness in training DNN models \cite{weinshall2018curriculum,hacohen2019power}. A curriculum-based adversarial training has also been proposed \cite{cai2018curadv}, which however focuses only on adversarial robustness.
For simultaneous spatial and adversarial robustness, we propose CuSP ({\bf Cu}rriculum based {\bf S}patial-Adversarial Robustness training for {\bf P}areto-Optimality), where the difficulty of spatial augmentation and adversarial strength is gradually increased each time the learning rate is updated as part of a schedule. In particular, each time the learning rate is updated, we increase the spatial augmentation range $[-\theta^{\circ}, \theta^{\circ}]$ as well as the adversarial perturbation bound $\epsilon$ and adversarial train on a randomly perturbed input within that range. This strategy reflects an optimization procedure where the difficulty of robustness handled by the model is increased when the model treads more slowly (lower learning rate) along areas of the loss surface. The outline of the proposed CuSP approach ({\bf Cu}rriculum based {\bf S}patial-Adversarial Robustness training for {\bf P}areto-Optimality) is given in Algorithm \ref{alg:cur-pareto}. 
\begin{algorithm}[]
\SetAlgoLined
\SetKwInOut{Input}{Input}
\SetKwInOut{Parameters}{Parameters}
\SetKwInOut{Output}{Output}
\Input{Training data $(X,Y)$}
\Parameters{No. of epochs: K, Learning rate: $\eta$, Adversarial training (AT) method: PGD or TRADES, Range of spatial transformations: $\theta_{1} \leq \theta_{2} \leq \dotsc \leq \theta_{l}$, \\ Adversarial perturbation bounds: $\epsilon_{1} \leq \epsilon_{2} \leq \dotsc \leq \epsilon_{s}$, \\ Learning rate schedule: $\eta_1 > \eta_2 > \dotsc > \eta_{t}$}
\Output{Model parameters $\phi$}
Initialize model parameters $\phi$; $i,j,k \leftarrow 1$\\
$\theta \leftarrow \theta_{i}, \epsilon \leftarrow \epsilon_{j}, \eta \leftarrow \eta_k$\\ 
\For{epochs $t = 1, \dotsc, E$}{
\lIf{$k \leftarrow k+1$; $\eta \leftarrow \eta_k$ \tcc{Change in learning rate schedule}}{$i \leftarrow i + 1$; $j \leftarrow j + 1$; Update $\theta \leftarrow \theta_{i}$ and $\epsilon \leftarrow \epsilon_{j}$}
$\phi \leftarrow AdversarialTraining_{\phi}(T_{\theta}(X) + A_{\epsilon}(T_{\theta}(X), \eta))$ \\ \tcc{Minimize loss on input randomly transformed within $[-\theta,\theta]$ and an adversarial attack of norm at most $\epsilon$} 
}
\caption{\footnotesize CuSP: {\bf Cu}rriculum-based {\bf S}patial-Adversarial Robustness Training for {\bf P}areto-Optimality}
\label{alg:cur-pareto}
\end{algorithm}

\begin{table}
\caption{\footnotesize Comparison of performance of proposed CuSP for CIFAR10 on StdCNN/VGG16 with other baseline strategies. (Aug $\theta$): denotes training data augmented with random rotations in the range $[-\theta, +\theta]$; $a \rightarrow b$: denotes $a$ sequentially followed by $b$ during training. The top $7$ values in each column are in boldface.} 
\footnotesize
    \centering
    \begin{tabular}{c | l | c  c  c }
    \hline
    & \bf Training Method&\bf Adv (PGD)&\bf Std &\bf Spatial\\
    &&\bf Accuracy(\%)&\bf Accuracy(\%)&\bf Accuracy(\%)\\
\hline
1 & Natural (Aug 0) & 00.05$\pm$0.06 & {\bf 92.89$\pm$0.34} & 33.99$\pm$0.54 \\
2 & Natural (Aug 180) & 00.00$\pm$0.00 & {\bf 85.00$\pm$1.36} & {\bf 83.78$\pm$0.95} \\
3 & PGD (Aug 0) & {\bf 44.88$\pm$0.22} & {\bf 80.08$\pm$0.58} & 33.07$\pm$0.81 \\
4 & PGD (Aug 180) & 32.79$\pm$0.73 & 54.60$\pm$0.51 & 53.69$\pm$0.48 \\
\hline
5 & PGD (Aug 0) $\rightarrow$ Aug 180 & 00.00$\pm$0.00 & {\bf 85.46$\pm$0.25} & {\bf 83.99$\pm$0.34} \\
6 & Aug 180 $\rightarrow$ PGD (Aug 0) & {\bf 45.80$\pm$1.05} & {\bf 81.03$\pm$0.14} & 33.19$\pm$0.33 \\
7 & PGD (Aug 0) $\rightarrow$ PGD (Aug 180) & 35.19$\pm$0.31 & 58.96$\pm$0.21 & 57.84$\pm$0.15 \\
8 & Aug 180 $\rightarrow$ PGD (Aug 180) & 35.94$\pm$0.35 & 60.27$\pm$0.39 & 59.02$\pm$0.21 \\
\hline
9 & CuSP (PGD, 30-60-180, $\{\frac{2}{255}, \frac{4}{255}, \frac{8}{255}\}$) & {\bf 38.63$\pm$0.11} & 65.92$\pm$0.35 & 53.31$\pm$0.13 \\
10 & CuSP (PGD, 60-120-180, $\{\frac{2}{255}, \frac{4}{255}, \frac{8}{255}\}$) & {\bf 37.18$\pm$0.14} & 65.88$\pm$0.11 & {\bf 59.09$\pm$0.18} \\
11 & CuSP (PGD, 120-150-180,  $\{\frac{2}{255}, \frac{4}{255}, \frac{8}{255}\}$) & {\bf 36.01$\pm$0.36} & 64.96$\pm$0.10 & {\bf 61.41$\pm$0.61} \\
12 & CuSP (PGD, 180, $\{\frac{2}{255}, \frac{4}{255}, \frac{8}{255}\}$) & 33.63$\pm$0.38 & 62.17$\pm$0.50 & {\bf 61.07$\pm$0.43} \\
13 & CuSP (PGD, 120-150-180,$\{\frac{8}{255}\}$) & 35.57$\pm$0.16 & 58.08$\pm$0.26 & 55.54$\pm$0.19 \\
14 & PGD (Aug 0) $\rightarrow$ & & & \\
 & CuSP (PGD, 120-150-180,  $\{\frac{2}{255}, \frac{4}{255}, \frac{8}{255}\}$) & {\bf 36.92$\pm$0.28} & {\bf 66.25$\pm$0.68} & {\bf 62.51$\pm$0.12} \\
15 & Aug 180 $\rightarrow$ & & & \\
 & CuSP (PGD, 120-150-180,  $\{\frac{2}{255}, \frac{4}{255}, \frac{8}{255}\}$) & {\bf 37.16$\pm$0.12} & {\bf 66.58$\pm$0.12} & {\bf 63.04$\pm$0.11} \vspace{.1cm} \\
\hline
\end{tabular}
    \label{tradeoff-table-cifar10-pgd-vgg16}
\end{table}

To study the goodness of the proposed strategy, we look at three metrics: \begin{inparaenum}[(a)] \item spatial accuracy on large random transformations $\prob{f(T_{180}(X))=Y}$, where $T_{180}$ denotes a random rotation in the range $[-180^{\circ}, 180^{\circ}]$, \item
adversarial accuracy at large perturbations $\prob{f(X+A_{8/255}(X))=Y}$, where $A_{8/255}$ denotes an adversarial attack (PGD or TRADES) of $\ell_{\infty}$-norm bounded by $\epsilon = 8/255$, and finally, \item natural accuracy on the unperturbed data $\prob{f(X)=Y}$. 
\end{inparaenum} The last one is included for completeness because we want better spatial and adversarial robustness simultaneously but not lose on natural accuracy in the bargain \cite{Tsipras2019odds,zhang2019theoretically}. 

\begin{figure}[h]
    \centering
    \includegraphics[scale=0.45]{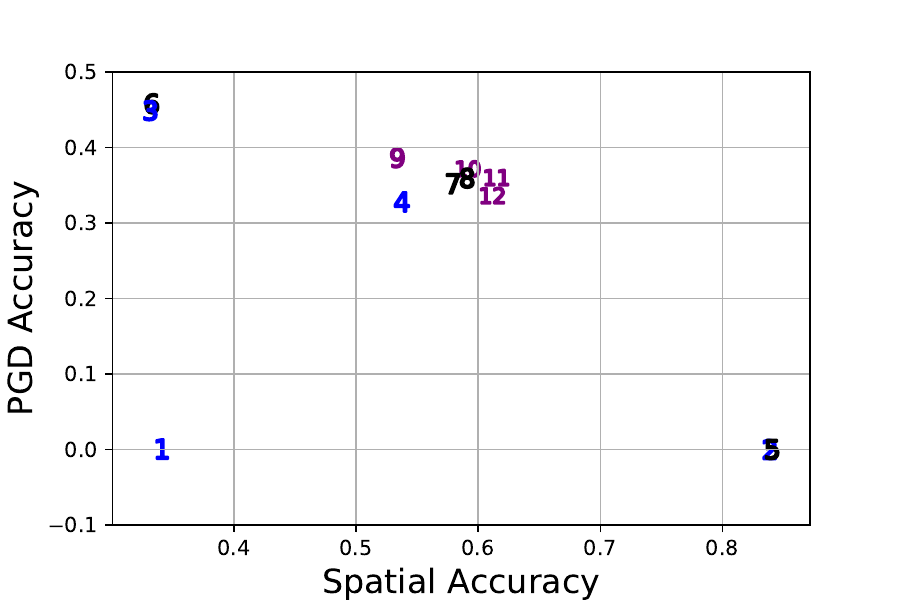}
    \includegraphics[scale=0.45]{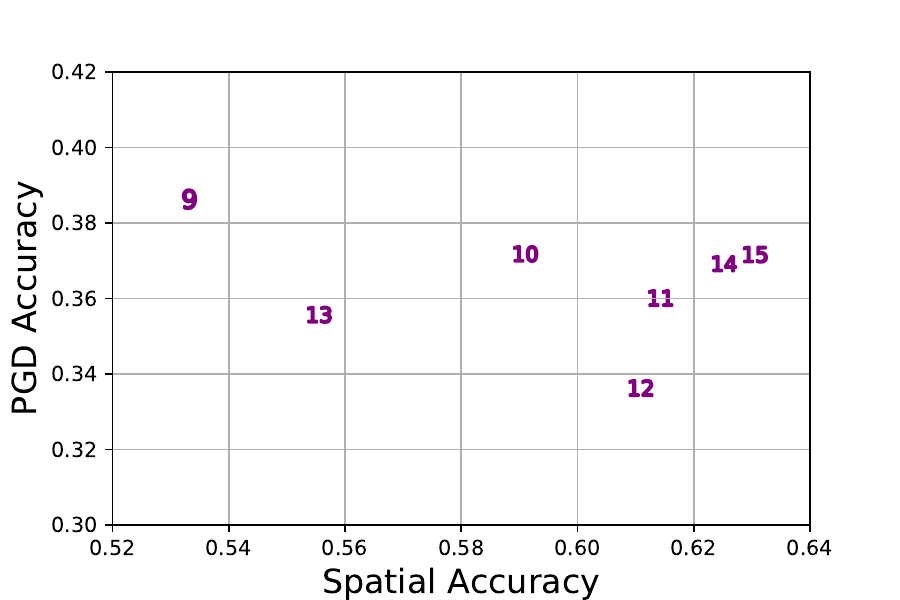}
    \caption{Visualization of performance of CuSP based on PGD against other baseline strategies on CIFAR10 for StdCNN/VGG16 model (each index corresponds to a row in Table \ref{tradeoff-table-cifar10-pgd-vgg16}). On the left, we compare CuSP against various baselines. On the right, we zoom in to compare different variants of CuSP.}
    \label{sample-trade-off-vgg16}
\end{figure}

\uline{\textit{Results:}} We studied CuSP across models and datasets in our experiments, and report the results of VGG16 on CIFAR10 in Table \ref{tradeoff-table-cifar10-pgd-vgg16}. The remainder of these results are presented in Appendix \ref{suppl:pareto-optimal}. We follow the learning rate schedule 75-90-100 as used in TRADES \cite{zhang2019theoretically} where the learning rate starting with $0.1$ decays by a factor of $0.1$ successively at the $75$-th, $90$-th and $100$-th epochs. Our algorithm works with given sets of values for $\theta$ and $\epsilon$ and does not optimize for the choice of these values. We empirically studied various combinations of $(\theta, \epsilon)$ values as $\theta$ varies over $\{0, 30, 60^{\circ}, \dotsc, 180\}$ and $\epsilon$ varies over $\{0, 2/255, 4/255, 8/255\}$. 
While each of these options improves performance over the baselines, we found the list of $\theta = \{120^{\circ}, 150^{\circ}, 180^{\circ}\}$ to give the best results on this particular setting. We also include results for CuSP training after a warm start, viz., PGD with no augmentation and natural training with $[-180^{\circ}, +180^{\circ}]$ augmentation. Figure \ref{sample-trade-off-vgg16} compares the performance of CuSP against other baseline training strategies, visualized in the context of the Pareto frontier for PGD accuracy and spatial accuracy.
\section{Conclusions and Future Work}
In this work, we present a comprehensive discussion on the trade-off between spatial and adversarial robustness in neural network models. While a few recent efforts have studied similar trade-offs, we study random spatial transformations in this work, which are more popularly used in practice than any adversarial spatial transformations. We study the trade-off both theoretically and empirically, and prove a quantitative trade-off between spatial and adversarial robustness in a simple statistical setting used in earlier work. Our experiments show that as equivariant models (StdCNNs and GCNNs) are trained with progressively larger spatial transformations, their spatial invariance improves but at the cost of their adversarial robustness. Similarly, adversarial training with perturbations of progressively increasing norms improves the robustness of well-known neural network models, but with a resulting drop in their spatial invariance. We also provide a curriculum learning-based training method to obtain a solution that is close to Pareto-optimal in comparison to other baseline adversarial training methods. While CuSP gradually increases both the range of spatial transformation $\theta$ in the augmentation and the adversarial perturbation bound $\epsilon$, we conjecture that tuning these further with the learning rate schedule along with a warm start for model parameters could give further improvements.
Recently, \cite{hendrycks2018benchmarking,hendrycks2019nae,hendrycks2020many} have shown that study of other common corruptions is important in the study of state-of-the-art NN models. Similar studies between spatial invariance and such common corruptions is an interesting direction of future work.

\noindent \textbf{Broader Impact.}
Adversarial robustness is one of the most important problems in better understanding the functioning of neural networks. Another important problem is how networks can efficiently handle various spatial transformations like rotation, translation, etc. The trade-off studied and the algorithm proposed to find a better Pareto-optimal in this paper tries to understand the interaction between the two which is a natural question. Hence, we believe this study is towards better understanding of the networks learning capability. There is no known detrimental social impact of our work.

\noindent \textbf{Acknowledgements and Funding Transparency Statement.}
Sandesh Kamath would like to thank Microsoft Research India for funding a part of this work through his postdoctoral research fellowship at IIT Hyderabad.

\bibliographystyle{plain}
\bibliography{spatial-robustness}

\newpage
\appendix
\section*{Supplementary Material : Can we have it all? On the Trade-off between Spatial and Adversarial Robustness of Neural Networks}



The supplementary material contains proofs, additional experiments to show that our results continue to hold across different datasets and different models, and additional empirical evidence for the spatial vs adversarial robustness trade-off that we could not fully include in the main paper owing to space constraints. 

\section{Proof of Proposition 1}
\vspace{-6pt}
This section provides the proof for Proposition \ref{prop:accuracy}. The parameter $\delta$ below is mostly for illustration and not optimized.  We consider $d$ to be large, as the small adversarial perturbation we consider has $\ell_{\infty}$ norm $\leq 4/\sqrt{d}$.

\addtocounter{theorem}{-2}
\begin{prop} 
There exists $p \geq 1/2$ and a cyclic code with relative distance $\delta \geq 3/8$ such that given the input distribution defined as above, the classifier of maximum accuracy on input $(X, Y)$ has accuracy at least $97\%$. Similarly, the classifier of maximum accuracy on the transformed input $(r_{j}(X), Y)$ also has accuracy at least $97\%$. However, when the classifier of maximum accuracy on $(X, Y)$ is applied to $(r_{j}(X), Y)$, for any $j$, it has accuracy at most $85\%$.
\end{prop}
\begin{proof}
Let $\rho(x_{0})$ be equal to $p/(1-p)$, if $x_{0}=1$, and $(1-p)/p$, if $x_{0}=-1$. The Bayes classifier $f^{*}$ that maximizes the accuracy is: 
\begin{align*}
f^{*}(x) & = \sign{\frac{\prob{Y=1, X=x}}{\prob{Y=-1, X=x}} - 1} \\
& = \sign{\rho(x_{0})~ \frac{\prod_{t=1}^{d} \exp\left(\frac{-(x_{t} - 2c_{t}/\sqrt{d})^{2}}{2}\right)}{\prod_{t=1}^{d} \exp\left(\frac{-(x_{t} + 2c_{t}/\sqrt{d})^{2}}{2}\right)} - 1} \\
& = \sign{\frac{\prod_{t=1}^{d} \exp\left(\frac{-(x_{t} - 2c_{t}/\sqrt{d})^{2}}{2}\right)}{\prod_{t=1}^{d} \exp\left(\frac{-(x_{t} + 2c_{t}/\sqrt{d})^{2}}{2}\right)} - \frac{1}{\rho(x_{0})}} \\
& = \sign{\sum_{t=1}^{d} c_{t} x_{t} + \frac{\sqrt{d}}{4}~ \log\rho(x_{0})}, \\
\end{align*}
because the quadratic terms and the constants cancel out. Note that the above classifier of maximum accuracy can be represented as a single decision tree node at $x_{0}$ followed by a linear classifier in the remaining coordinates $x_{1}, x_{2}, \dotsc, x_{d}$. Moreover, $c_{t} = \pm 1$ and $X_{t} \given Y=y$ is normally distributed as $N(2c_{t}y/\sqrt{d}, 1)$, so $\sum_{t=1}^{d} c_{t} X_{t} \given Y=y$ is normally distributed as $Z_{y} = N(2y\sqrt{d}, d)$. As $p \rightarrow 1/2$, we have $\log\rho(x_{0}) \rightarrow 0$. It is known that $\prob{Z \geq \mu - 2\sigma} = \prob{Z \leq \mu + 2\sigma} \approx 0.977$, for any normally distributed $Z$ with mean $\mu$ and standard deviation $\sigma$. So the $\sign{Z_{y}}$ matches $y$ with probability at least $97\%$, and the accuracy of $f^{*}$ is at least $97\%$. 

Similarly, the classifier $f^{*}_{j}$ that maximizes the accuracy on the transformed data $(\tilde{X}, Y) = (r_{j}(X), Y)$ is given by
\[
f^{*}_{j}(\tilde{x}) = \sign{\sum_{t=1}^{d} \left(r_{j}(c)\right)_{t} \tilde{x}_{t} + \frac{\sqrt{d}}{4}~ \log\rho(\tilde{x}_{0})},
\]
and the same proof mutatis mutandis works to show that its accuracy on $(\tilde{X}, Y)$ is at least $97\%$.

Let $S_{j}$ be the subset of coordinates $t$ with $(r_{j}(c))_{t} = c_{t}$. Since our code is binary, we get $(r_{j}(c))_{t} = -c_{t}$, for $t \notin S_{j}$. Therefore,
\[
f^{*}_{j}(x) = \sign{\sum_{t \in S_{j}} c_{t} x_{t} - \sum_{t \notin S_{j}} c_{t} x_{t} + \frac{\sqrt{d}}{4}~ \log\rho(x_{0})}.
\]
Since $c_{t} = \pm 1$ and $X_{t} \given Y=y$ is normally distributed as $N(2c_{t}y/\sqrt{d}, 1)$, we have $c_{t}X_{t} \given Y=y$ normally distributed as $N(2y/\sqrt{d}, 1)$. Moreover, we get that $\sum_{t \in S_{j}} c_{t} X_{t} - \sum_{t \notin S_{j}} c_{t} X_{t} \given Y=y$ is normally distributed as $N(2y(2\size{S_{j}} - d)/\sqrt{d}, 2\size{S_{j}} - d)$. Note that $\size{S_{j}} \leq (1-\delta) d$. Thus, for $\delta \geq 3/8$ we get
\[
\frac{\abs{2y(2\size{S_{j}} - d)}}{\sqrt{2\size{S_{j}} - d}} \leq 2 \sqrt{1-2\delta} = 1.
\]
For the existence of a cyclic code with relative distance $\delta \geq 3/8$, see Chapter 8 of \cite{roth2006intro}. It is known that $\prob{Z \geq \mu - \sigma} = \prob{Z \leq \mu + \sigma} \approx 0.841$, for any normally distributed $Z$ with mean $\mu$ and standard deviation $\sigma$. Using  $p$ sufficiently close to $1/2$, we get that $\prob{f^{*}_{j}(X) = Y}$ is at most $85\%$. Since $r(x)$ is a random transformation $r_{j}(x)$ where $j$ is picked uniformly at random from $\{1, 2, \dotsc, d/m\}$, the same bound holds for the data transformed using $r(x)$.
\end{proof}

\section{Proof of Theorem 2}
\vspace{-6pt}
We now prove Theorem \ref{theorem:trade-off} on the trade-off between spatial and adversarial robustness.

\begin{theorem} \label{theorem:trade-off}
Given the input distribution defined as above, any $\eta > 0$ and a classifier $f: \mathbb{R}^{d+1} \rightarrow \{-1, 1\}$, if the adversarial accuracy is at least $1-\eta$, then the spatial accuracy is at most $\dfrac{\eta~ p}{(1-p)}$. Similarly, if the spatial accuracy is at least $1-\eta$ then the adversarial accuracy is at most $1 - \dfrac{(1-p)(1-\eta)}{p}$.

\end{theorem}

\begin{proof}
Observe that $r_{j}(X)$ only permutes the coordinates of $X$. Hence, 
\begin{align}
\min_{\mathcal{A}} \operatorname{Pr}(f(X + \mathcal{A}(X))=Y) \leq \operatorname{Pr}(f(r_{j}(X)+\mathcal{A}(r_{j}(X)))=Y)
\end{align}
The known adversarial robustness vs. standard accuracy trade-off proof [43] implies, for all $j$,
\begin{align*}
\operatorname{Pr}(f(r_{j}(X)+\mathcal{A}(r_{j}(X)))=Y) \leq 1 - \dfrac{1-p}{p}~ \operatorname{Pr}(f(r_{j}(X))=Y).
\end{align*}
(For completeness, we can add a proof similar to lines 564-565 in the supplementary.) By averaging the above over all $j$, we get
\begin{align*}
\min_{\mathcal{A}} \operatorname{Pr}(f(X + \mathcal{A}(X))=Y) & \leq 1 - \dfrac{m}{d}~ \dfrac{1-p}{p} \sum_{j=1}^{d/m} \operatorname{Pr}(f(r_{j}(X))=Y) \\
& = 1 - \dfrac{1-p}{p} \operatorname{Pr}(f(r(X)=Y).
\end{align*}
Now if the adversarial robustness (i.e., the LHS above) is at least $1-\eta$, then the spatial robustness can be upper bounded as 
\begin{align}
\operatorname{Pr}(f(r(X))=Y) \leq \frac{\eta~ p}{1-p}.
\end{align}
The other side of the trade-off can be proved similarly. If the spatial robustness $\operatorname{Pr}(f(r(X)=Y)$ is at least $1-\eta$ then there exists at least one index $j$ such that $\operatorname{Pr}(f(r_{j}(X))=Y) \geq 1-\eta$. So again using the above adversarial robustness vs. standard accuracy trade-off for that particular index $j$ we get
\begin{align*}
\operatorname{Pr}(f(r_{j}(X)+\mathcal{A}(r_{j}(X)))=Y) & \leq 1 - \dfrac{1-p}{p}~ \operatorname{Pr}(f(r_{j}(X))=Y) \\
& \leq 1 - \frac{(1-p)(1-\eta)}{p}.
\end{align*}
Hence, the adversarial robustness can be upper bounded as follows, using the observation that $r_{j}(X)$ only permutes the coordinates of $X$.
\begin{align*}
\min_{\mathcal{A}} \operatorname{Pr}(f(X + \mathcal{A}(X))=Y) & \leq \operatorname{Pr}(f(r_{j}(X)+\mathcal{A}(r_{j}(X)))=Y) \\
& \leq 1 - \frac{(1-p)(1-\eta)}{p},
\end{align*}
\end{proof}

\section{Tighter Analysis and a Modified Version of Theorem 2}
\vspace{-6pt}
We now give a tighter analysis and prove a modified version of Theorem \ref{theorem:trade-off} towards exhibiting a trade-off between spatial and adversarial robustness. It would be interesting to close the gap between our theoretical trade-off and the progressive or gradual trade-off observed in our experiments. 
\begin{theorem}
{(Theorem \ref{theorem:trade-off}, modified)}
Given the input distribution defined as above and a classifier $f: \R^{d+1} \rightarrow \{-1, 1\}$, suppose $\prob{f(X + \mathcal{A}(X)) = Y} \geq 1 - \eta$, for all adversarial attacks $\mathcal{A}$ of $\ell_{\infty}$ norm bounded by $4/\sqrt{d}$. Then 
{\small
\[
\left(1 - \frac{m}{d}\right)~ (1-\eta) \leq \prob{f(r(X)) = Y} \\
\leq 1 - \frac{m}{d}~ \left(1 - \frac{\eta~ p}{1-p}\right).
\]
}
Note that as $p \rightarrow 1/2$ and $\eta \rightarrow 0$, both the bounds approach $1 - m/d$, which decreases as $m$ increases. For $m=d/2$, this accuracy is as bad as that of a random classifier.
\end{theorem}
\begin{proof}
Let $\mathcal{A}(x)$ denote an adversarial perturbation for $(x, y)$ given by $(\mathcal{A}(x))_{0} = 0$, and $(\mathcal{A}(x))_{t} = -4c_{t}y/\sqrt{d}$, for $1 \leq t \leq d$. Note that $\norm{\mathcal{A}(x)}_{\infty} = 4/\sqrt{d}$, for all $x \in \mathcal{X}$. Let $G_{y}$ denotes $(X_{1}, \dotsc, X_{d}) |~ Y=y$, the last $d$ coordinates conditioned on $Y=y$. We observe that $X + A(X) |~ Y=y$ has its last $d$ coordinates distributed as $G_{-y}$, i.e., the adversarial attack that takes $X$ to $X + \mathcal{A}(X)$ turns $G_{y}$ into $G_{-y}$.
{\small
\begin{equation*}
\begin{split}
\prob{f(X + \mathcal{A}(X)) \neq Y} \\
& \hspace{-1.5in} = \frac{1}{2} \sum_{y \in \{-1, 1\}} \prob{f((X_{0}, G_{-y})) = -y} \\
& \hspace{-1.5in} = \frac{1}{2} \big \{\sum_{y \in \{-1, 1\}} p~ \prob{f((y, G_{-y})) = -y} \\
& \hspace{-0.5in} + (1-p)~ \prob{f((-y, G_{-y})) = -y} \big\} \\
& \hspace{-1.5in} \geq \frac{1-p}{2p} \big \{\sum_{y \in \{-1, 1\}} (1-p)~ \prob{f((y, G_{-y})) = -y} \\
& \hspace{-0.5in} + p~ \prob{f((-y, G_{-y})) = -y} \big \}, \qquad \text{using $p \geq 1/2$} \\
& \hspace{-1.5in} = \frac{1-p}{2p} \big \{\sum_{y \in \{-1, 1\}} (1-p)~ \prob{f((-y, G_{y})) = y} \\
& \hspace{-0.5in} + p~ \prob{f((y, G_{y})) = y} \big \} \\
& \hspace{-1.5in} = \frac{1-p}{p}~ \prob{f(X) = Y},
\end{split}
\end{equation*}
}
From the above statement, we get that if the adversarial accuracy $\prob{f(X + \mathcal{A}(X)) = Y} \geq 1 - \eta$ then the standard accuracy $\prob{f(X) = Y} \leq \eta~ p/(1-p)$. Using this we get:
{\small
\begin{align*} 
& \prob{f(r(X)) = Y} \\
& = \frac{m}{d} \sum_{j=1}^{d/m} \prob{f(r_{j}(X) = Y} \\
& = \frac{m}{d} \prob{f(X) = Y} + \frac{m}{d} \sum_{j=1}^{d/m - 1} \prob{f(r_{j}(X) = Y} \\
& \leq \frac{m}{d} \cdot \frac{\eta~ p}{1-p} + \frac{m}{d} \cdot \left(\frac{d}{m} - 1\right) \\
& = 1 - \frac{m}{d}~ \left(1 - \frac{\eta~ p}{1-p}\right).
\end{align*}
}
Two important things to note are as follows. First, the above upper bound on the accuracy after random transformation $r$ decreases as $m$ increases. Note that larger rotations mean larger $m$, e.g., a random rotation from $\{0^{\circ}, 90^{\circ}, 180^{\circ}, 270^{\circ}\}$ can be modeled by $m=d/4$. Second, for $m=d/2$ and $p \rightarrow 1/2$, this bound tends to $1/2 + \eta/2$. When $\eta$ is small, this is only slightly better than a random binary classifier.

Observe that the transformation $r_{j}(x)$ can also be achieved as an adversarial perturbation $x + \mathcal{A}_{j}(x)$, where $(\mathcal{A}_{j}(x))_{0} = 0$ and $(\mathcal{A}_{j}(x))_{t} = (r_{j}(x))_{t} - x_{t} \in \{-4y/\sqrt{d}, 0, 4y/\sqrt{d}\}$, for $1 \leq t \leq d$. Using this observation, we get
{\small
\begin{align*} 
\prob{f(r(X)) = Y} & = \frac{m}{d} \sum_{j=1}^{d/m} \prob{f(r_{j}(X) = Y} \\
& \geq \frac{m}{d} \sum_{j=1}^{d/m - 1} \prob{f(X + \mathcal{A}_{j}(X) = Y} \\
& \geq \left(1 - \frac{m}{d}\right)~ (1-\eta).
\end{align*}
}
Note that for $m = d/2$, this bound becomes $(1-\eta)/2$. When $\eta$ is small, this is only slightly worse than a random binary classifier. In other words, for $m = d/2$, high adversarial robustness implies that spatial robustness must be as bad as that of a random classifier.
\end{proof}

\section{Choice of Evaluation Metrics: Fooling Rate and Invariance vs Adversarial and Spatial Accuracy}
\begin{figure*}[!h]
\begin{center}
\includegraphics[width=0.24\linewidth]{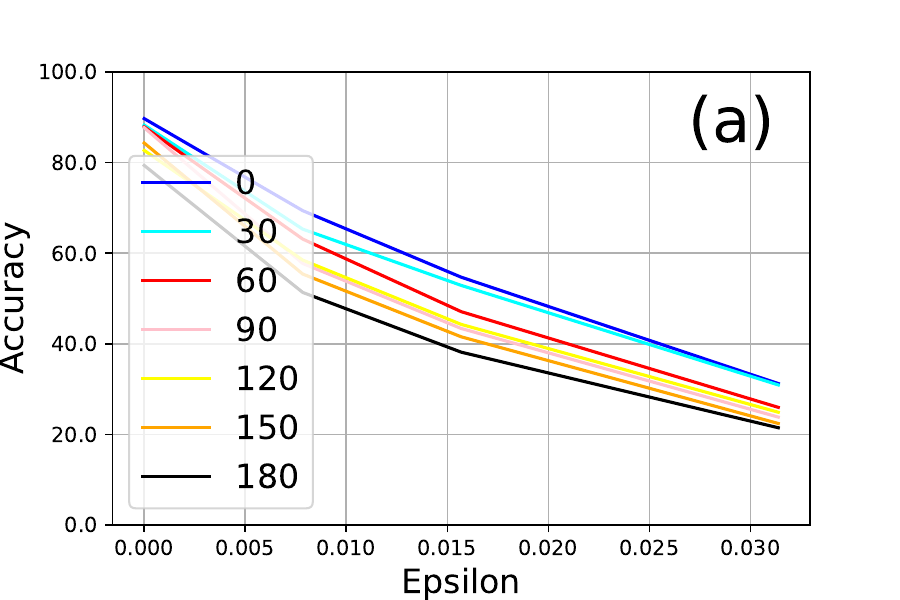}
\includegraphics[width=0.24\linewidth]{plots/stdcnn_vgg16_cifar10_fixedtrain_pgd_foolingrate_notitle.pdf}
\rulesep
\includegraphics[width=0.24\linewidth]{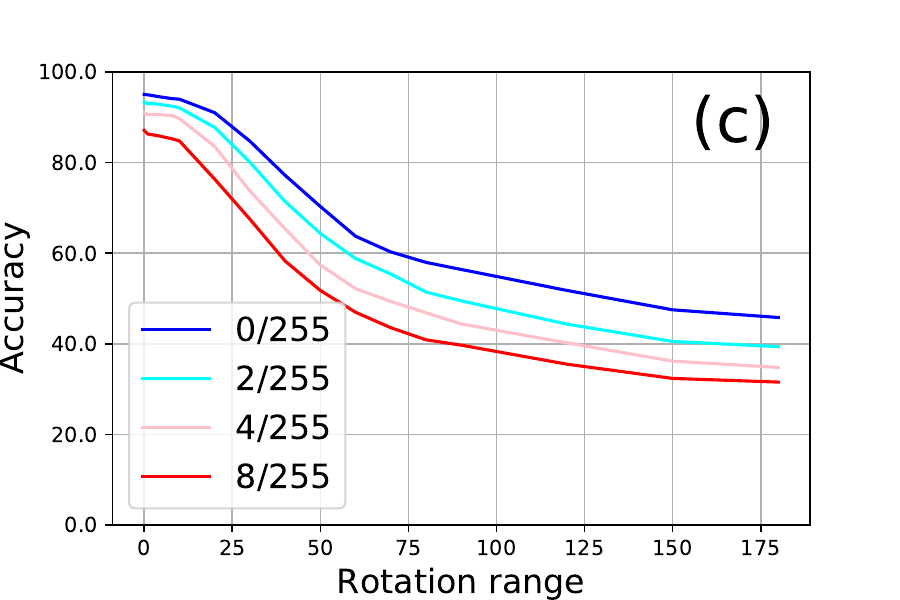}
\includegraphics[width=0.24\linewidth]{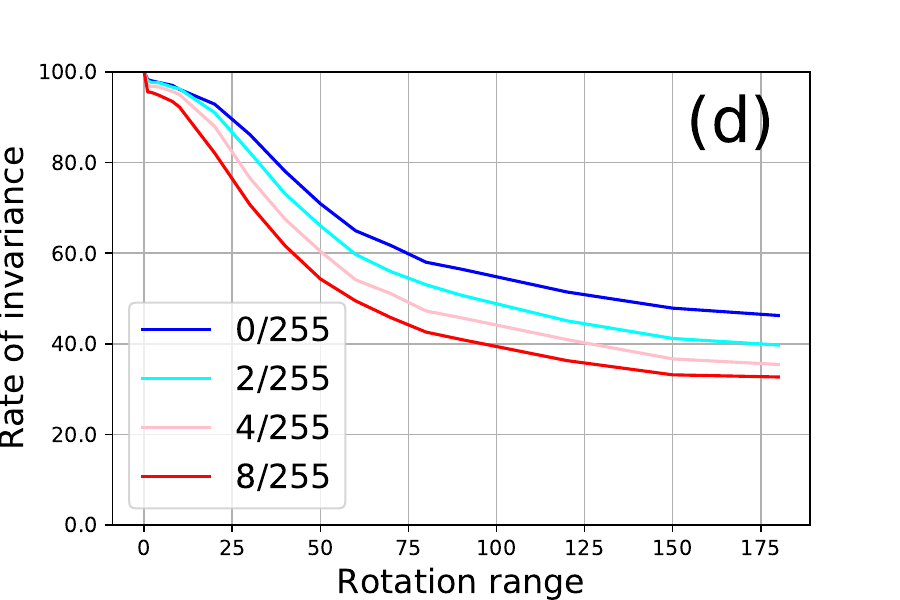}
\end{center}
\vspace{-10pt}
\caption{(a-b) Plots of accuracy and (1-fooling rate) for  StdCNN/VGG16 trained on CIFAR10; (c-d) Plots of accuracy and rate of invariance for ResNet-based model from \protect \cite{Madry18} adversarially trained on CIFAR10. Note the same starting points on (b) \& (d), which allow for better comparison.}
\label{fool-accuracy}
\end{figure*}

\vspace{-6pt}
In the main paper, our theoretical results are about the adversarial accuracy $\prob{f(X + \mathcal{A}(X)) = Y}$ and the spatial accuracy $\prob{f(r(X)) = Y}$. However, the experiments show the fooling rate $\prob{f(X + \mathcal{A}(X)) = f(X)}$ and the rate of invariance $\prob{f(r(X)) = f(X)}$, respectively. In fact, they are reasonably proxies for each other, respectively, when the standard accuracy $\prob{f(X) = Y}$ is high. In Fig \ref{fool-accuracy}, we plot the adversarial and the spatial accuracy in (a) and (c), alongside (1 - fooling rate) and rate of invariance in (b) and (d). As evident, our results and claims on the trade-off continue to hold in either case. In the main paper, we present our results using the fooling rate and the rate of invariance because that ensures the same starting point for all the curves, and provides a better visual representation to compare adversarial and spatial robustness of different models.

\section{Details of Experimental Setup} \label{suppl:setup-exp-details}
We describe in detail the settings for all our experiments. 

\uline{\textit{Datasets:}} We use the standard benchmark train-validation-test splits of all the datasets used in this work, that is publicly available. MNIST dataset consists of $70,000$ images of $28 \times 28$ size, divided into $10$ classes: $55,000$ used for training, $5,000$ for validation and $10,000$ for testing. CIFAR10 dataset consists of $60,000$ images of $32 \times 32$ size, divided into $10$ classes: $40,000$ used for training, $10,000$ for validation and $10,000$ for testing. CIFAR100 dataset consist of $60,000$ images of $32 \times 32$ size, divided into $100$ classes: $40,000$ used for training, $10,000$ for validation and $10,000$ for testing. Tiny-ImageNet dataset consists of $100,000$ images of $64 \times 64$ size, divided into $200$ classes: $80,000$ used for training, $10,000$ for validation and $10,000$ for testing. 

\uline{\textit{Spatially Robust Model Architectures:}} 
StdCNNs are known to be translation-equivariant by design, and GCNNs \cite{Cohen16} are rotation-equivariant by design through clever weight sharing \cite{Kondor18}. Equivariant models, especially GCNNs, when trained with random rotation augmentations have been observed to come very close to being truly rotation-invariant \cite{Cohen16,Cohen17,Cohen18} (or spatially robust in our context). We hence use both StdCNNs and equivalent GCNNs trained with suitable data augmentations for our studies with spatially robust architectures. In particular, 
for each StdCNN we use, the corresponding GCNN architecture is obtained by replacing the layer operations with equivalent GCNN operations as in \cite{Cohen16}\footnote{https://github.com/adambielski/pytorch-gconv-experiments}. For the StdCNN, we use the architecture given in Table \ref{gcnn-table} for MNIST; VGG16 \cite{Simonyan14} and ResNet18 \cite{He16} for CIFAR10 and CIFAR100; and ResNet18 for the Tiny ImageNet dataset.

\uline{\textit{Adversarially Robust Model Architectures:}} 
For adversarial training, we use a LeNet-based architecture for MNIST\footnote{https://github.com/MadryLab/mnist\_challenge/} and a ResNet-based architecture for CIFAR10\footnote{https://github.com/MadryLab/cifar10\_challenge}. Both these models are exactly as given in \cite{Madry18}. For CIFAR100, we use the popularly used WideResNet-34\footnote{https://github.com/yaodongyu/TRADES/models/} architecture also used in \cite{zhang2019theoretically}. We use ResNet18 \cite{He16} for the Tiny ImageNet dataset.




\uline{\textit{Training Data Augmentation:}}
\emph{Spatial}: \begin{inparaenum}[(a)] \item \textbf{Aug - R} : Data is augmented with random rotations in the range $\pm \theta^{\circ}$ given $\theta$, along with random crops and random horizontal flips (for MNIST alone, we do not apply crop and horizontal flips); \item \textbf{Aug - T}: Data is augmented with random translations within $[-i, +i]$ range of pixels in the image, given $i$ (eg. for CIFAR10 with $i=0.1$ is $32*0.1 \approx \pm3 px$) in both horizontal and vertical directions; \item \textbf{Aug - RT}: Data is augmented with random rotations in $\pm i*180^{\circ}$ and random translations within $[-i, +i]$ range of pixels in the image (eg. for CIFAR10 with $i=0.1$ is $0.1*180^{\circ} = \pm 18^{\circ}$ rotation and $32*0.1 \approx \pm3 px$ translation), here no cropping and no horizontal flip is used. Random transformation is picked uniformly at random in the given transformation range, e.g., given $\theta$, rotations are picked uniformly at random from $[-\theta^{\circ}, +\theta^{\circ}]$ for augmentation. We use nearest neighbour interpolation and black filling to obtain the transformed image. \end{inparaenum} \emph{Adversarial} : \textbf{Adv - PGD}: Adversarial training using PGD-perturbed adversarial samples using an $\epsilon$-budget of given $\epsilon$. Our experiments with PGD use a random start, 40 iterations, and step size $0.01$ on MNIST, and a random start, 10 iterations, and step size $2/255$ on CIFAR10, CIFAR100 and Tiny ImageNet. Our results are best understood by noting the augmentation method mentioned in the figure caption. For example, in Fig \ref{cifar10-stdcnn-gcnn-vgg16-full}(a), the augmentation scheme used is \textbf{Aug-R}. The red line (annotated as 60 in the legend) corresponds to the model trained with random rotation augmentations in the range $\pm60^{\circ}$.

\uline{\textit{Hyperparameter Settings for CuSP algorithm:}}
The empirical evaluations of the CuSP algorithm are based on settings used in the TRADES method. For CIFAR10, we train the model for 120 epochs with the learning rate schedule 75-90-100 where the learning rate starting with 0.1 decays by a factor of 0.1 successively at the 75-th, 90-th and 100-th epochs. TRADES is used with $\frac{1}{\lambda}=5.0$. PGD-based perturbation is used with step size = $\frac{2}{255}$ or $0.007$ and number of steps = $10$ with the appropriate $\epsilon$ budget.

\vspace{-3pt}
\uline{\textit{Hardware Configuration:}}
We used a computing server with 4 Nvidia GeForce GTX 1080i GPUs to run all experiments in the paper.

\vspace{-3pt}
\uline{\textit{Tool License:}}
AdverTorch \cite{ding2019advertorch} provides the tool under GNU/ LGPL. TRADES and code for \cite{Madry18} are under MIT License.

\section{Experiments on MNIST} \label{suppl:exp-results-mnist}
\begin{figure*}[!h]
\begin{center}
\includegraphics[width=0.24\linewidth]{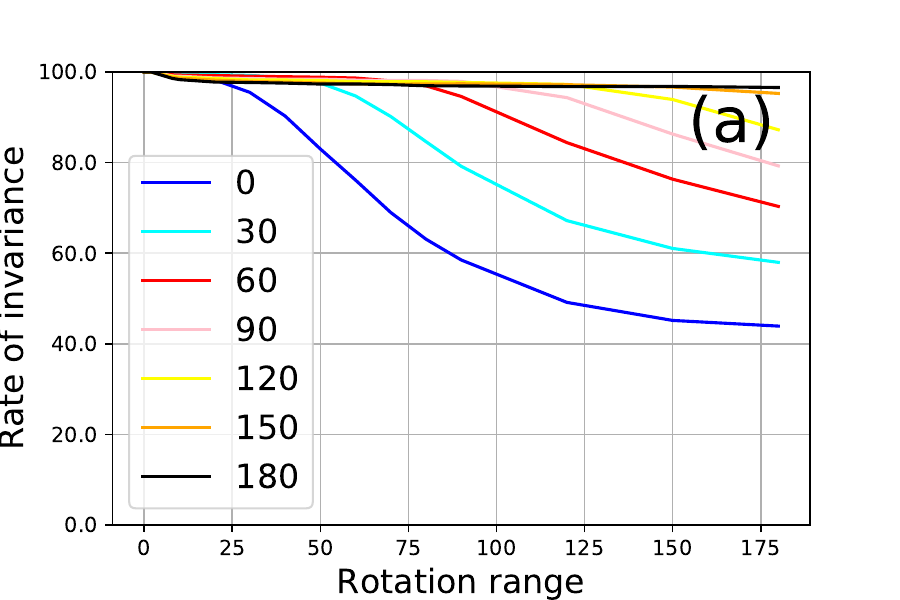} 
\includegraphics[width=0.24\linewidth]{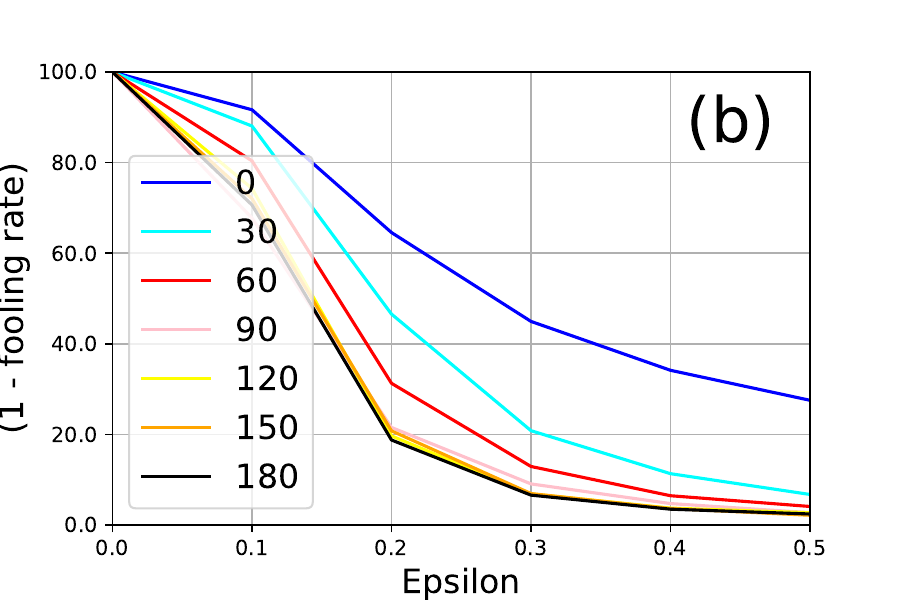}
\rulesep
\includegraphics[width=0.24\linewidth]{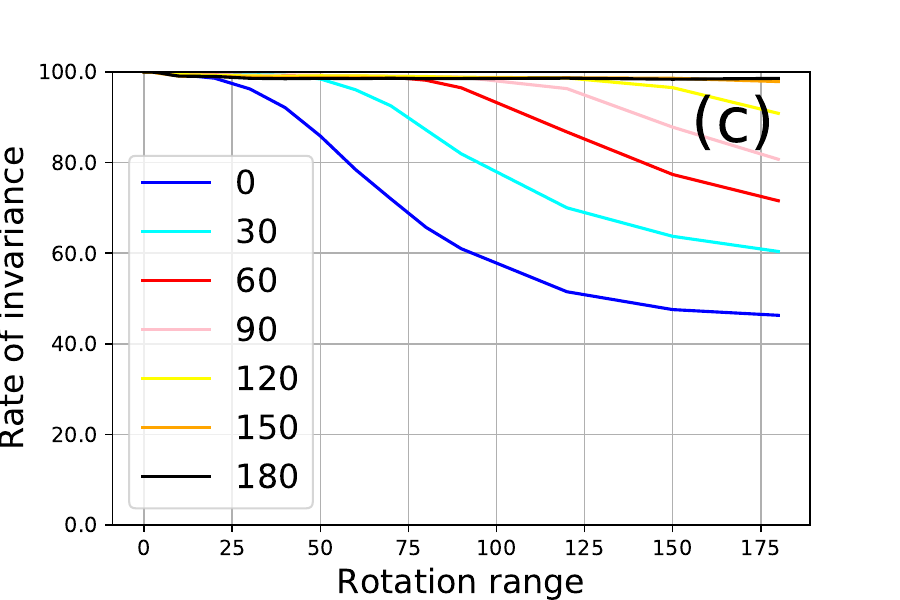} 
\includegraphics[width=0.24\linewidth]{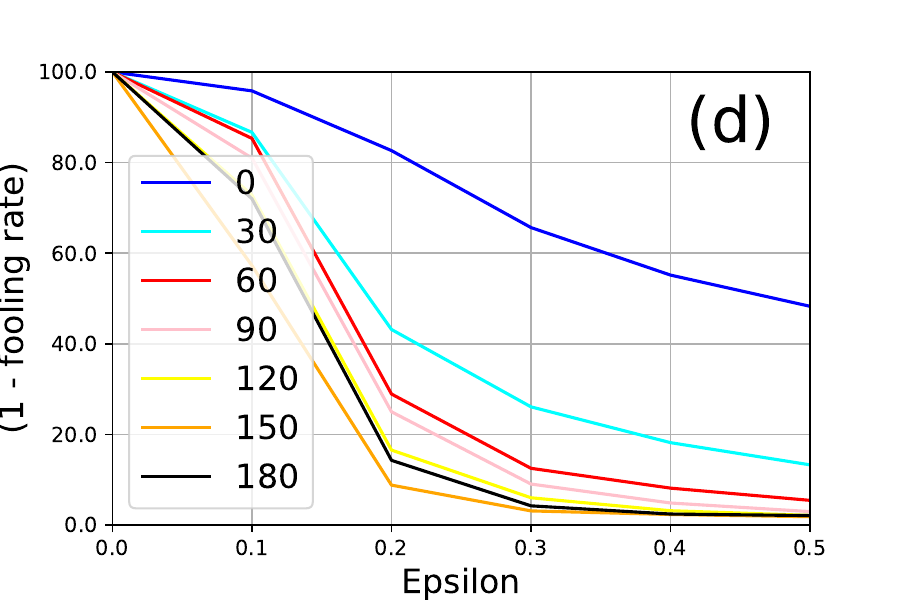}\\
\includegraphics[width=0.24\linewidth]{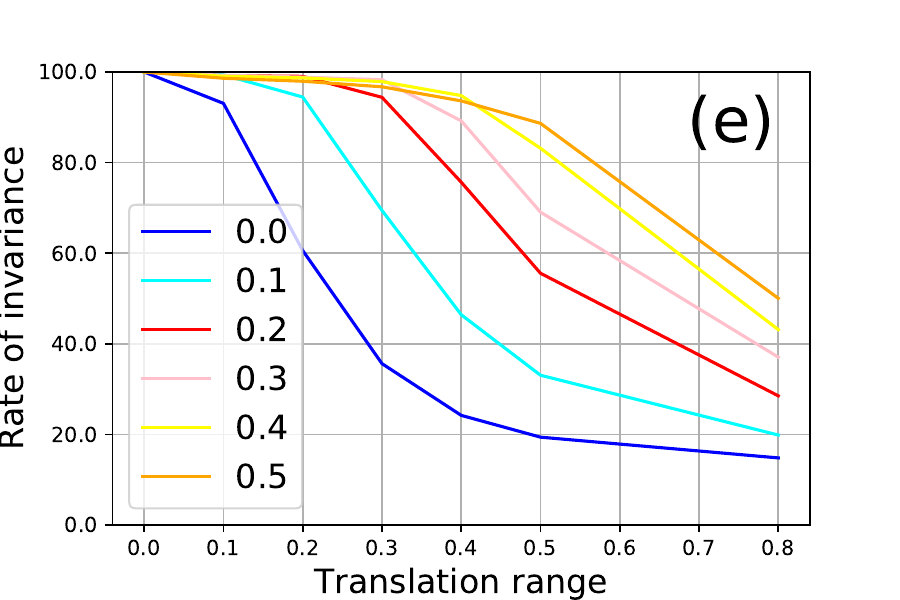}
\includegraphics[width=0.24\linewidth]{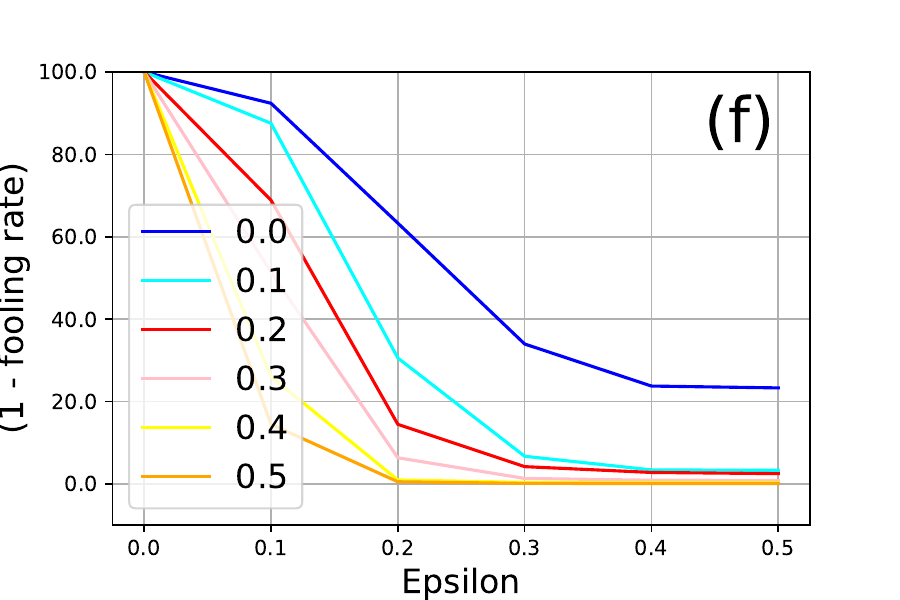}
\rulesep
\includegraphics[width=0.24\linewidth]{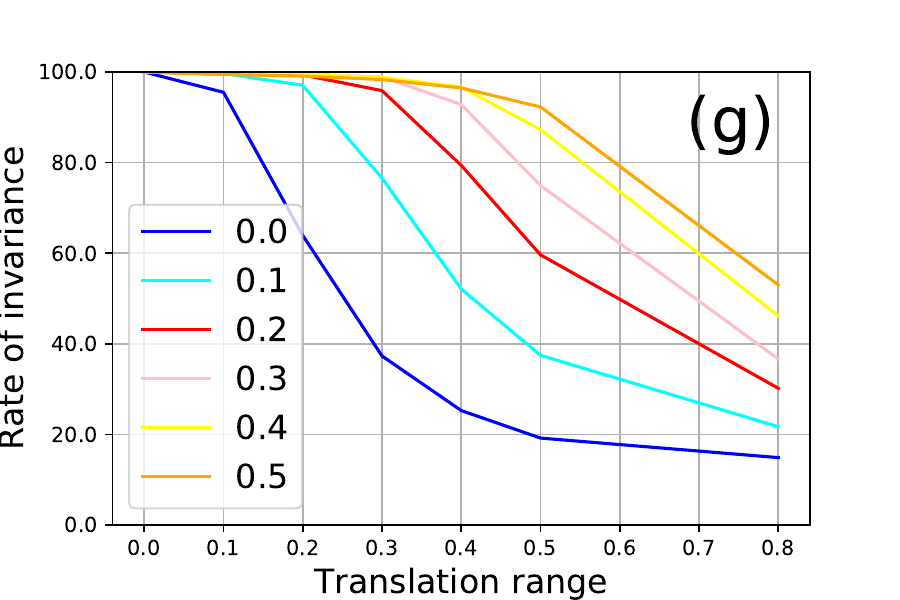}
\includegraphics[width=0.24\linewidth]{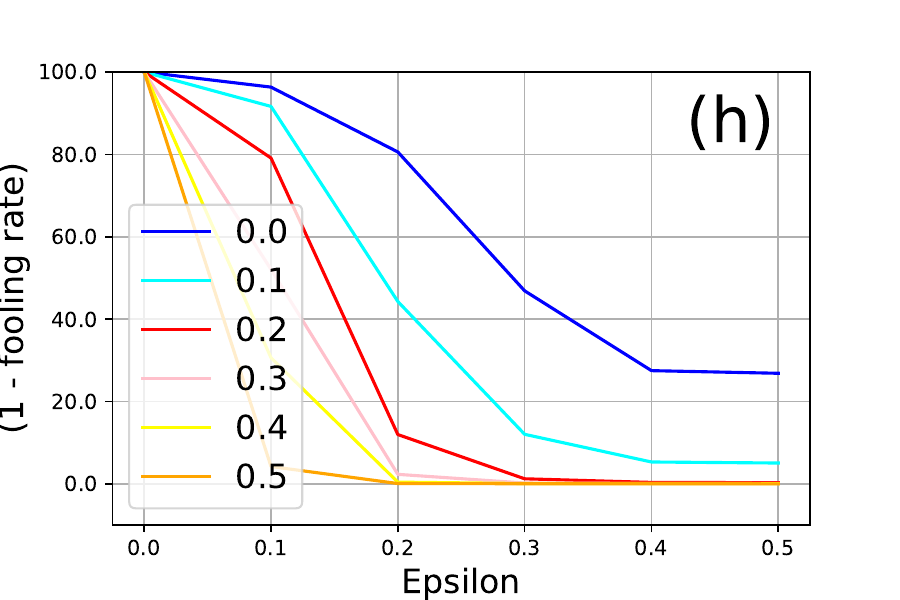}\\
\includegraphics[width=0.24\linewidth]{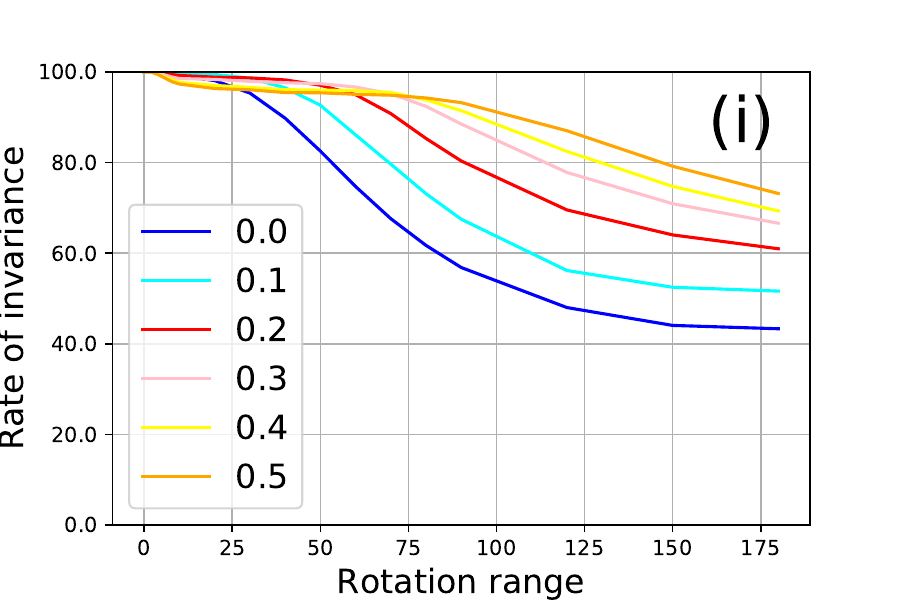}
\includegraphics[width=0.24\linewidth]{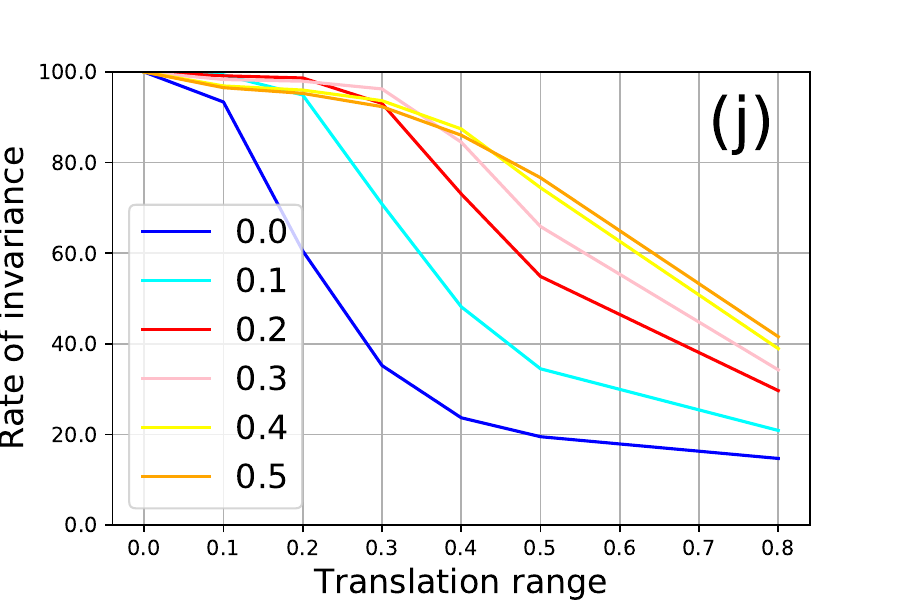}
\includegraphics[width=0.24\linewidth]{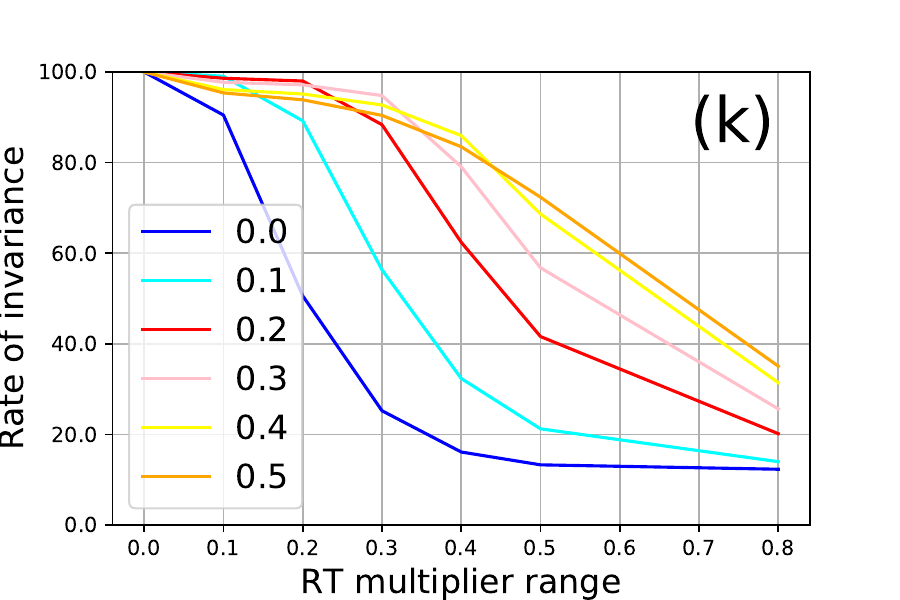}
\includegraphics[width=0.24\linewidth]{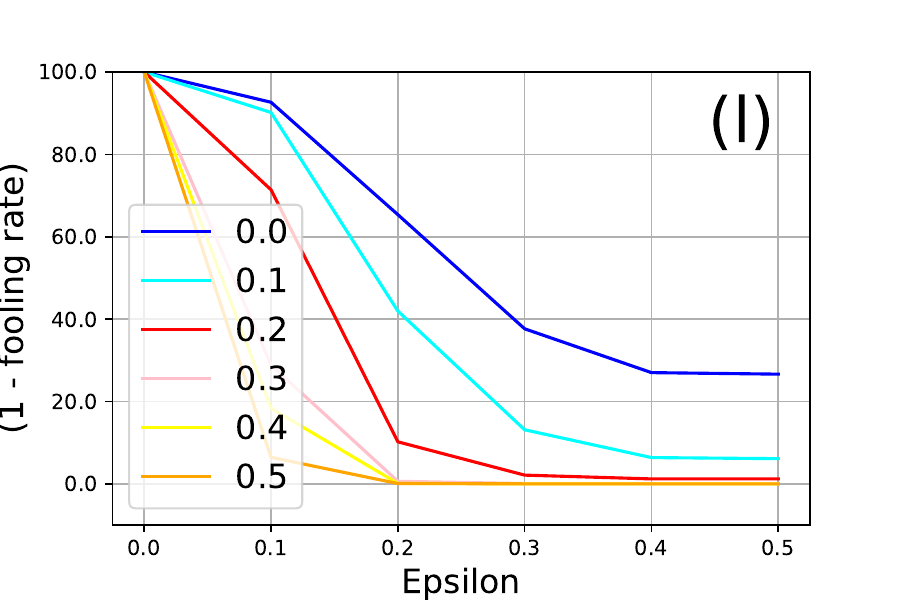}\\
\includegraphics[width=0.24\linewidth]{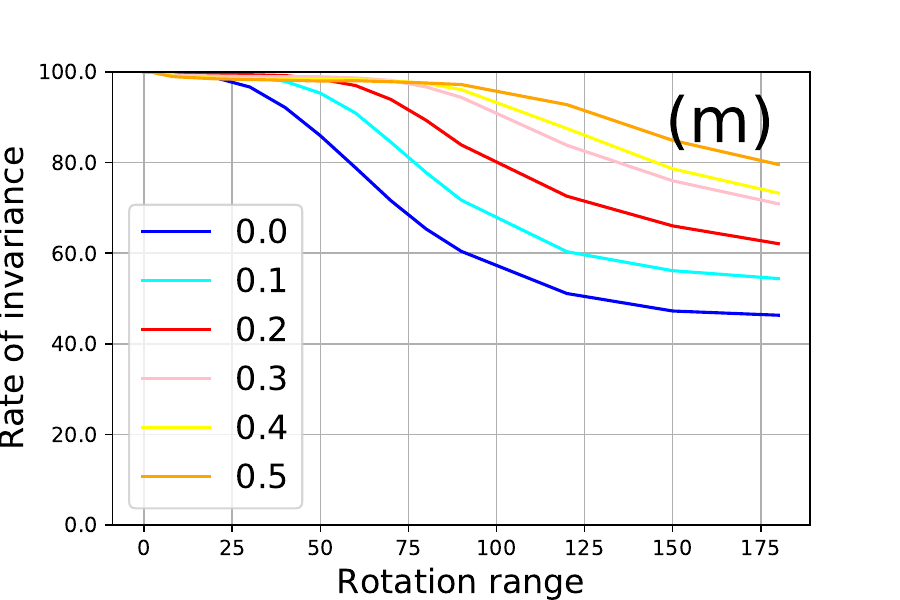}
\includegraphics[width=0.24\linewidth]{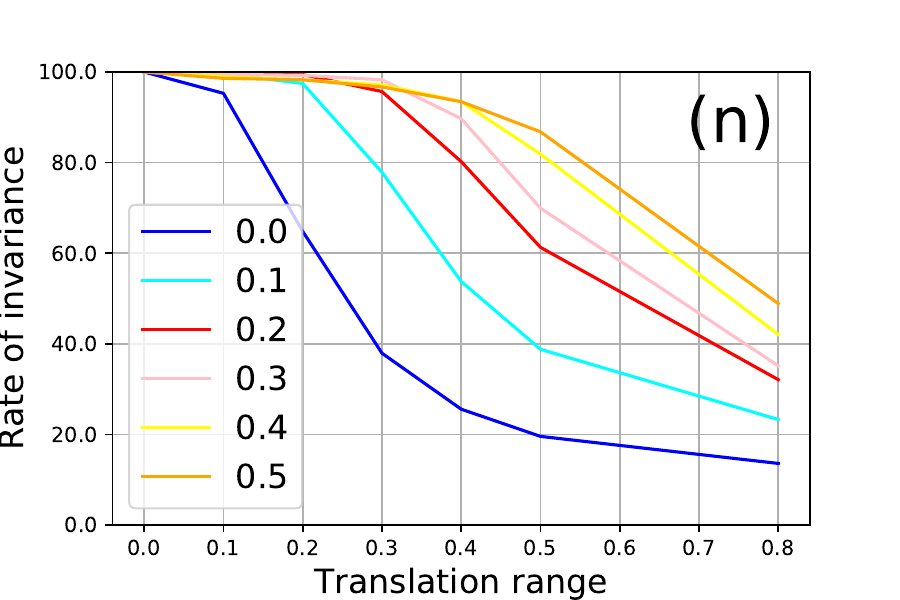}
\includegraphics[width=0.24\linewidth]{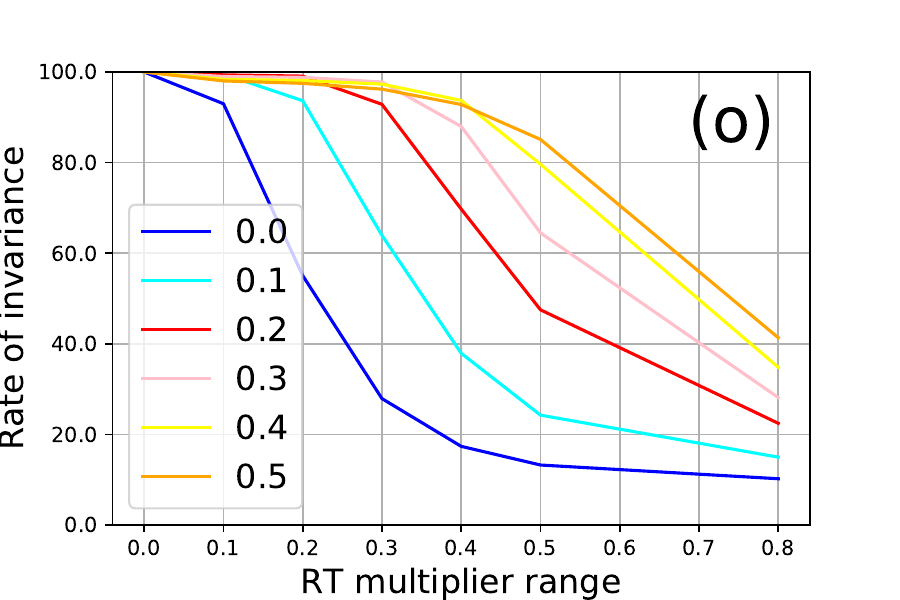}
\includegraphics[width=0.24\linewidth]{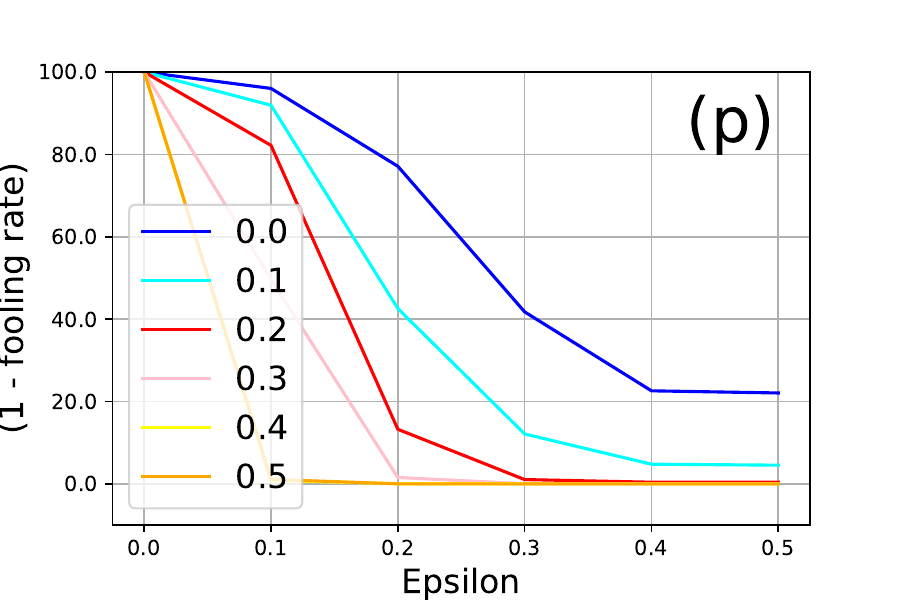}
\end{center}
\vspace{-4pt}
\caption{On MNIST, (a-b) \textbf{Aug - R} StdCNN, (c-d) \textbf{Aug - R} GCNN, (e-f) \textbf{Aug - T} StdCNN, (g-h) \textbf{Aug - T} GCNN, (i-l) \textbf{Aug - RT} StdCNN, (m-p) \textbf{Aug - RT} GCNN, invariance profiles of StdCNN/GCNN models and corresponding robustness profiles.}
\vspace{-8pt}
\label{mnist-stdcnn-gcnn-full}
\end{figure*}

\vspace{-6pt}
Table \ref{gcnn-table} shows the details of the model architecture used for our experiments on the MNIST dataset. Fig \ref{mnist-stdcnn-gcnn-full} contains the invariance and robustness profiles of StdCNN and GCNN models on MNIST. We observe the same trends noted in the main paper in these results too.
\begin{table}[!h] 
\caption{Architectures used for MNIST experiments\vspace{-12pt}} 
\footnotesize
\begin{center}
\begin{tabular}{ll}
\multicolumn{1}{c}{\bf Standard CNN}  &\multicolumn{1}{c}{\bf GCNN}
\\ \hline
Conv(10,3,3) + Relu & P4ConvZ2(10,3,3) + Relu \\
Conv(10,3,3) + Relu & P4ConvP4(10,3,3) + Relu \\
Max Pooling(2,2)  & Group Spatial Max Pooling(2,2)  \\
Conv(20,3,3) + Relu & P4ConvP4(20,3,3) + Relu \\
Conv(20,3,3) + Relu & P4ConvP4(20,3,3) + Relu \\
Max Pooling(2,2)  & Group Spatial Max Pooling(2,2) \\
FC(50) + Relu & FC(50) + Relu \\
Dropout(0.5) & Dropout(0.5) \\
FC(10) + Softmax & FC(10) + Softmax \\
\end{tabular}
\vspace{-8pt}
\end{center}
\label{gcnn-table}
\end{table}
\vspace{-10pt}


\vspace{-12pt}
\section{Additional Results on CIFAR10} \label{suppl:exp-results-cifar10}
\begin{figure*}[!h]
\begin{center}
\includegraphics[width=0.24\linewidth]{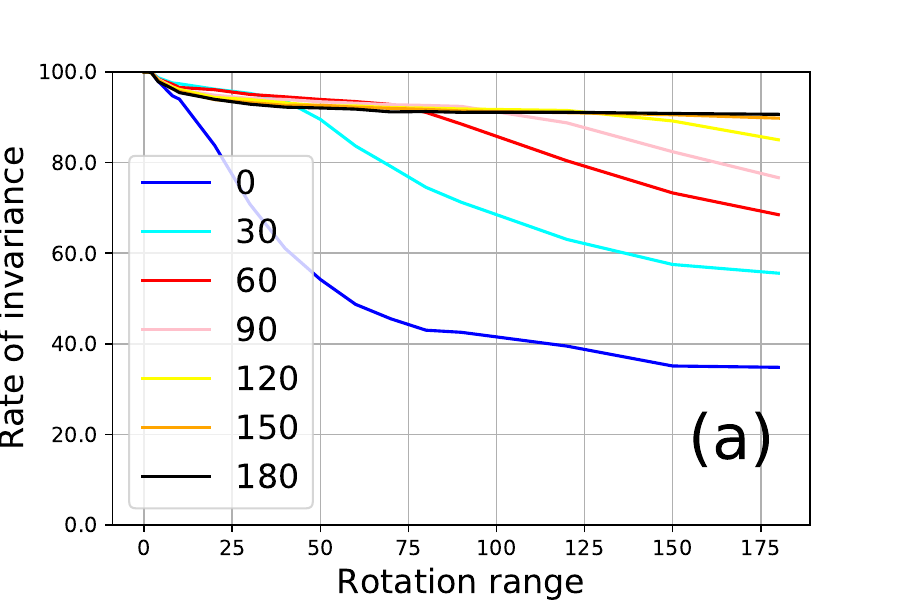}
\includegraphics[width=0.24\linewidth]{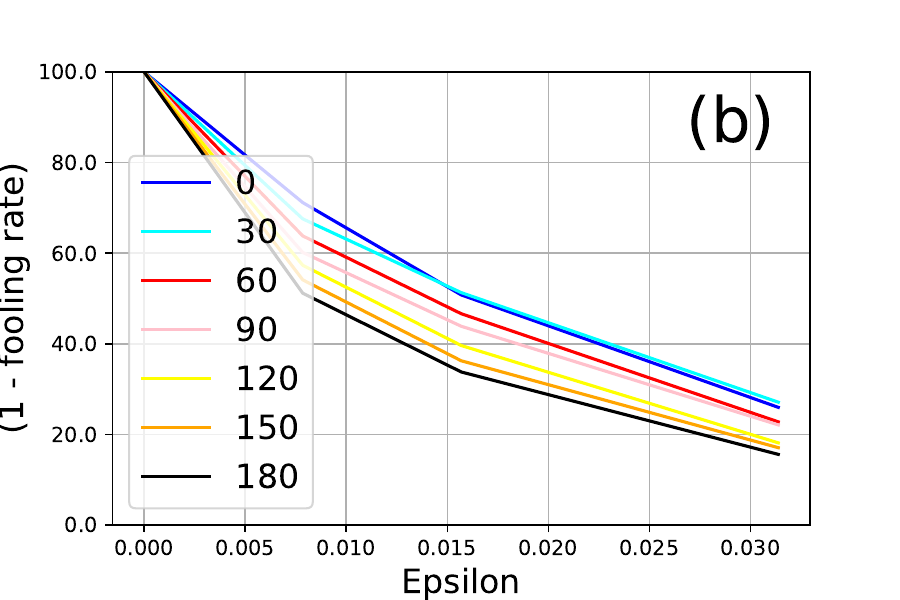}
\rulesep
\includegraphics[width=0.24\linewidth]{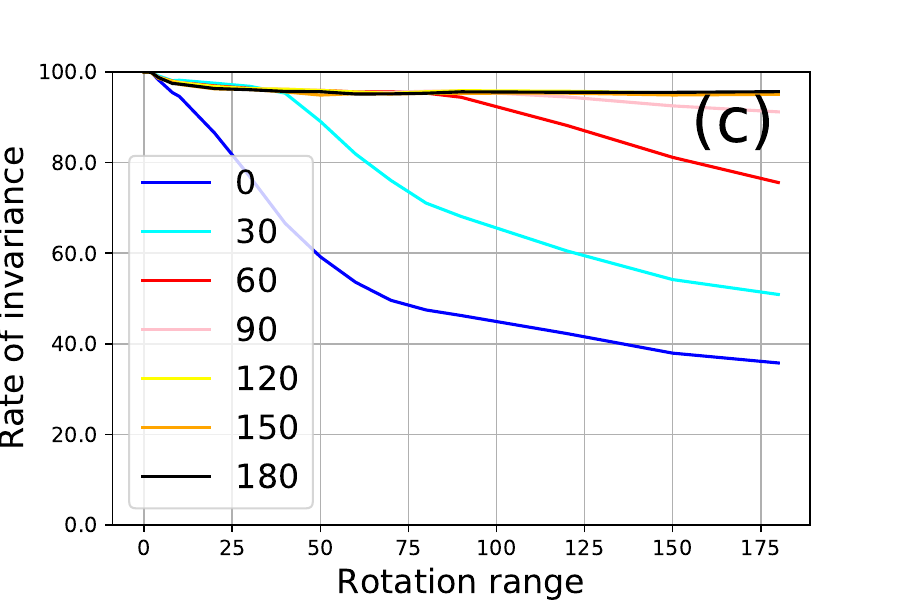}
\includegraphics[width=0.24\linewidth]{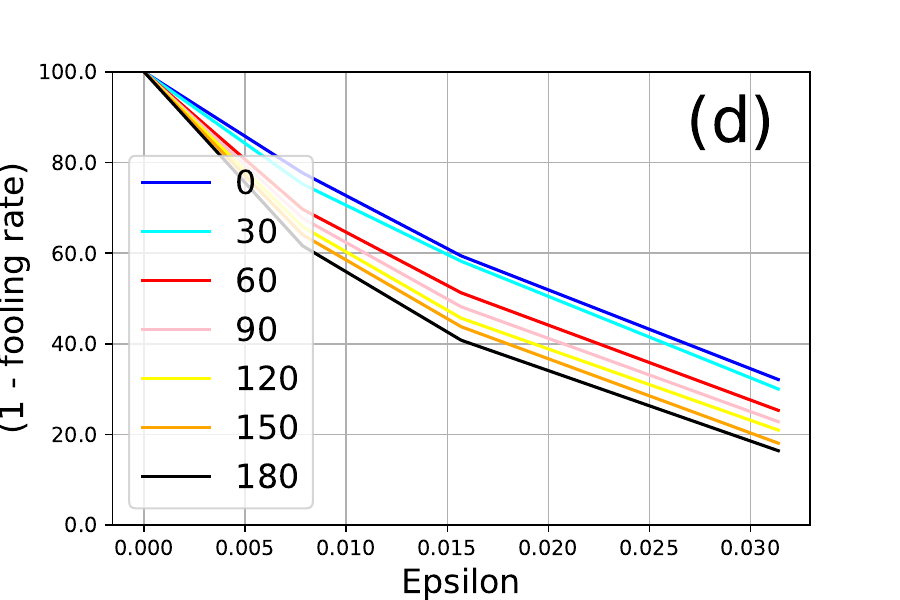}\\
\includegraphics[width=0.24\linewidth]{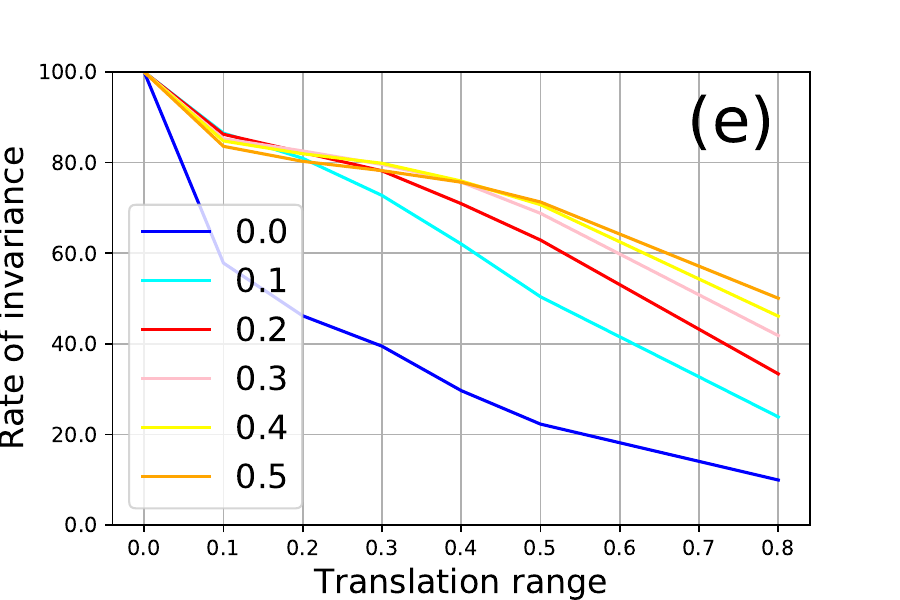}
\includegraphics[width=0.24\linewidth]{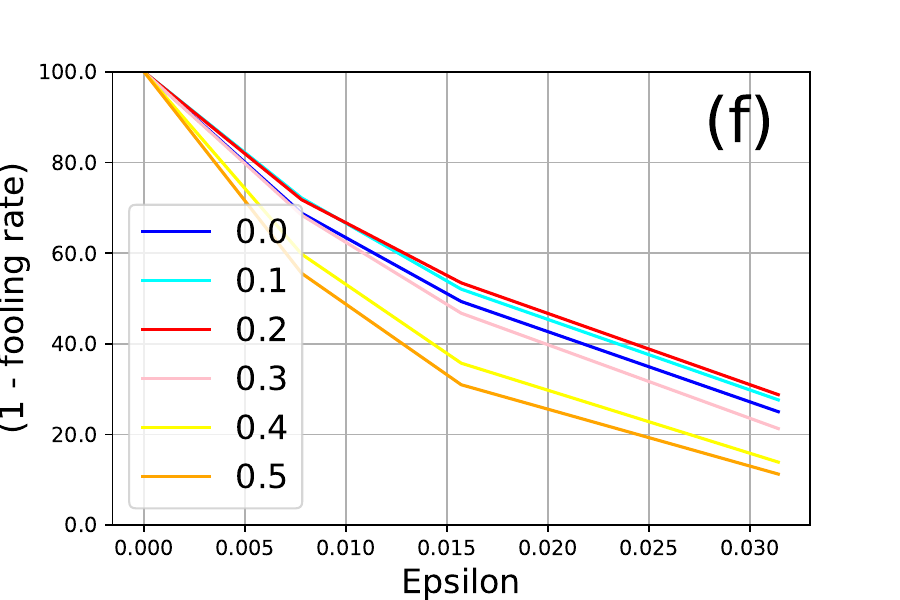}
\rulesep
\includegraphics[width=0.24\linewidth]{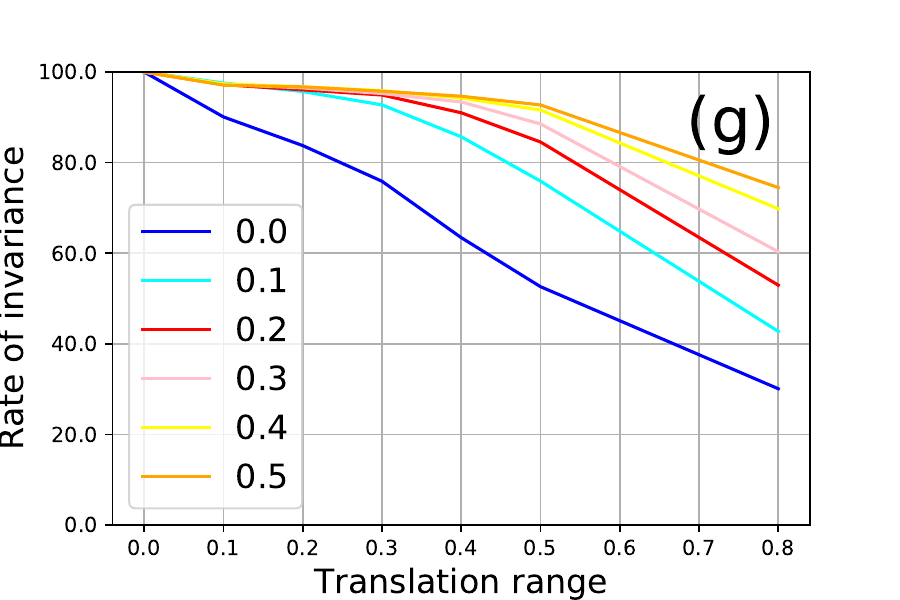}
\includegraphics[width=0.24\linewidth]{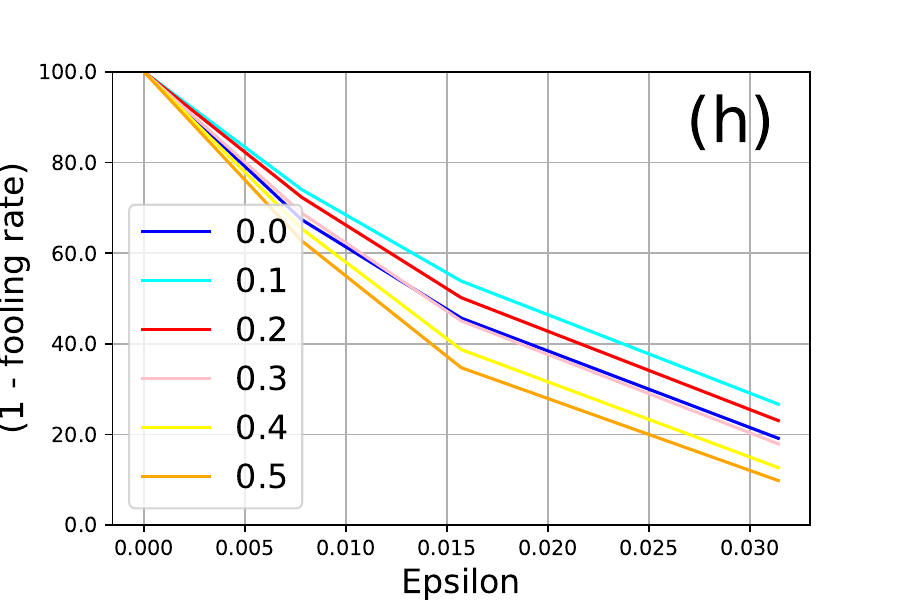}\\
\includegraphics[width=0.24\linewidth]{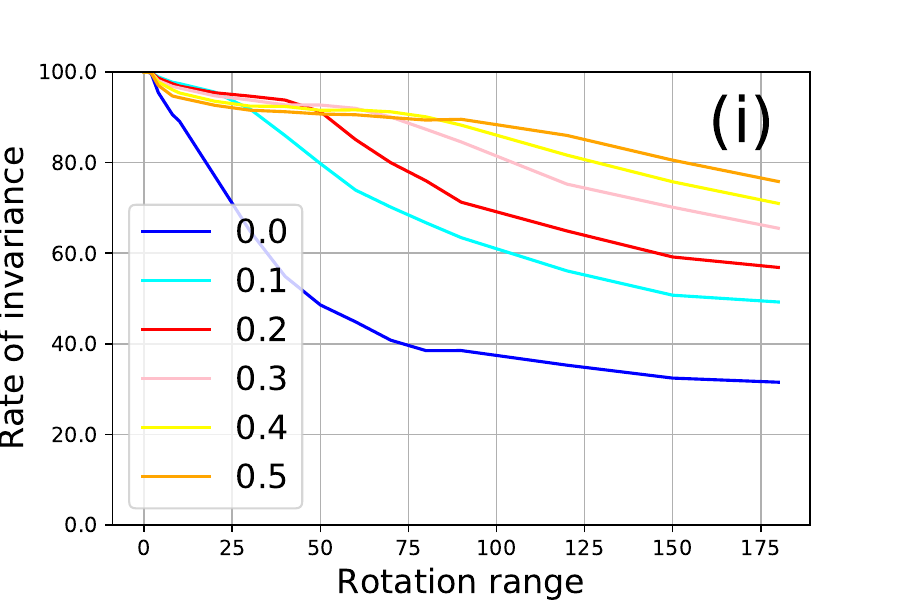}
\includegraphics[width=0.24\linewidth]{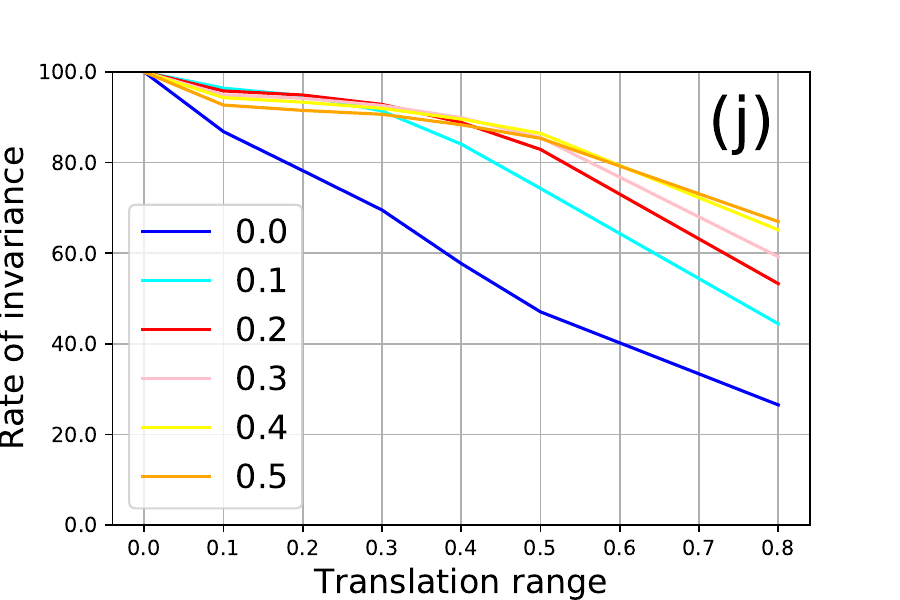}
\includegraphics[width=0.24\linewidth]{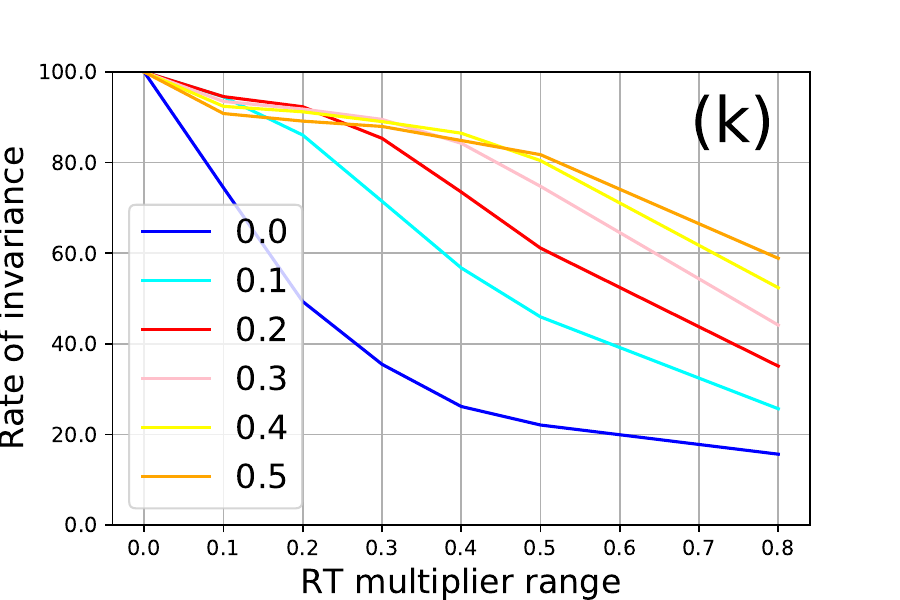}
\includegraphics[width=0.24\linewidth]{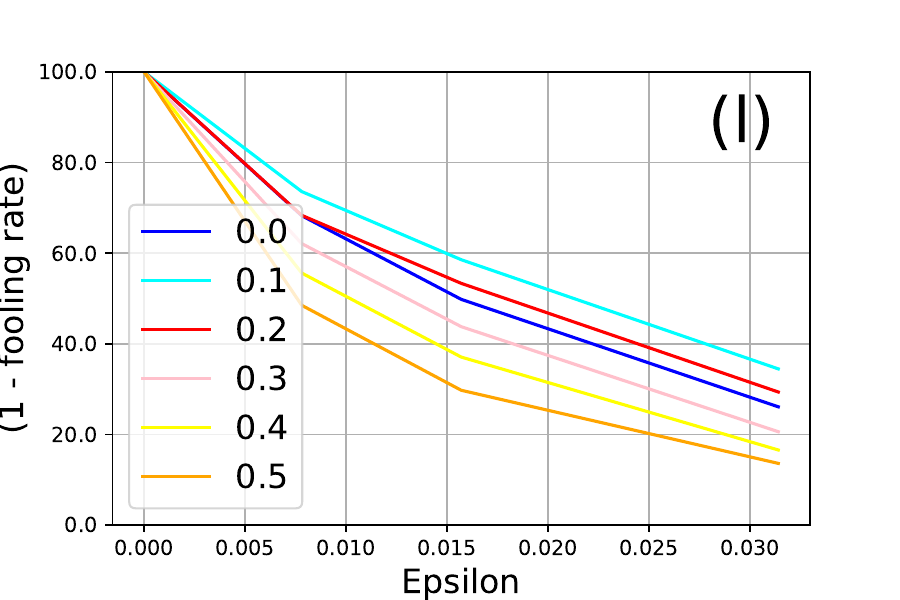}\\
\includegraphics[width=0.24\linewidth]{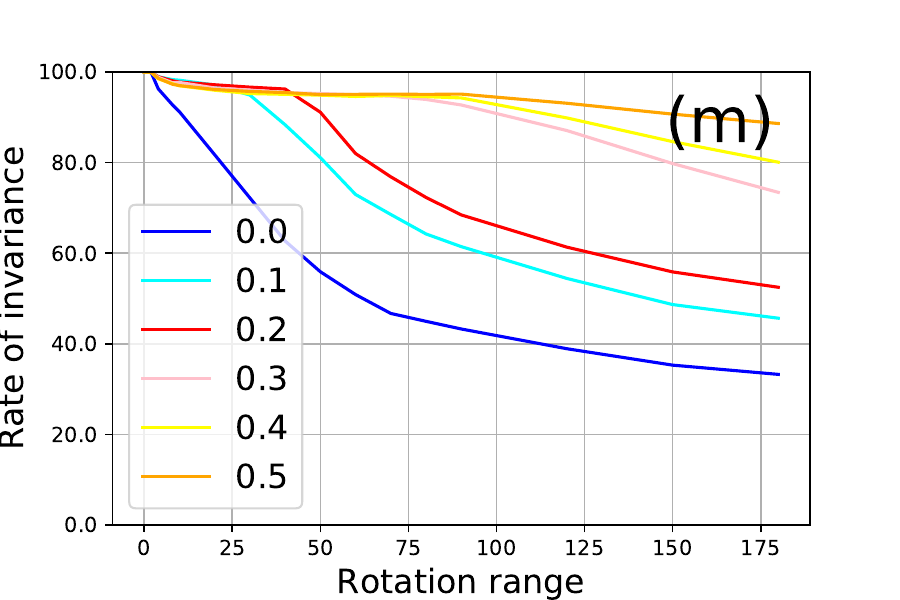}
\includegraphics[width=0.24\linewidth]{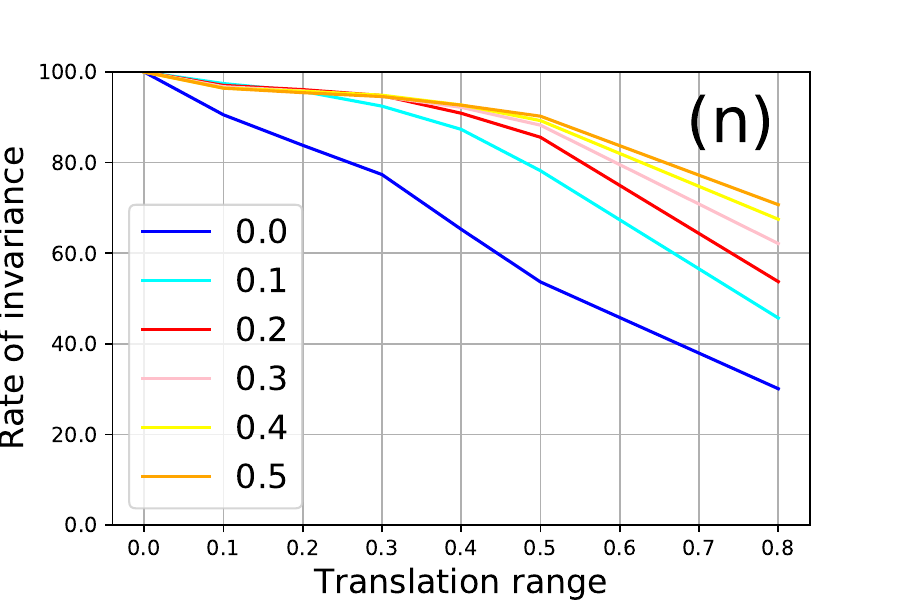}
\includegraphics[width=0.24\linewidth]{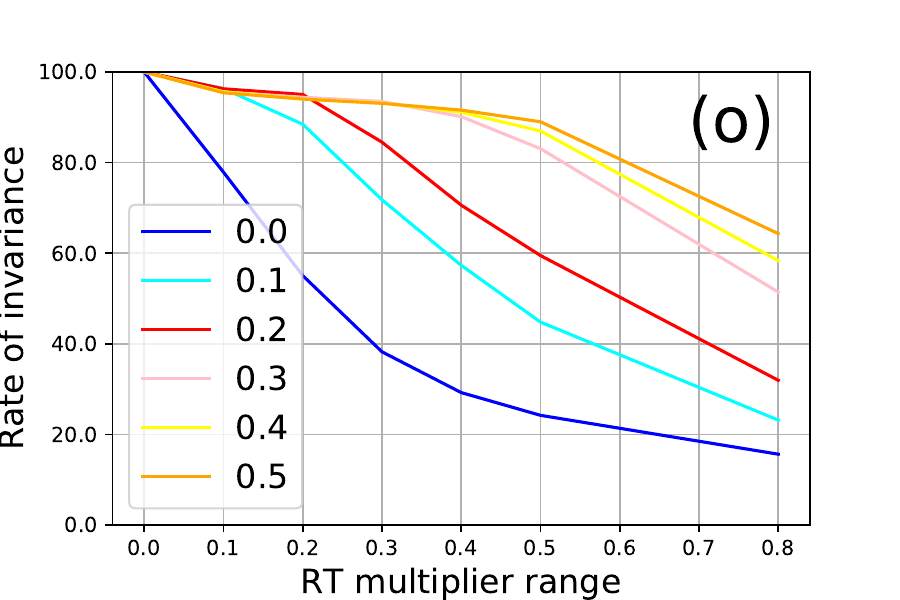}
\includegraphics[width=0.24\linewidth]{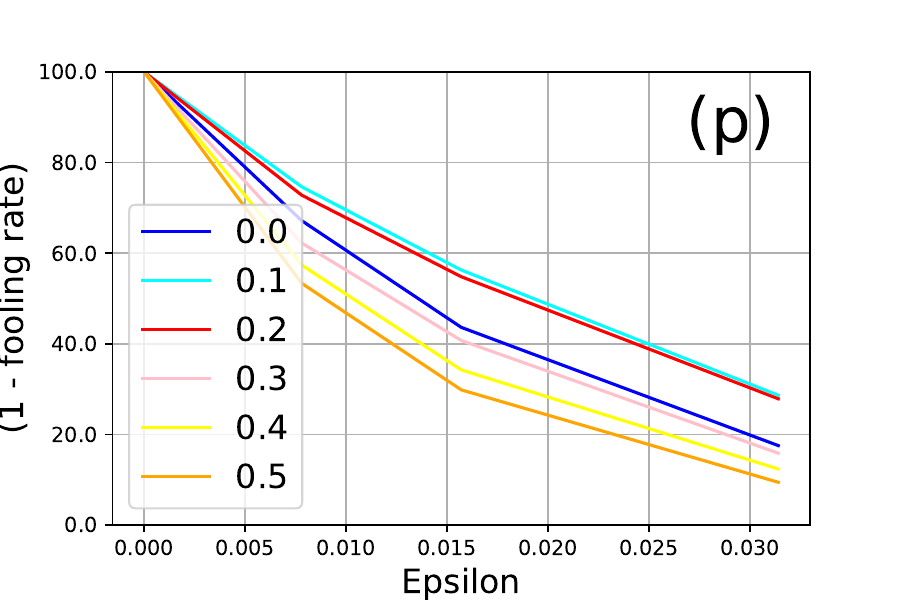}
\end{center}
\vspace{-10pt}
\caption{\textit{(Best viewed in color, zoomed in)} On CIFAR10, For ResNet18 model (a-b) \textbf{Aug - R} StdCNN, (c-d) \textbf{Aug - R} GCNN, (e-f) \textbf{Aug - T} StdCNN, (g-h) \textbf{Aug - T} GCNN, (i-l) \textbf{Aug - RT} StdCNN, (m-p) \textbf{Aug - RT}, invariance profiles of StdCNN/GCNN models and corresponding robustness profiles.}
\label{cifar10-stdcnn-gcnn-resnet18-full}
\end{figure*}

\vspace{-8pt}
In the main paper (Fig 1), we presented results for experiments carried out using VGG16 on CIFAR10. Please see Fig \ref{cifar10-stdcnn-gcnn-resnet18-full} for additional experiments on CIFAR10 using the ResNet18 model. These complement the experiments in the main paper and show that our results continue to hold across different models. 

\vspace{-7pt}
\section{Additional Results on CIFAR100} \label{suppl:exp-results-cifar100}
\begin{figure*}[!h]
\begin{center}
\includegraphics[width=0.24\linewidth]{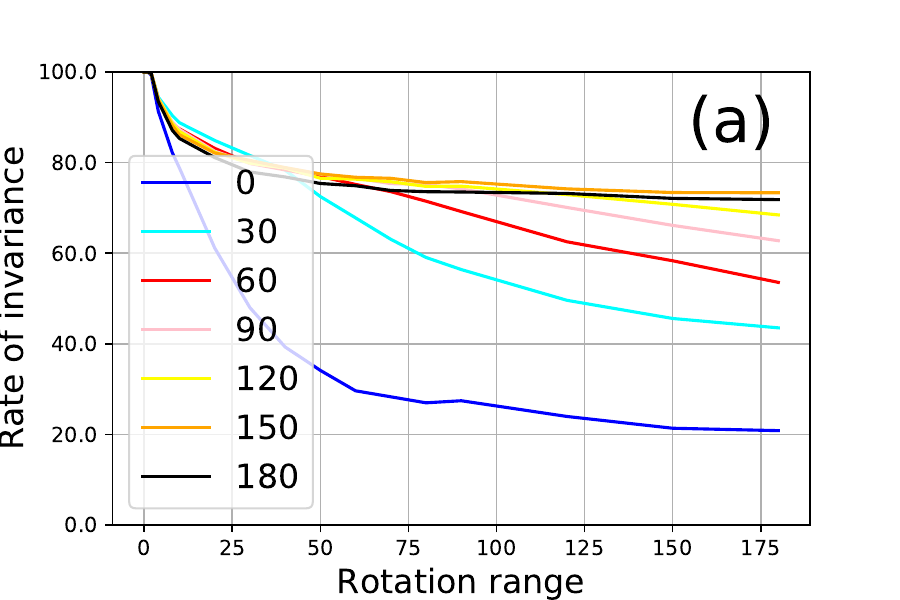}
\includegraphics[width=0.24\linewidth]{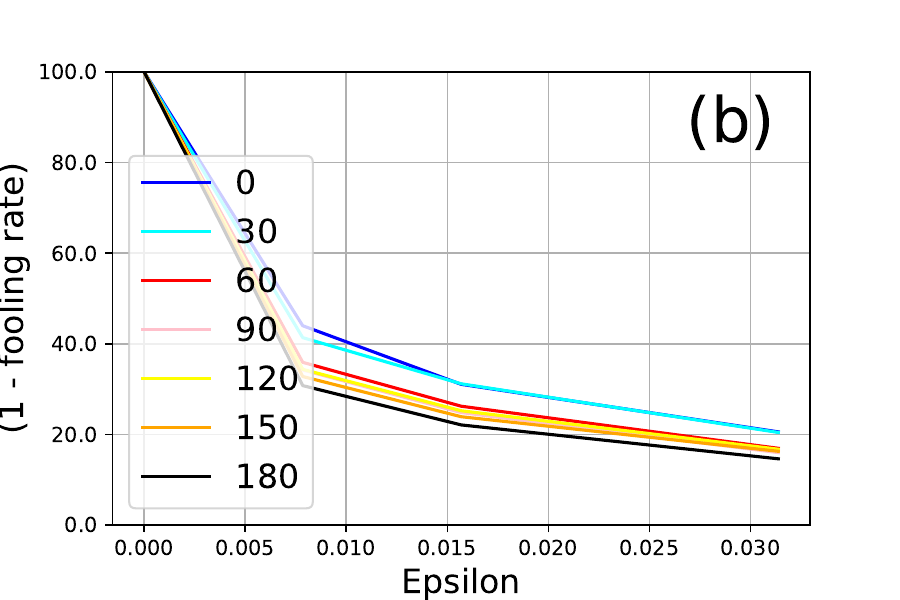}
\rulesep
\includegraphics[width=0.24\linewidth]{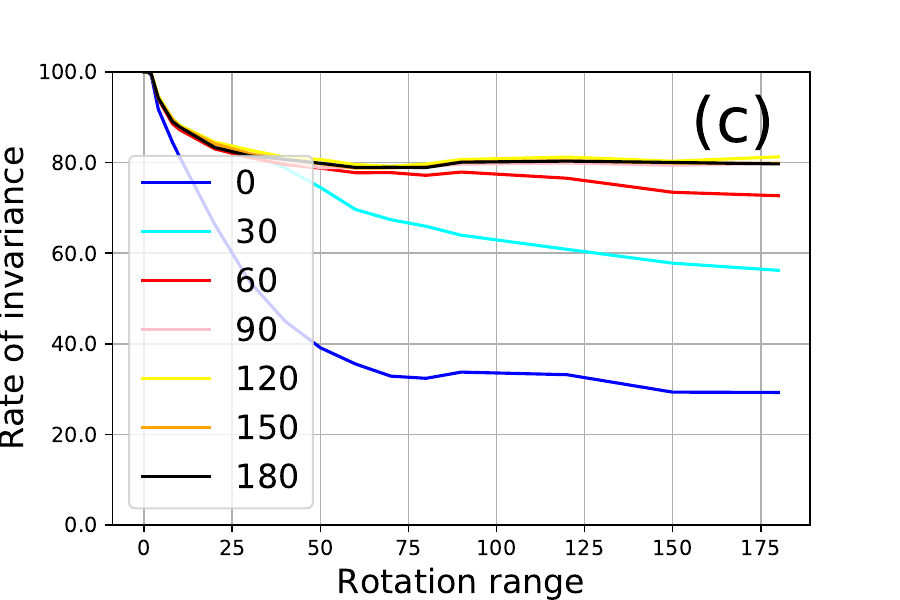}
\includegraphics[width=0.24\linewidth]{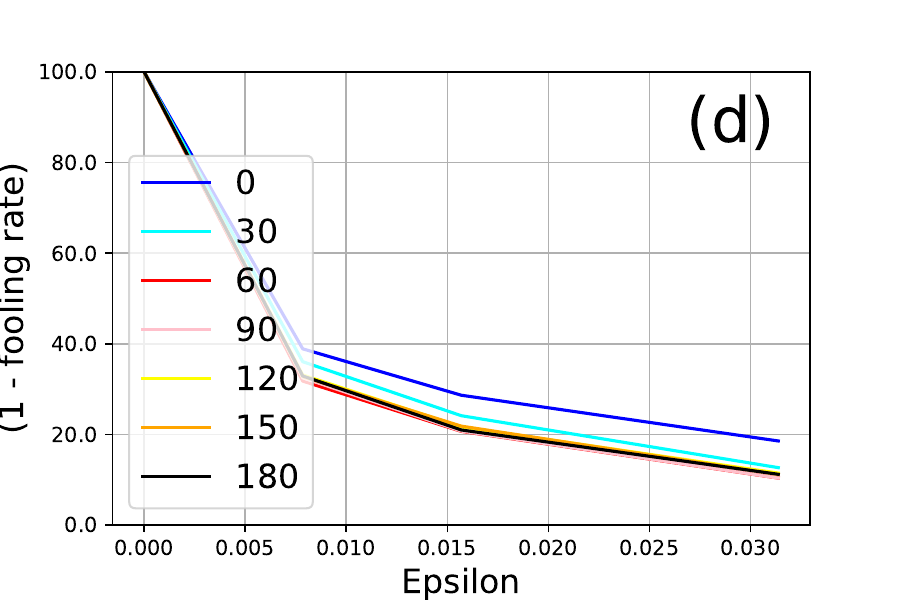}\\
\includegraphics[width=0.24\linewidth]{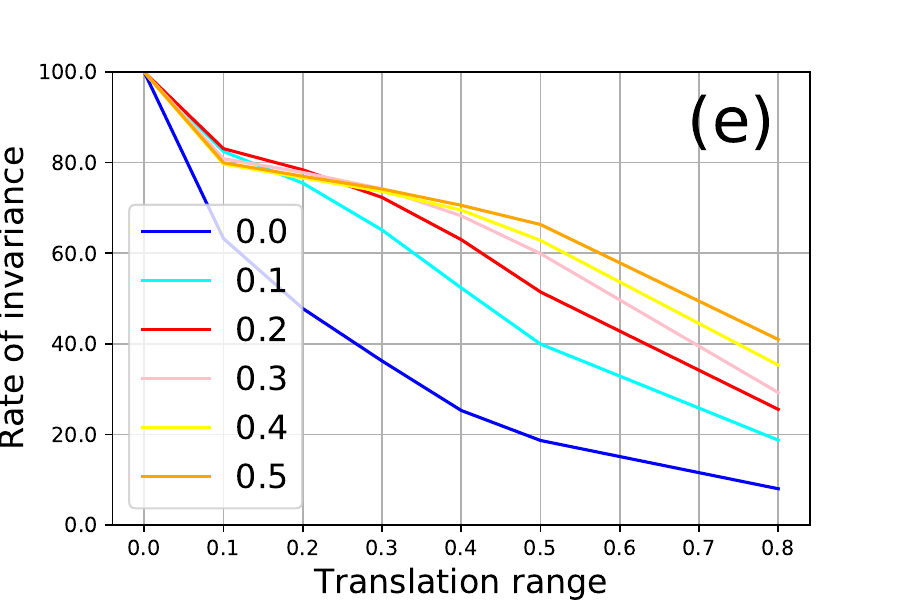}
\includegraphics[width=0.24\linewidth]{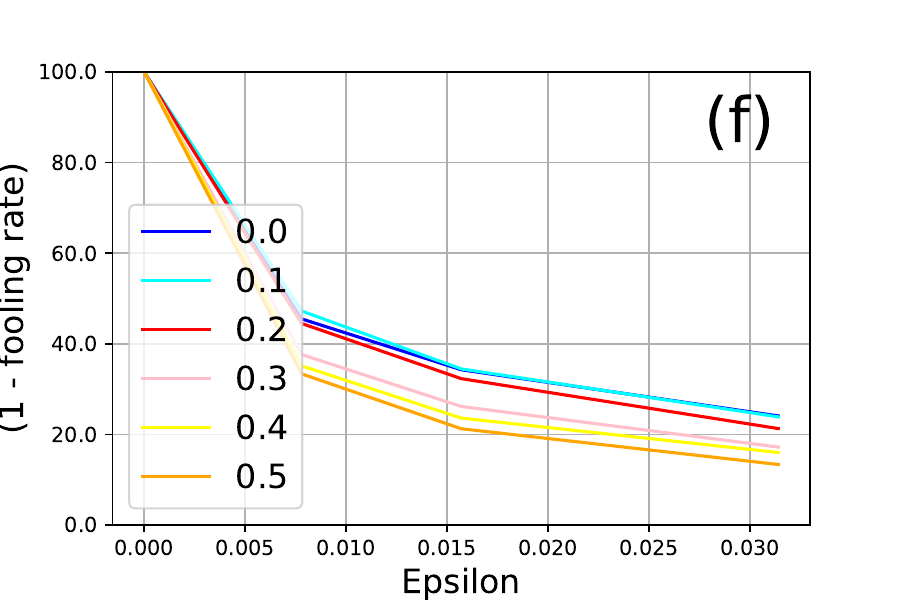}
\rulesep
\includegraphics[width=0.24\linewidth]{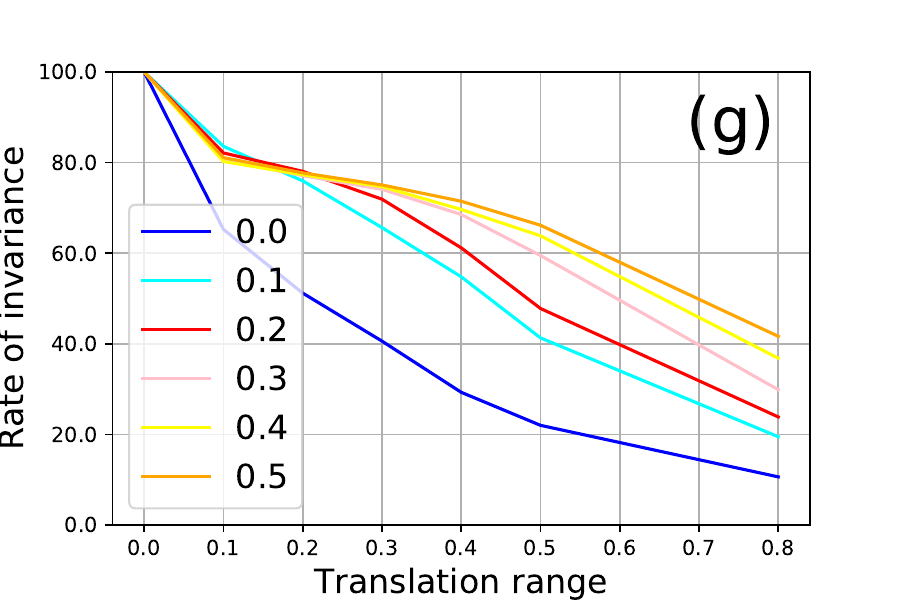}
\includegraphics[width=0.24\linewidth]{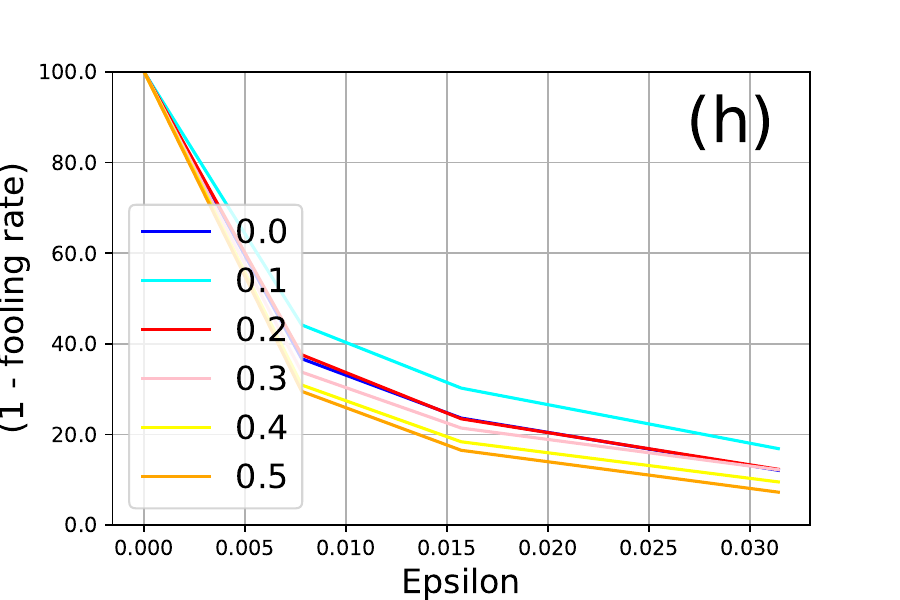}\\
\includegraphics[width=0.24\linewidth]{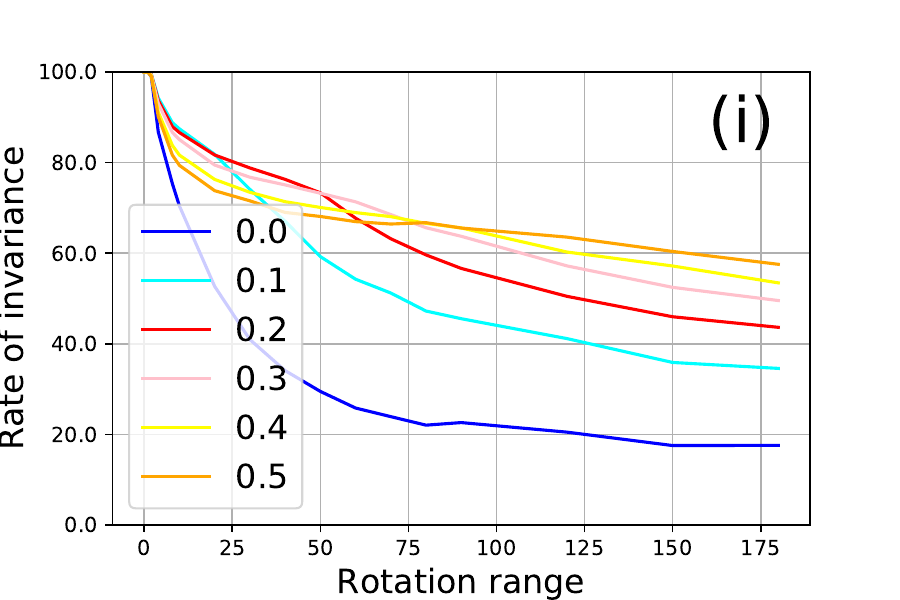}
\includegraphics[width=0.24\linewidth]{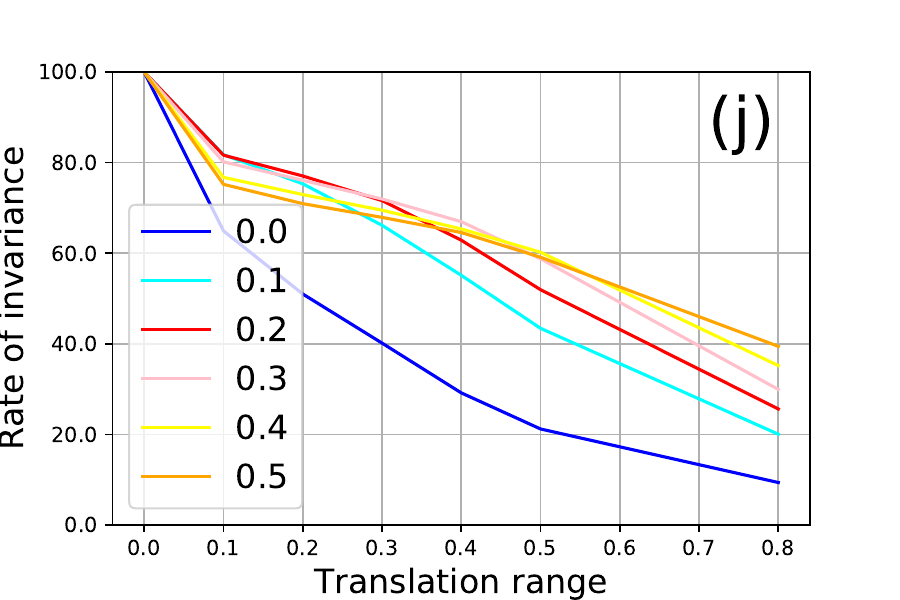}
\includegraphics[width=0.24\linewidth]{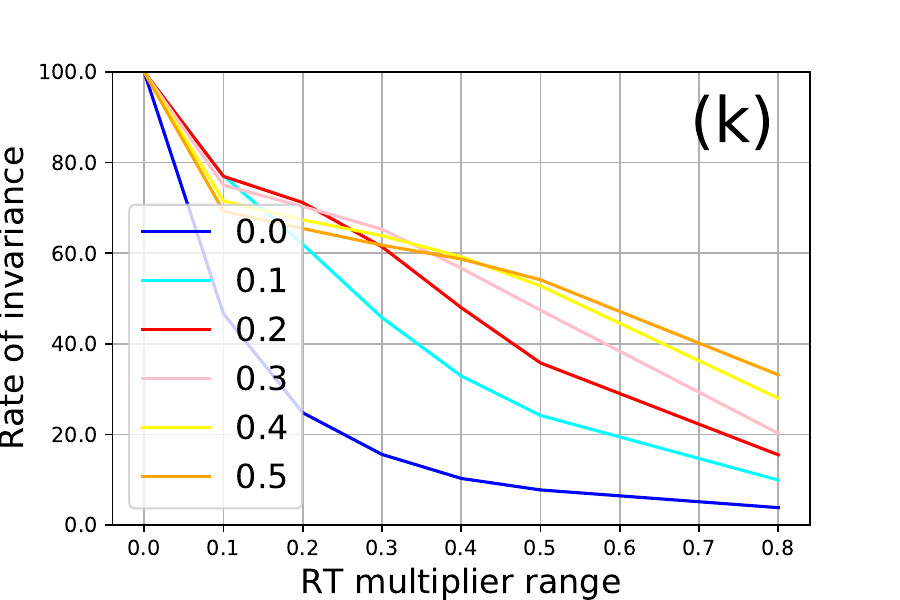}
\includegraphics[width=0.24\linewidth]{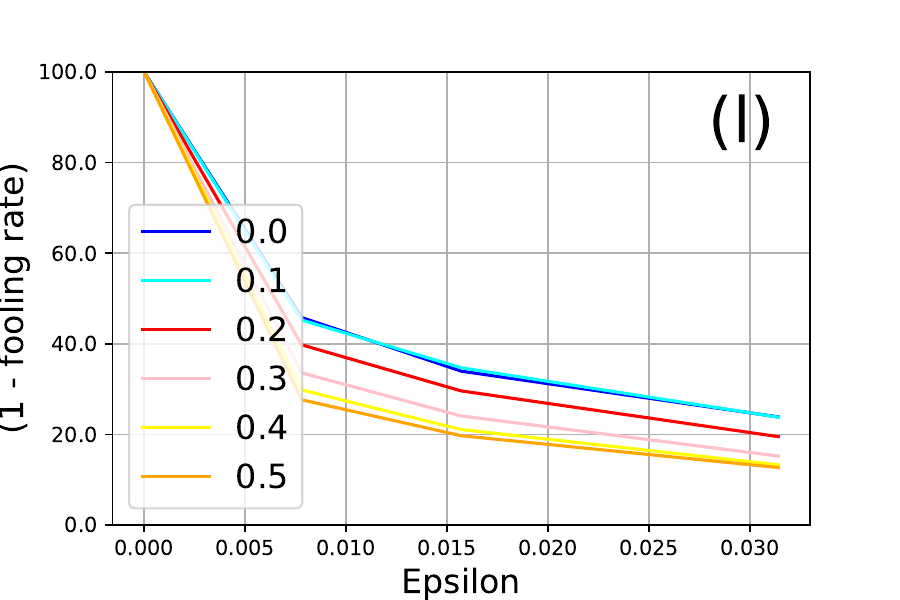}\\
\includegraphics[width=0.24\linewidth]{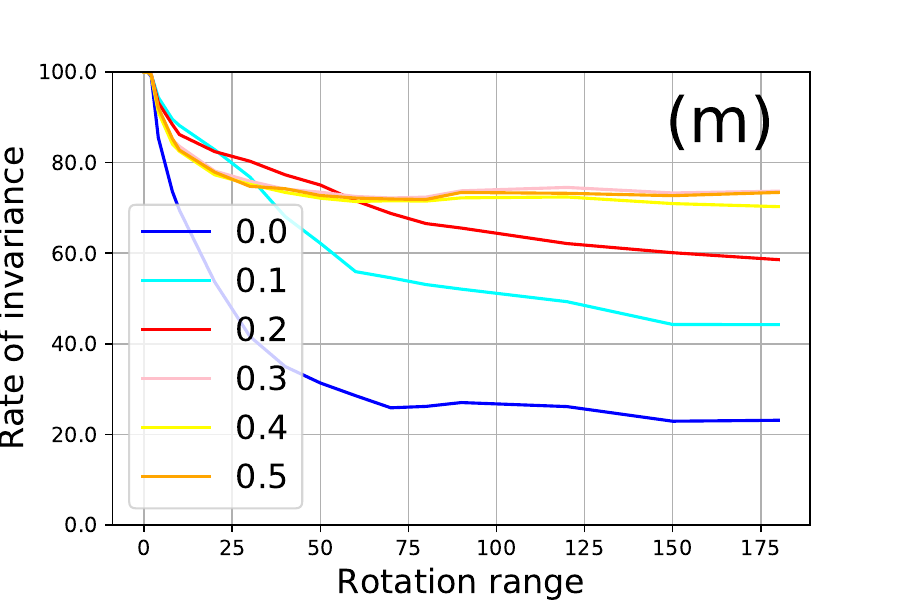}
\includegraphics[width=0.24\linewidth]{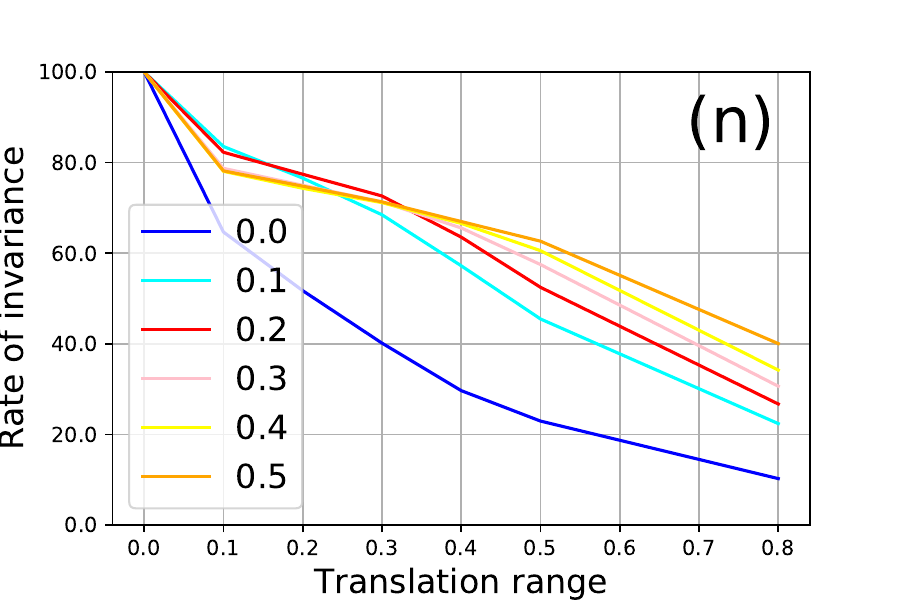}
\includegraphics[width=0.24\linewidth]{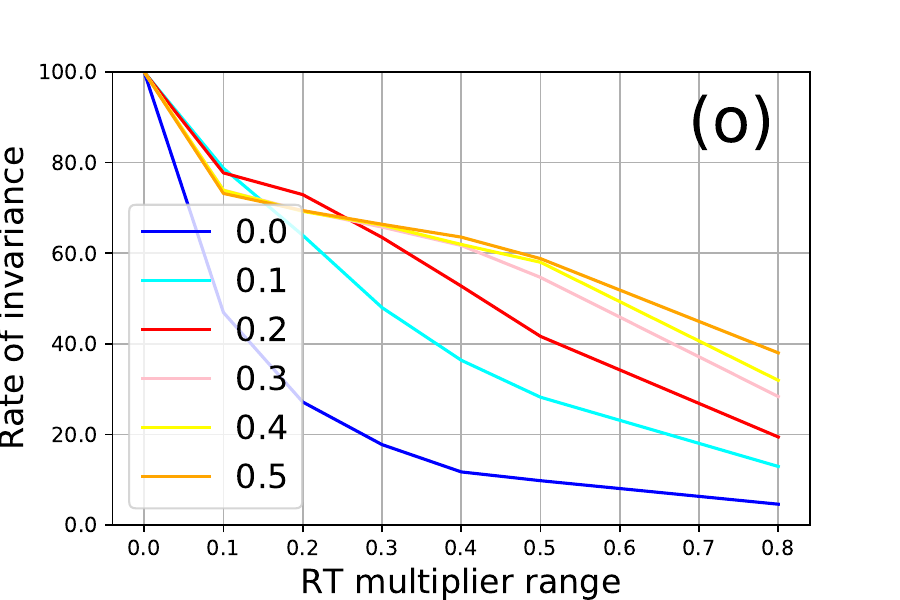}
\includegraphics[width=0.24\linewidth]{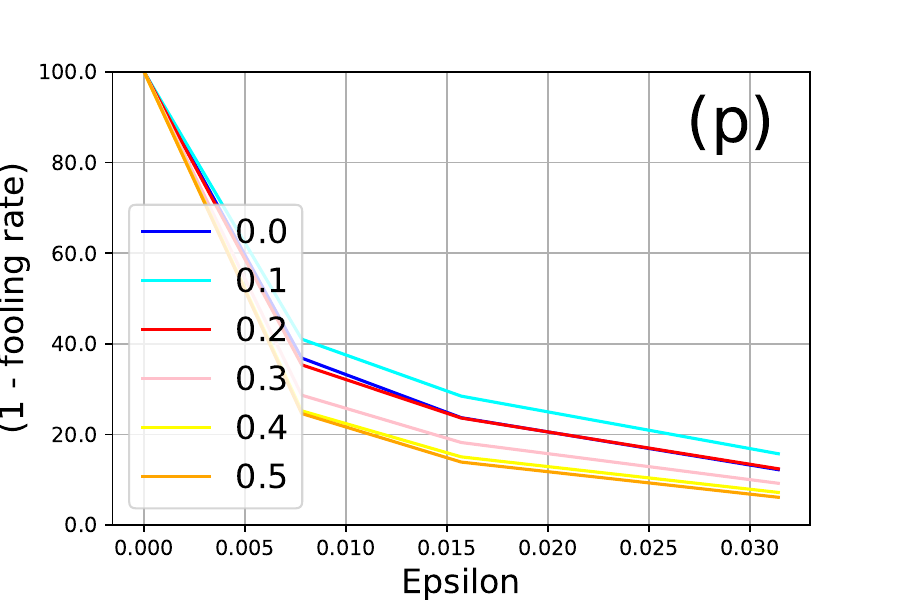}
\end{center}
\vspace{-10pt}
\caption{\textit{(Best viewed in color, zoomed in)} On CIFAR100, For VGG16 model (a-b) \textbf{Aug - R} StdCNN, (c-d) \textbf{Aug - R} GCNN, (e-f) \textbf{Aug - T} StdCNN, (g-h) \textbf{Aug - T} GCNN, (i-l) \textbf{Aug - RT} StdCNN, (m-p) \textbf{Aug - RT}, invariance profiles of StdCNN/GCNN models and corresponding robustness profiles.}
\label{cifar100-stdcnn-gcnn-vgg16-full}
\end{figure*}

\begin{figure*}[!h]
\begin{center}
\includegraphics[width=0.24\linewidth]{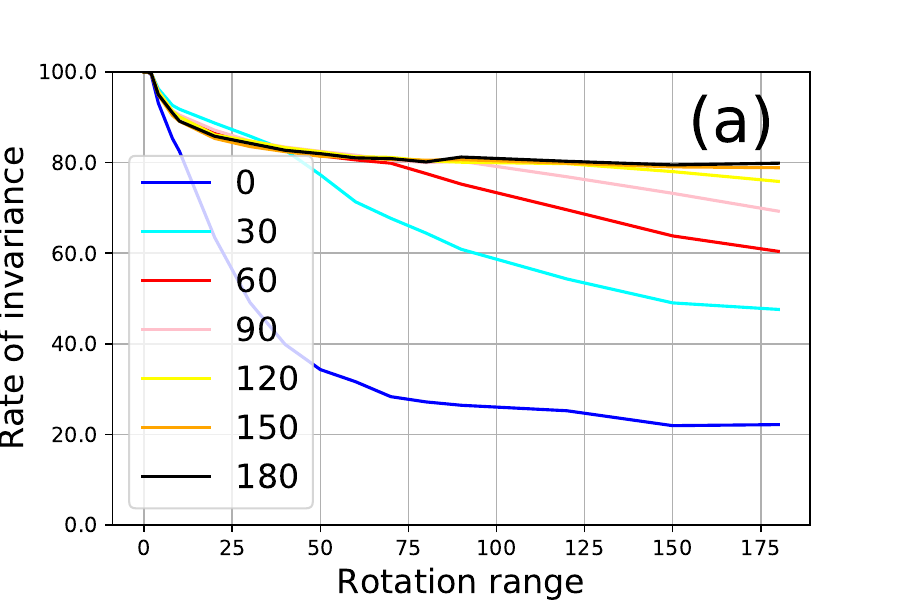} 
\includegraphics[width=0.24\linewidth]{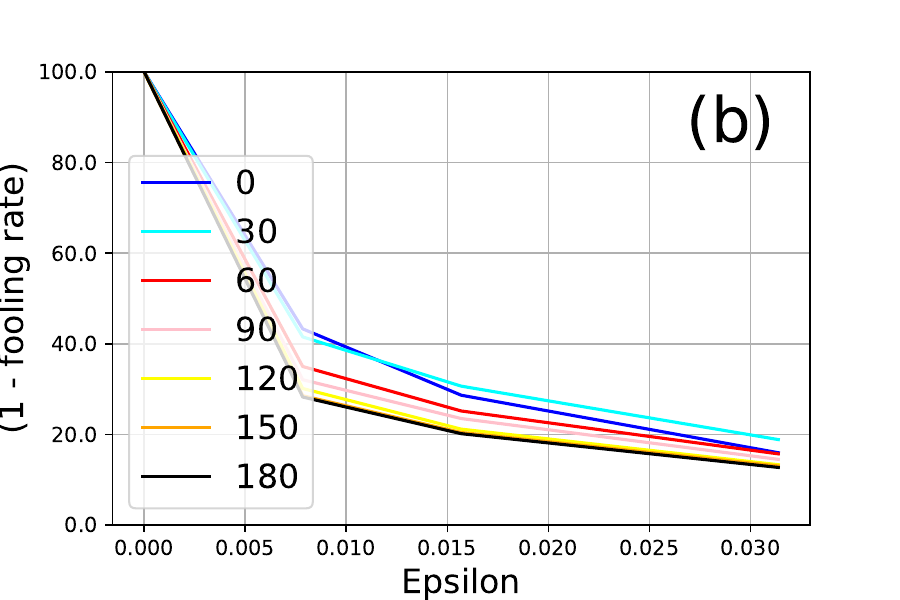}
\rulesep
\includegraphics[width=0.24\linewidth]{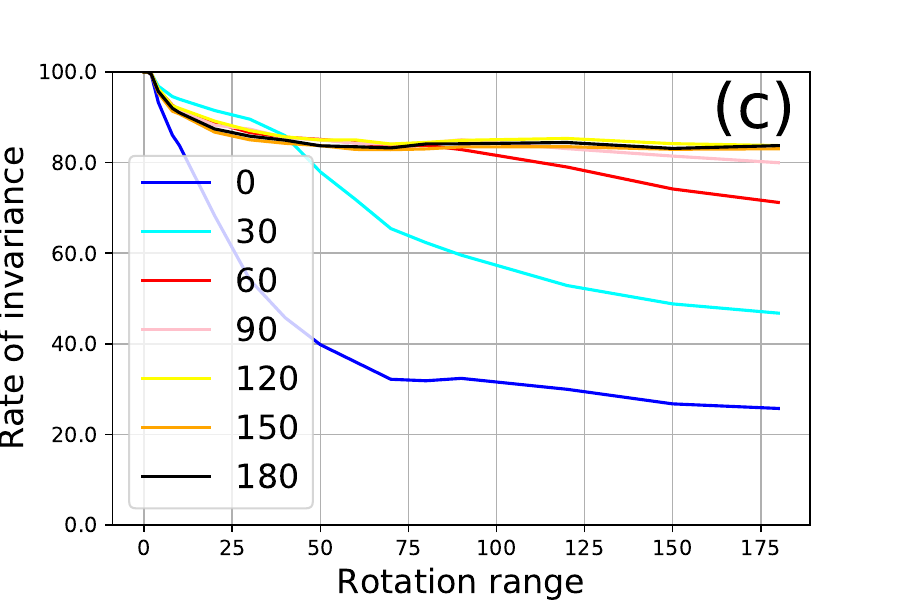} 
\includegraphics[width=0.24\linewidth]{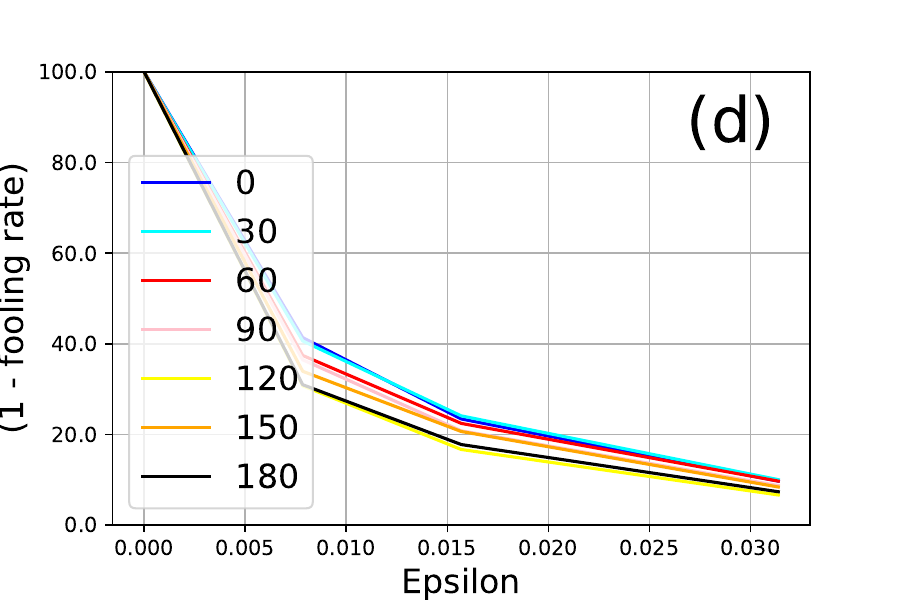}
\includegraphics[width=0.24\linewidth]{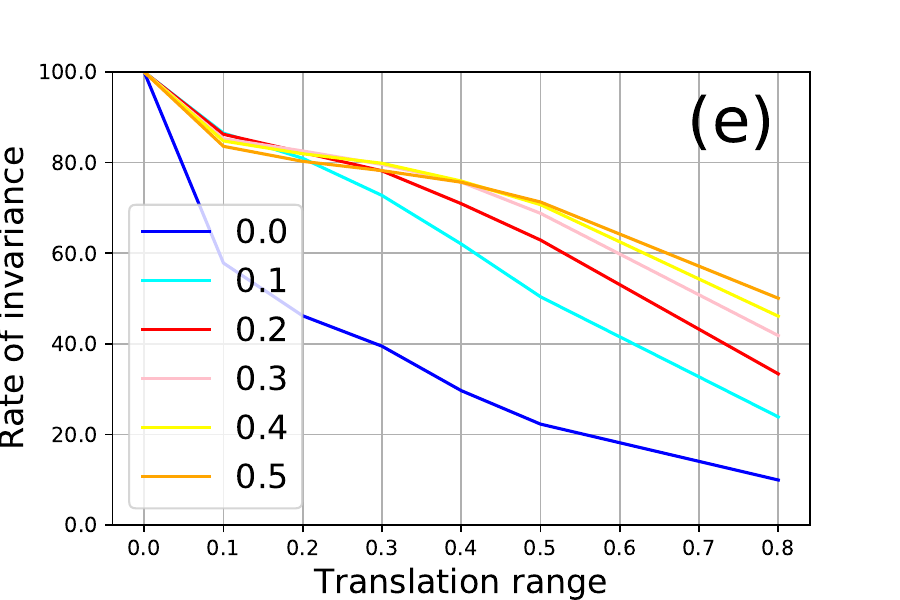}
\includegraphics[width=0.24\linewidth]{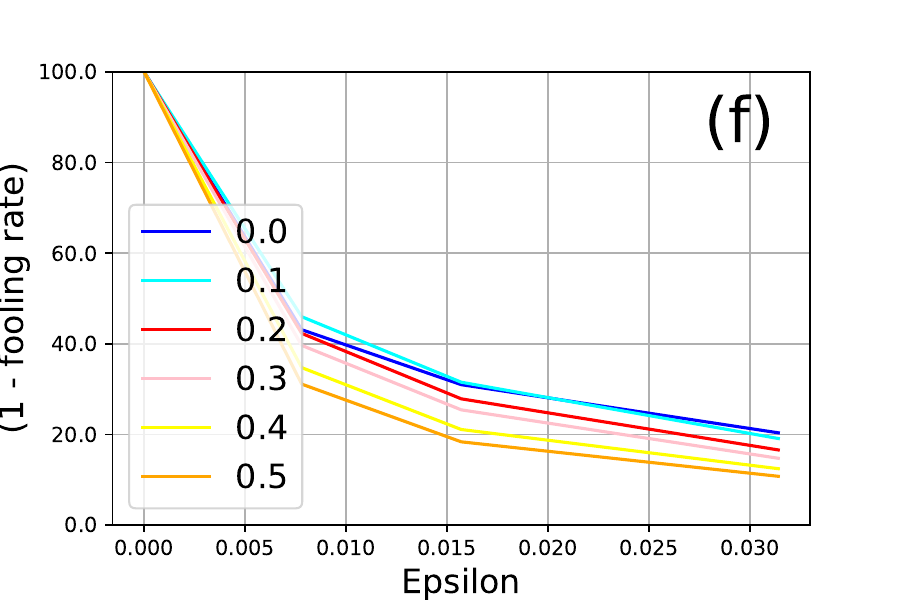}
\rulesep
\includegraphics[width=0.24\linewidth]{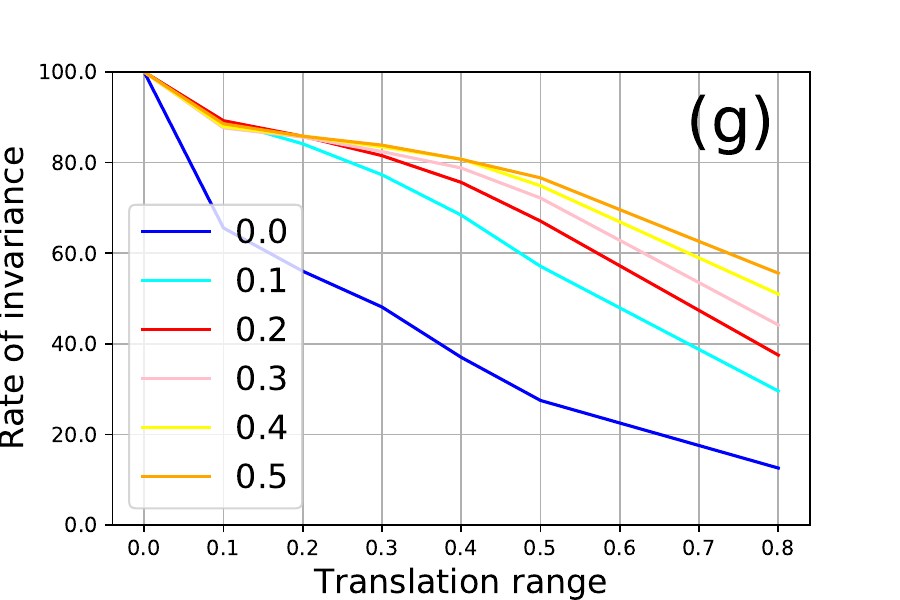}
\includegraphics[width=0.24\linewidth]{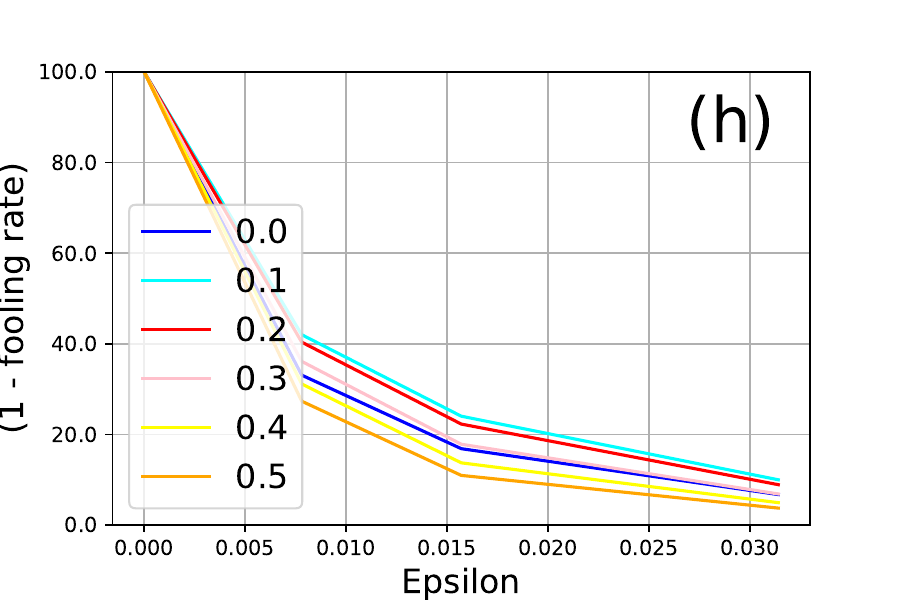}
\includegraphics[width=0.24\linewidth]{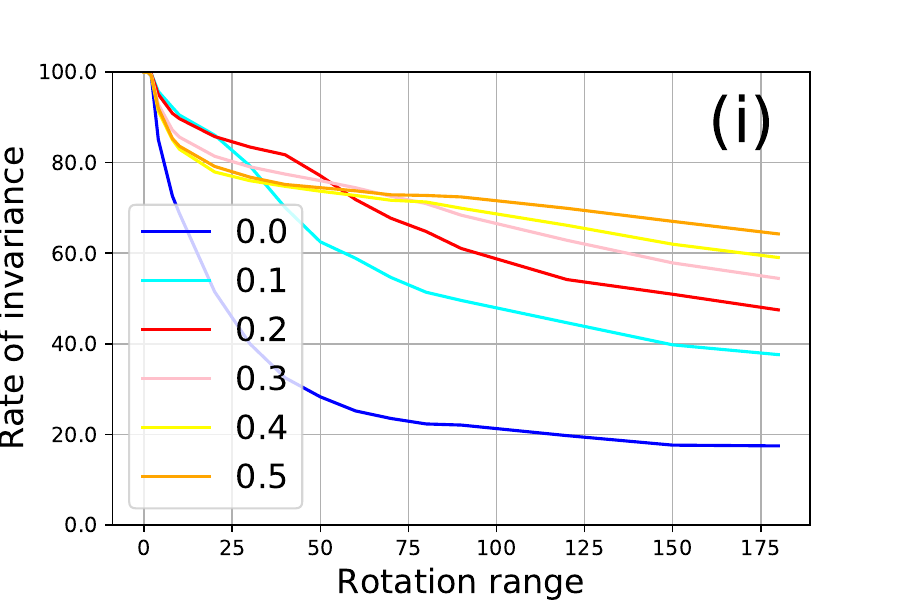}
\includegraphics[width=0.24\linewidth]{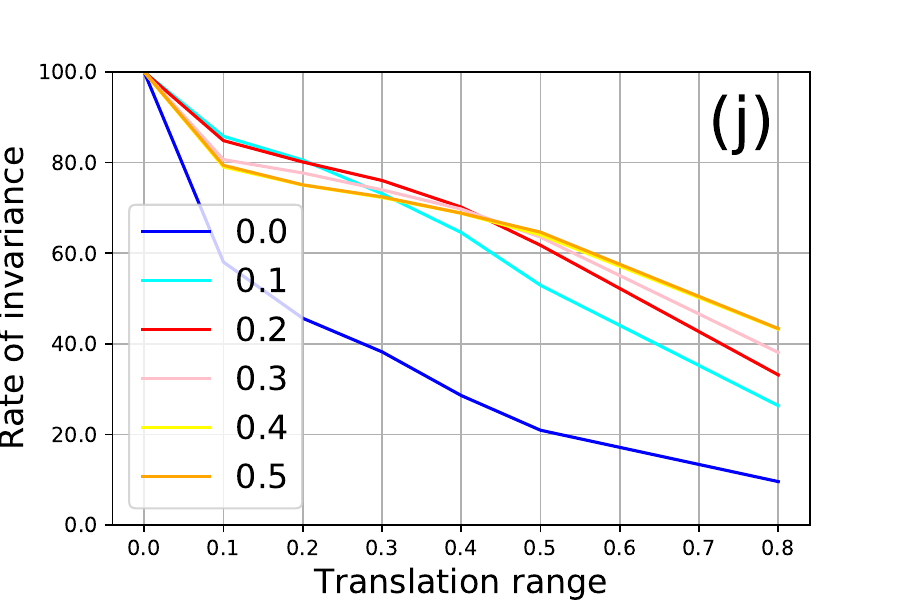}
\includegraphics[width=0.24\linewidth]{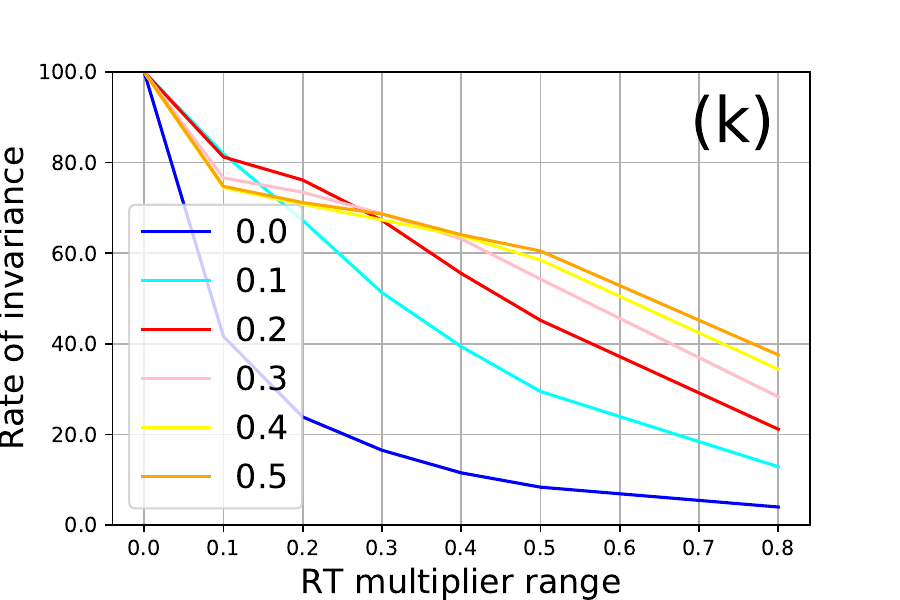}
\includegraphics[width=0.24\linewidth]{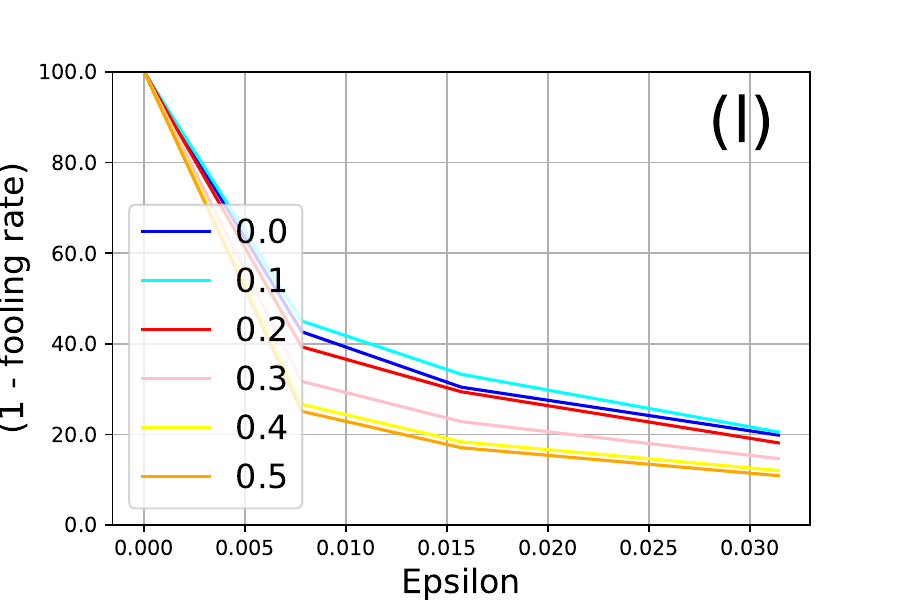}
\includegraphics[width=0.24\linewidth]{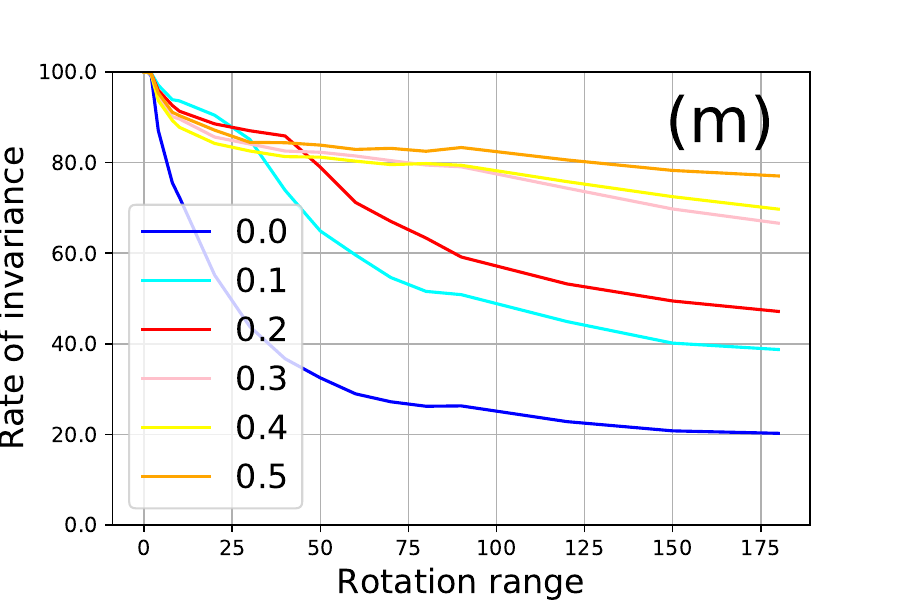}
\includegraphics[width=0.24\linewidth]{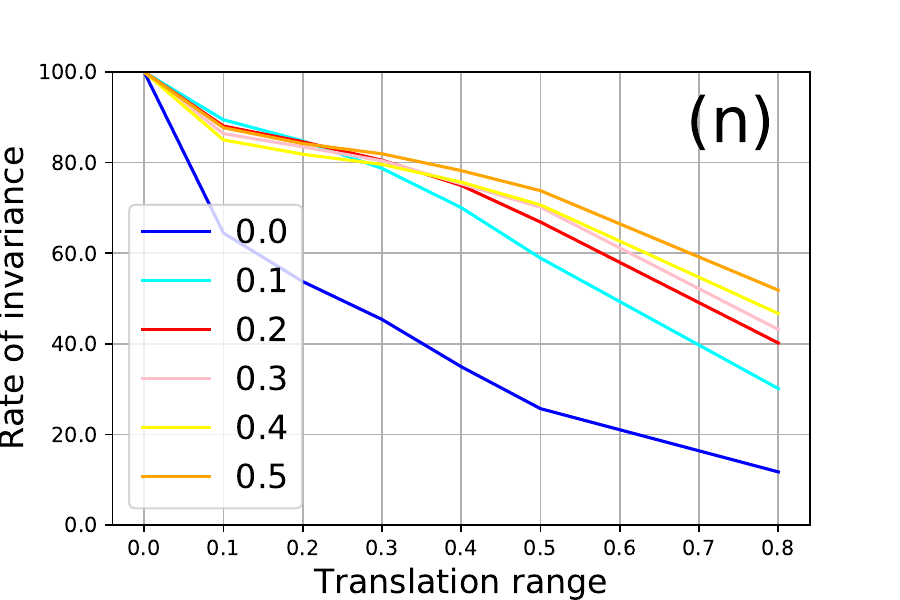}
\includegraphics[width=0.24\linewidth]{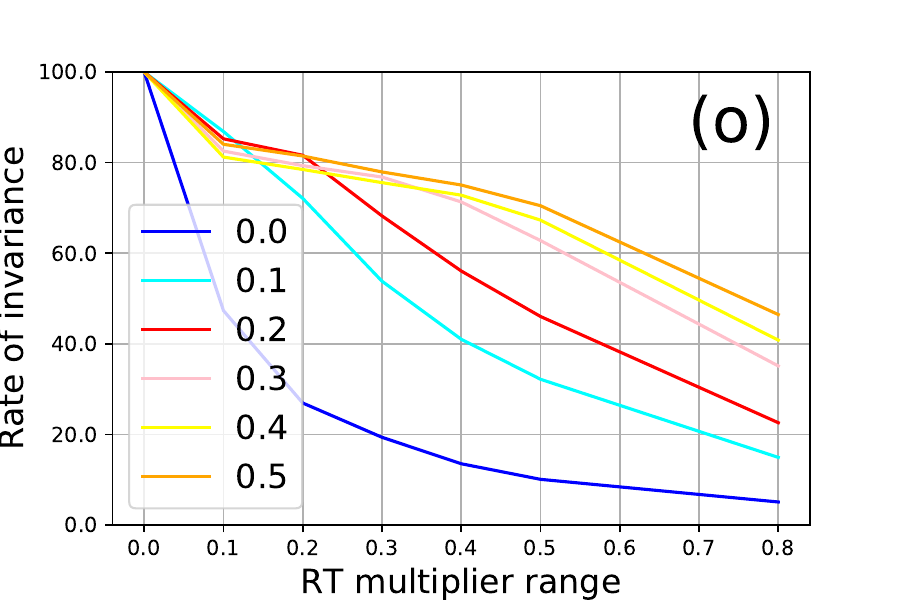}
\includegraphics[width=0.24\linewidth]{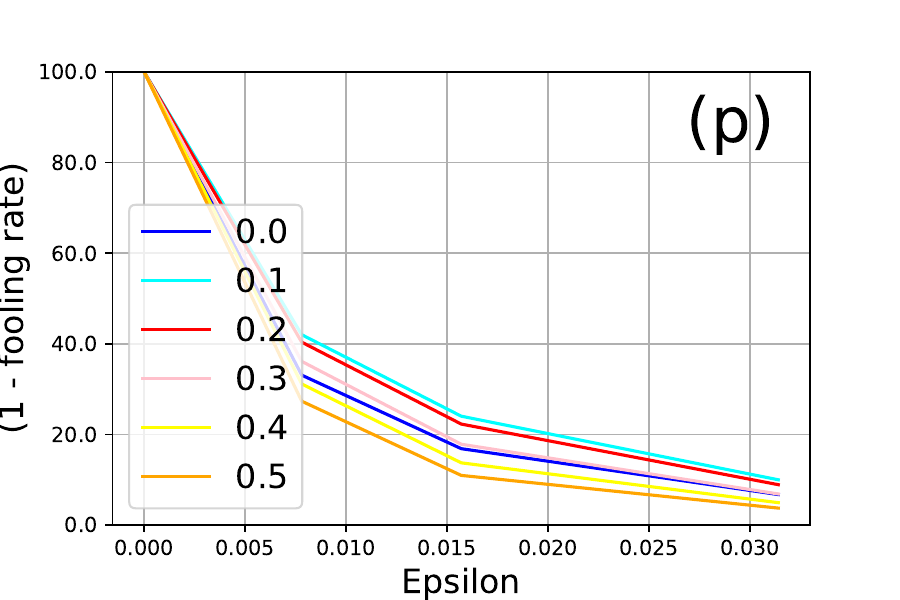}
\end{center}
\vspace{-10pt}
\caption{\textit{(Best viewed in color, zoomed in)} On CIFAR100, For ResNet18 model (a-b) \textbf{Aug - R} StdCNN, (c-d) \textbf{Aug - R} GCNN, (e-f) \textbf{Aug - T} StdCNN, (g-h) \textbf{Aug - T} GCNN, (i-l) \textbf{Aug - RT} StdCNN, (m-p) \textbf{Aug - RT}, invariance profiles of StdCNN/GCNN models and corresponding robustness profiles.}
\label{cifar100-stdcnn-gcnn-resnet18-full}
\end{figure*}

\vspace{-7pt}
In the main paper, we presented results for experiments carried out using VGG16 on CIFAR10. Please see Figs \ref{cifar100-stdcnn-gcnn-vgg16-full} and \ref{cifar100-stdcnn-gcnn-resnet18-full} for additional experiments on CIFAR100 using the VGG16 and ResNet18 models, respectively. These complement the experiments in the main paper and show that our results continue to hold across different models and datasets.

\vspace{-7pt}
\section{Additional Results on Tiny-ImageNet} \label{suppl:exp-results-tiny-imagenet}
\begin{figure*}[!h]
\begin{center}
\includegraphics[width=0.24\linewidth]{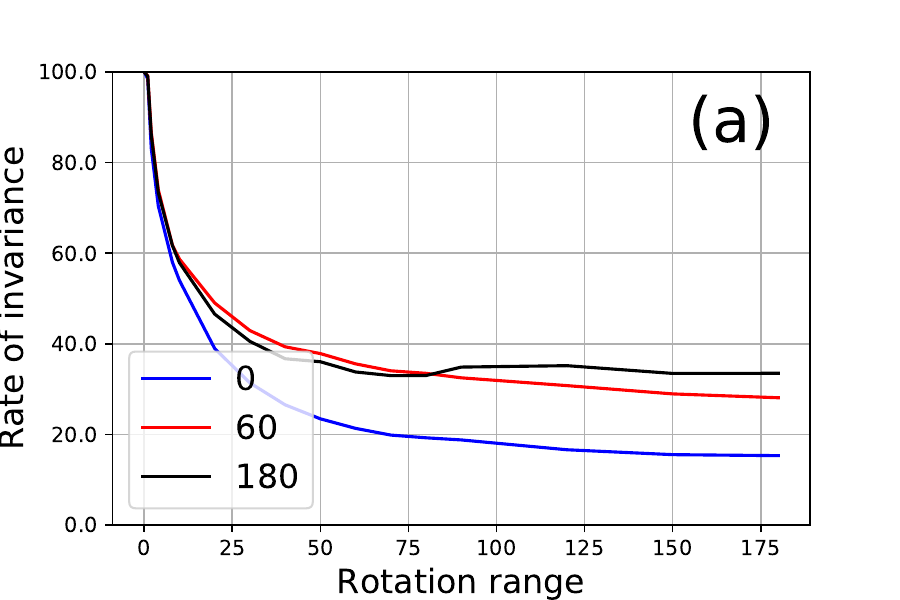}
\includegraphics[width=0.24\linewidth]{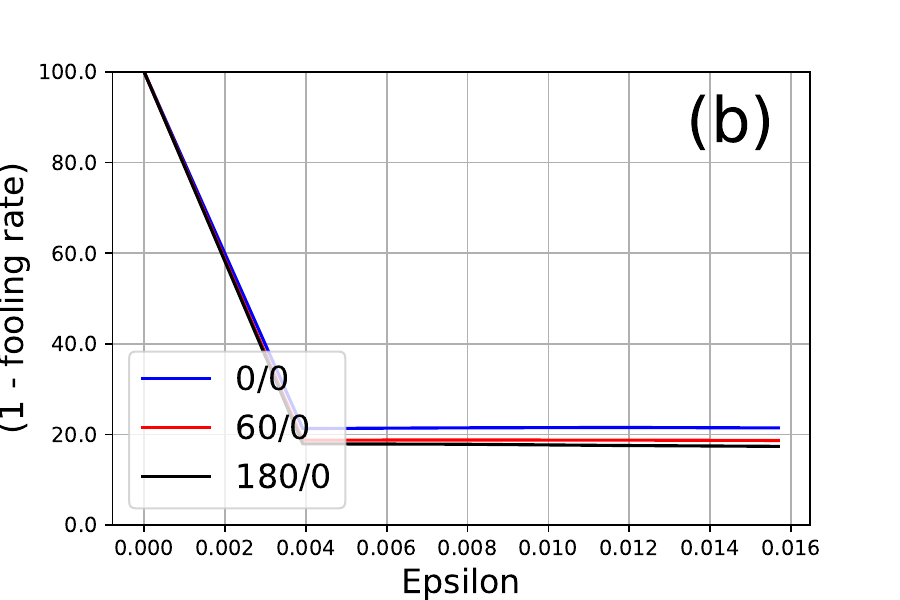}
\rulesep
\includegraphics[width=0.24\linewidth]{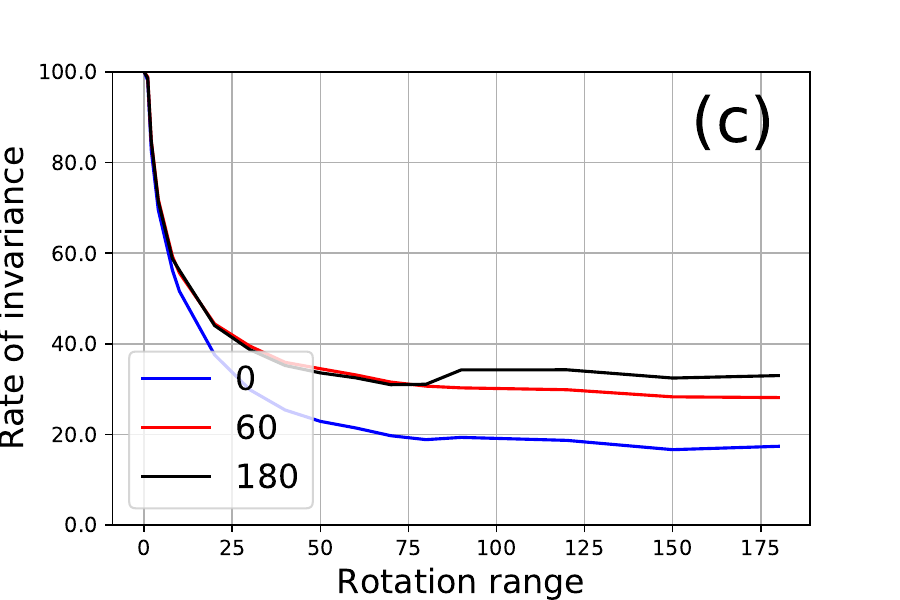}
\includegraphics[width=0.24\linewidth]{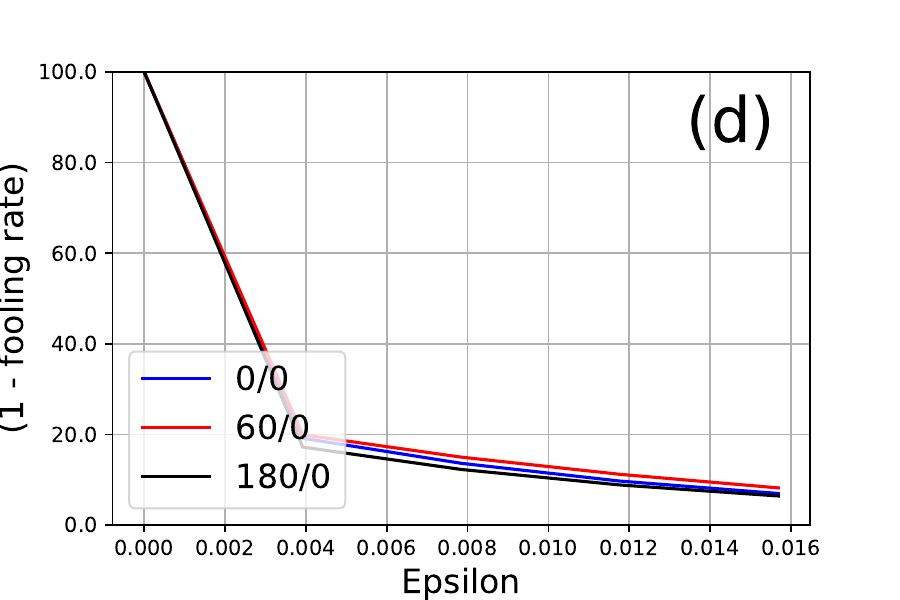}\\
\includegraphics[width=0.24\linewidth]{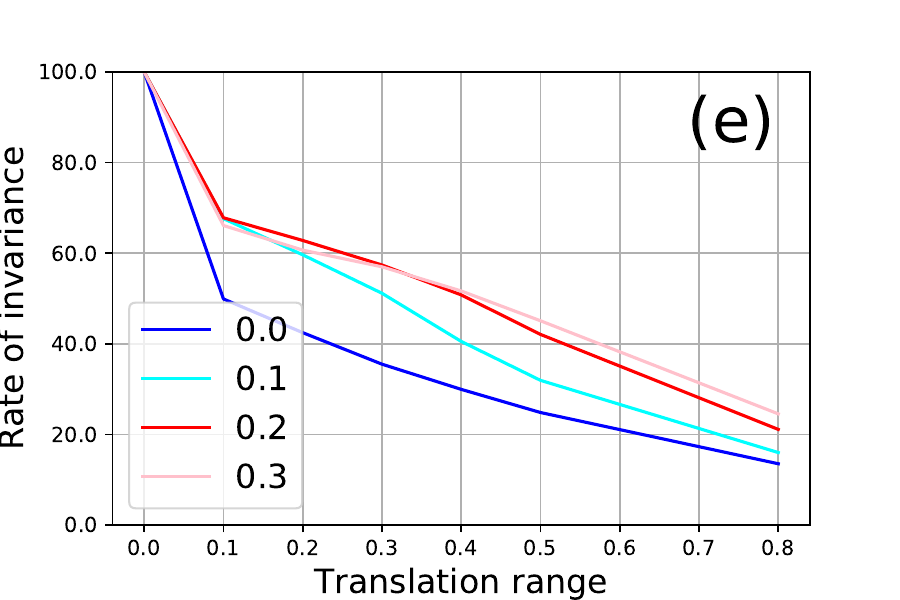}
\includegraphics[width=0.24\linewidth]{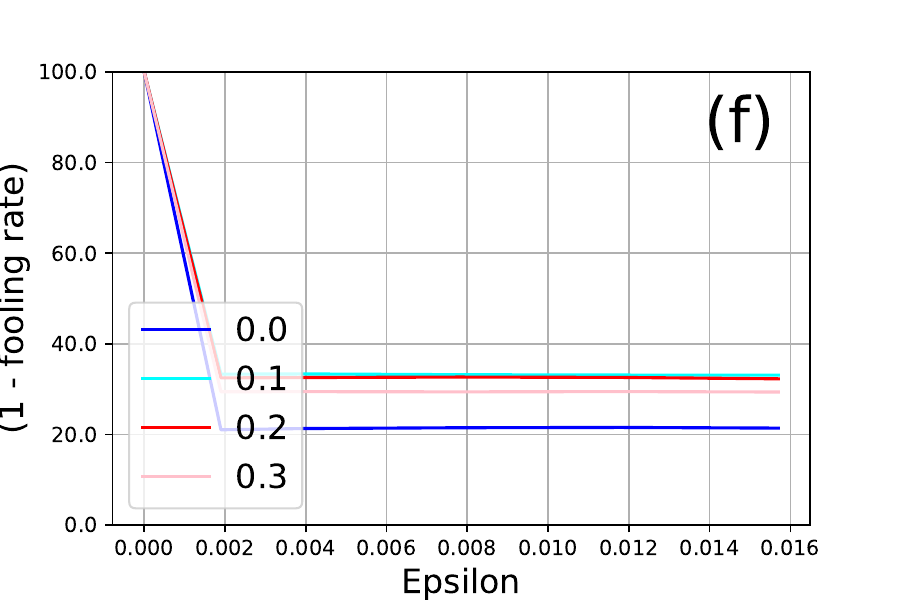}\\
\includegraphics[width=0.24\linewidth]{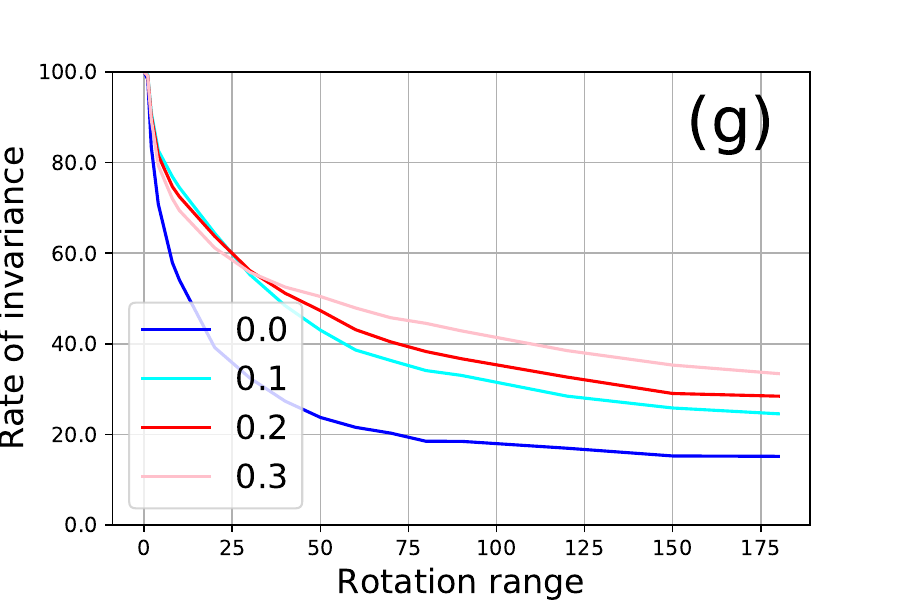}
\includegraphics[width=0.24\linewidth]{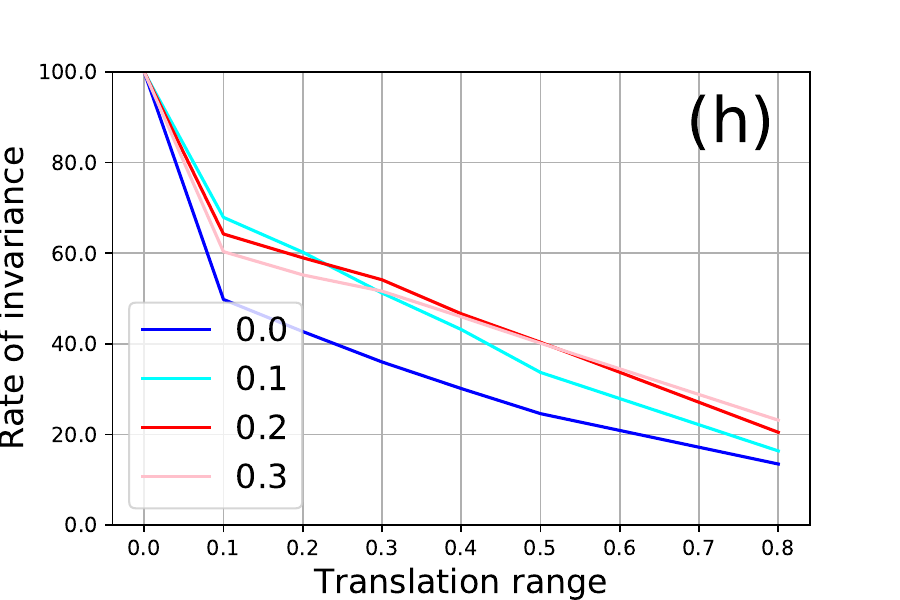}
\includegraphics[width=0.24\linewidth]{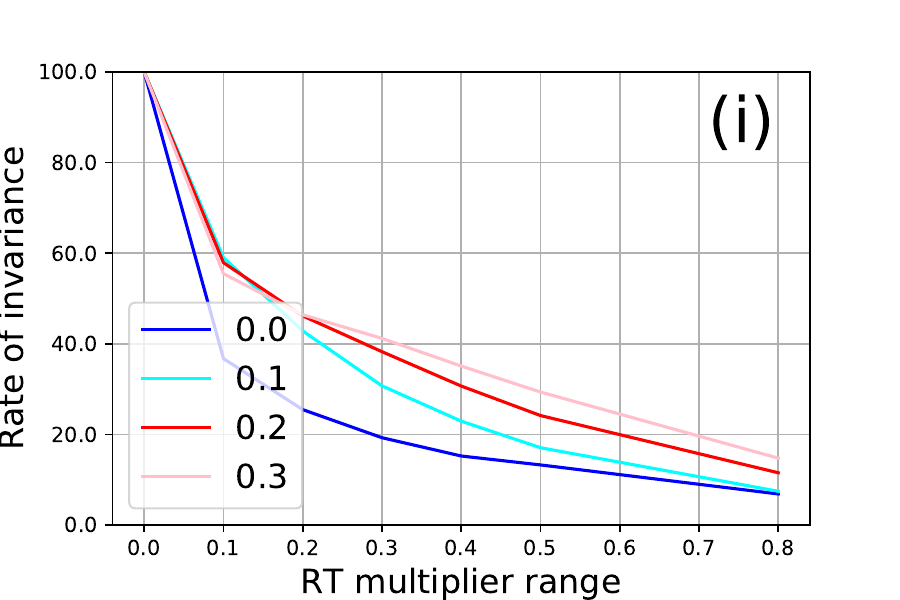}
\includegraphics[width=0.24\linewidth]{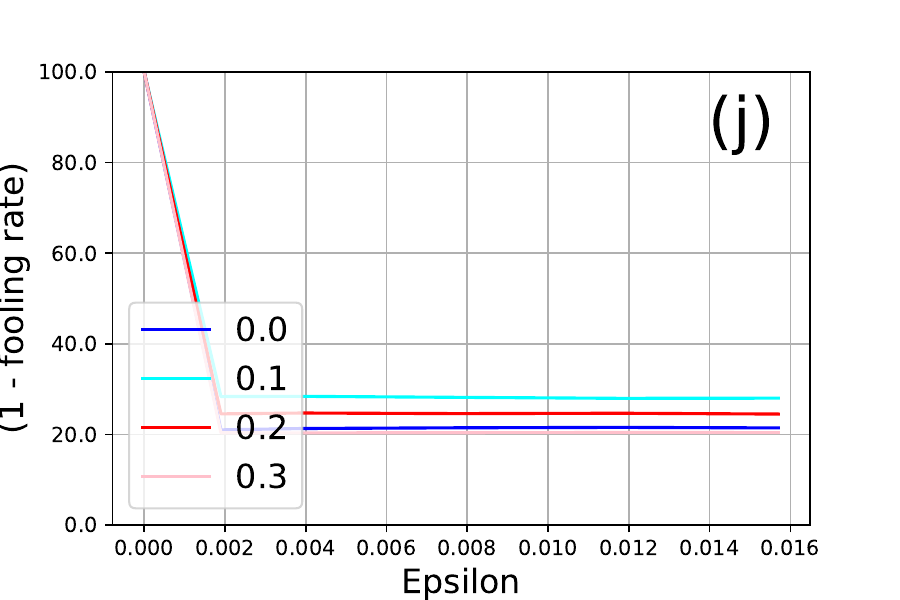}\\
\includegraphics[width=0.24\linewidth]{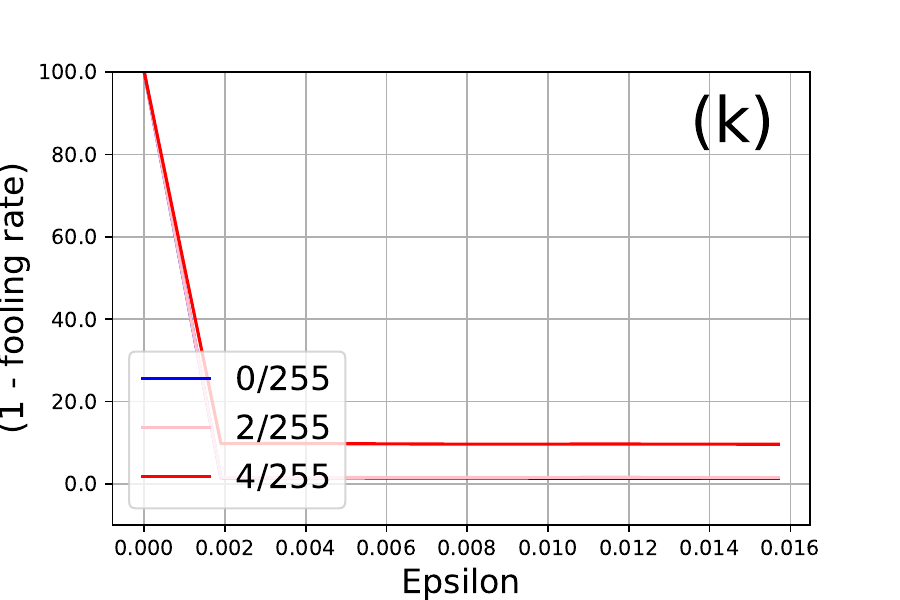}
\includegraphics[width=0.24\linewidth]{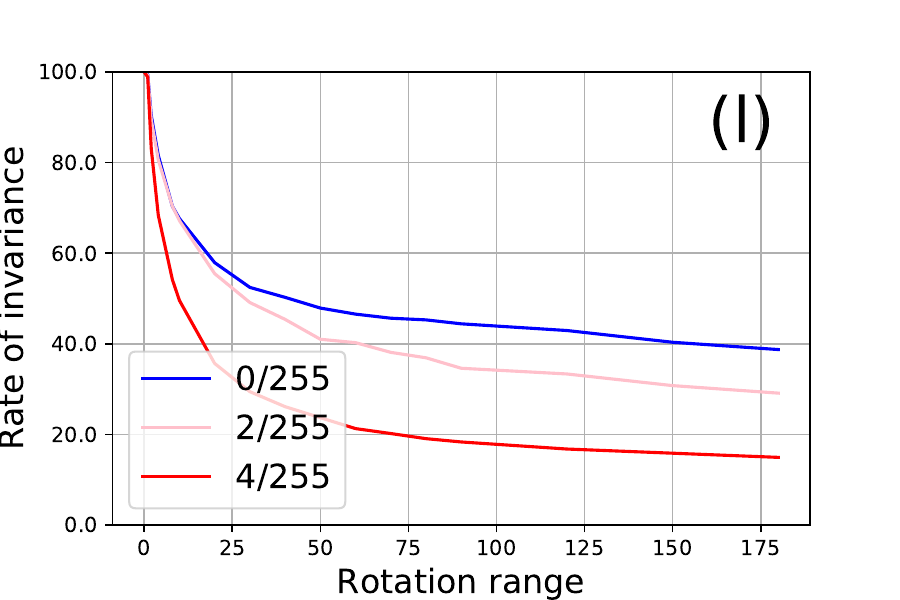}
\includegraphics[width=0.24\linewidth]{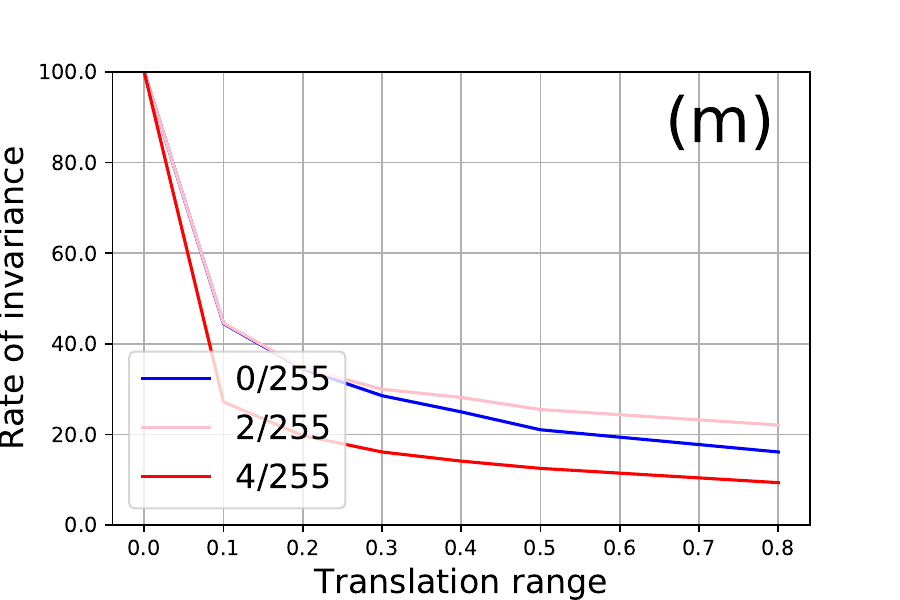}
\includegraphics[width=0.24\linewidth]{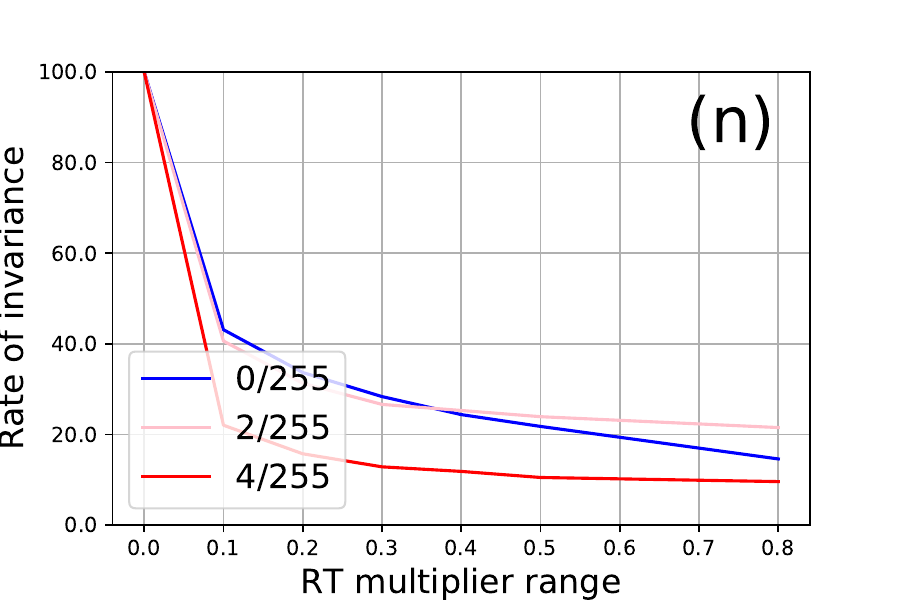}
\end{center}
\vspace{-10pt}
\caption{\textit{(Best viewed in color, zoomed in)} On Tiny-ImageNet for ResNet18 model: (a-b) \textbf{Aug - R} StdCNN, (c-d) \textbf{Aug - R} GCNN, (e-f) \textbf{Aug - T} StdCNN, (g-j) \textbf{Aug - RT} StdCNN, (k-n) \textbf{Adv - PGD} StdCNN, invariance profiles of StdCNN/GCNN models and corresponding robustness profiles.}
\label{tiny-stdcnn-gcnn-resnet-full}
\end{figure*}

\begin{figure*}[!h]
\begin{center}
\includegraphics[width=0.24\linewidth]{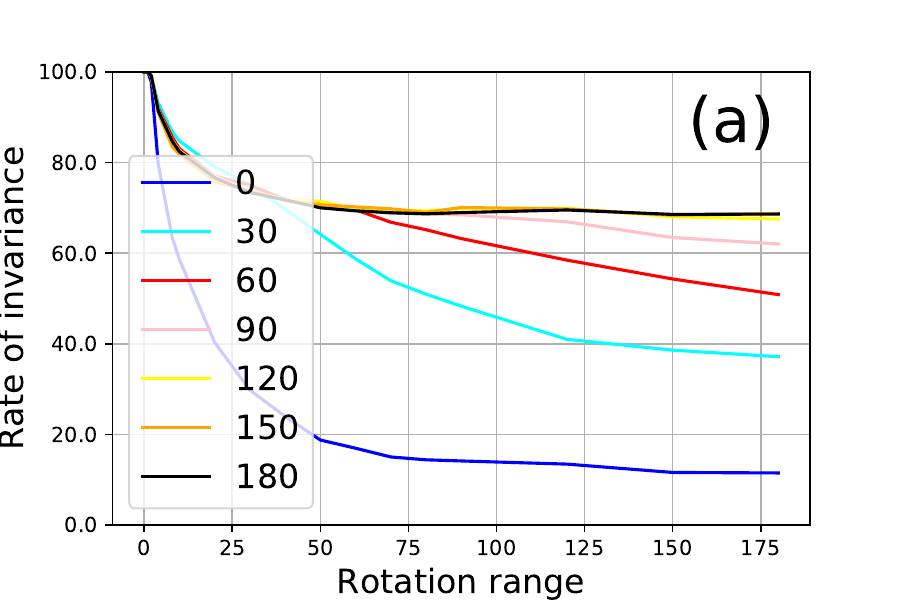}
\includegraphics[width=0.24\linewidth]{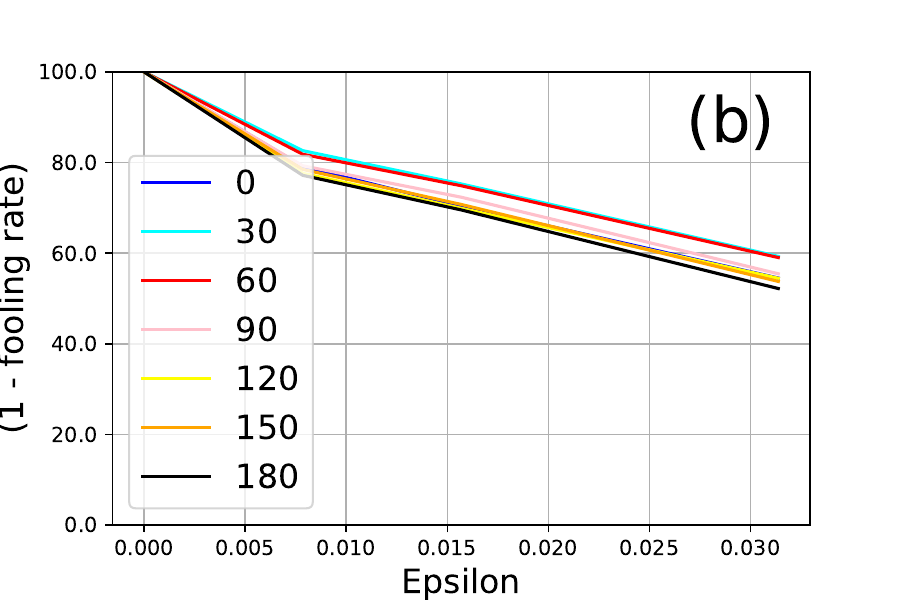}
\rulesep
\includegraphics[width=0.24\linewidth]{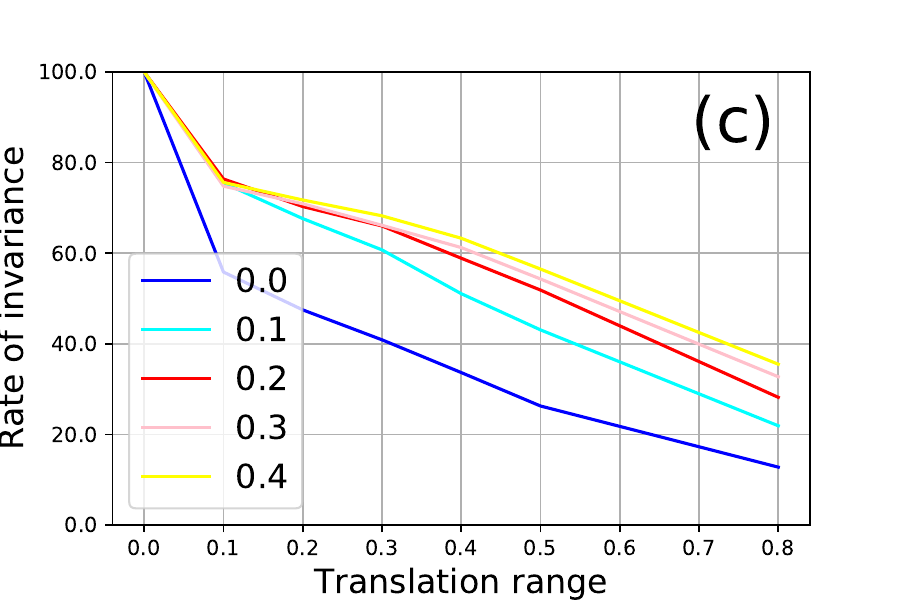}
\includegraphics[width=0.24\linewidth]{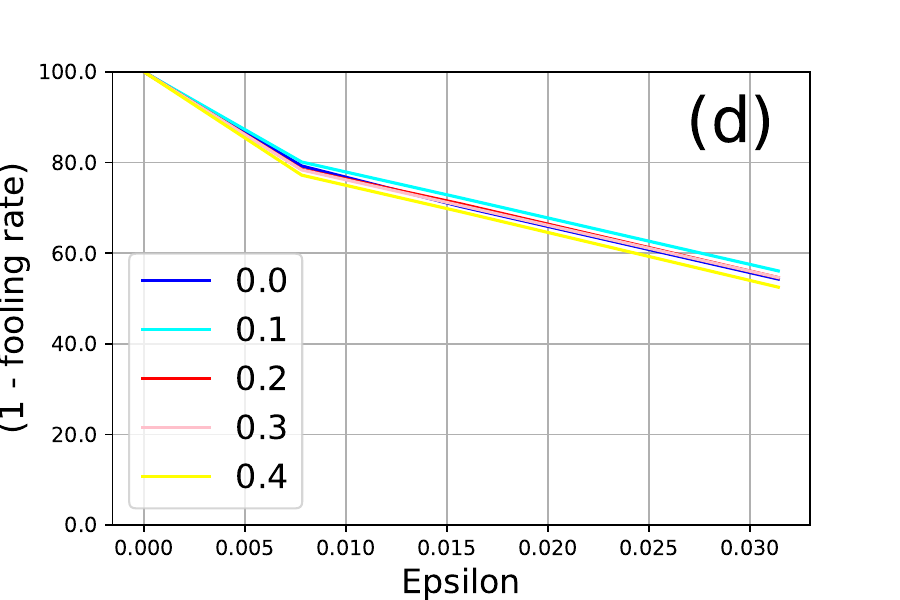}\\
\includegraphics[width=0.24\linewidth]{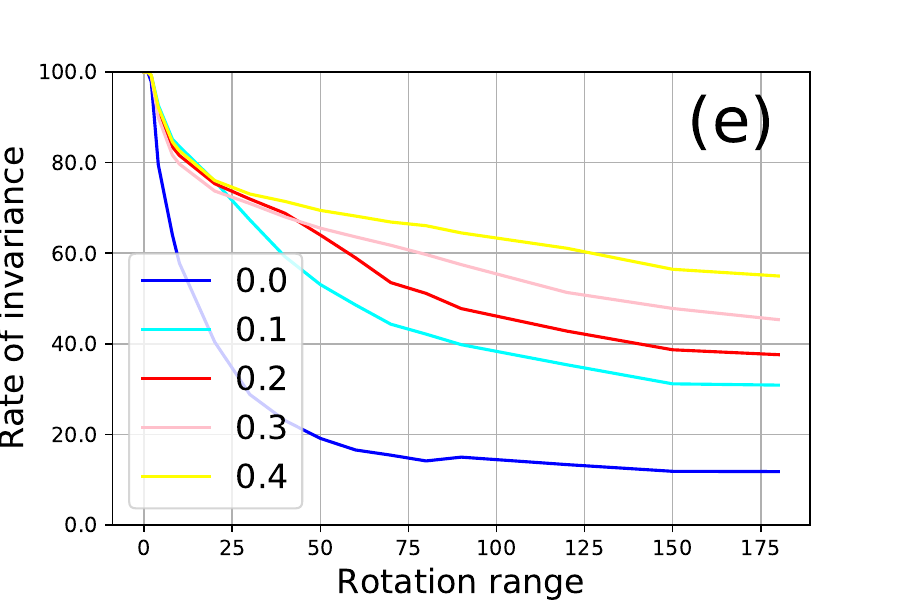}
\includegraphics[width=0.24\linewidth]{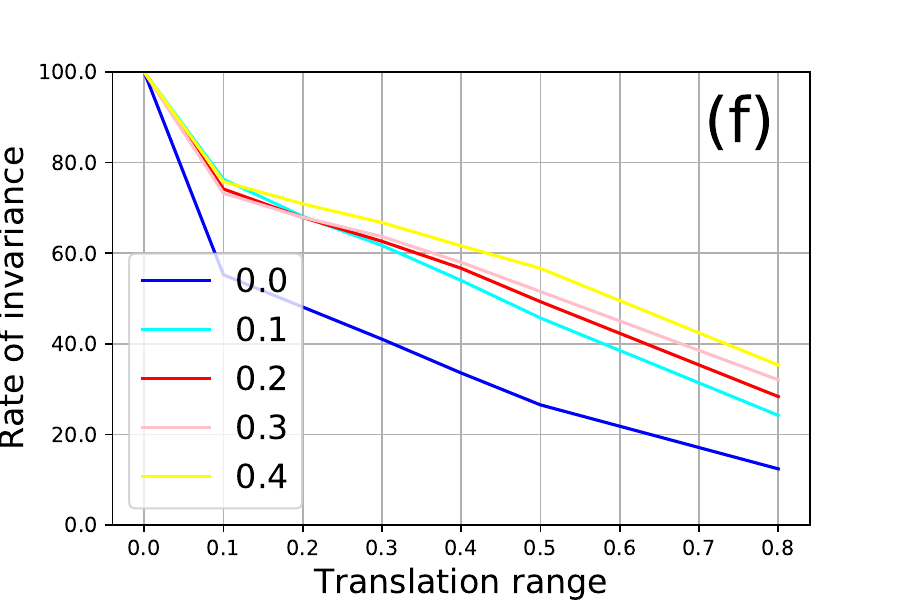}
\includegraphics[width=0.24\linewidth]{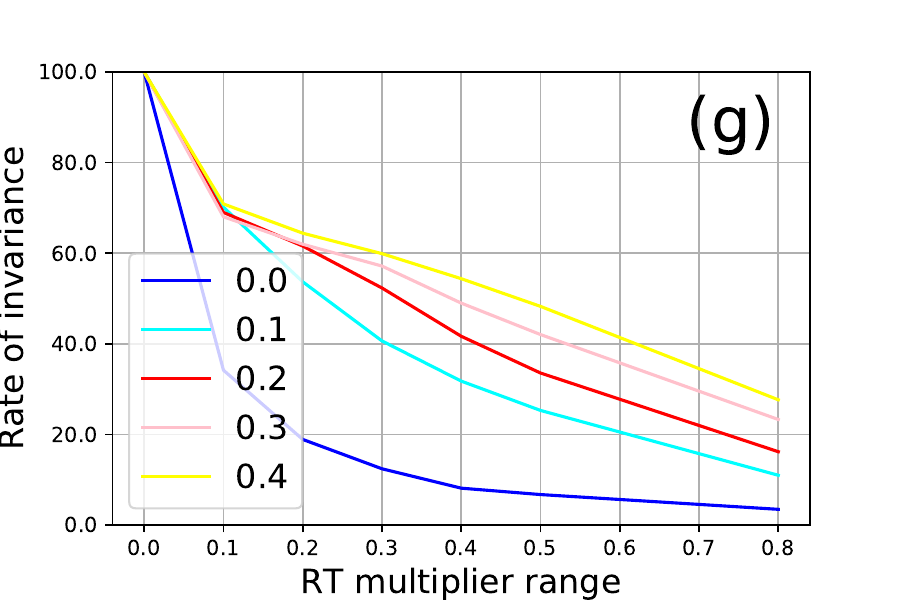}
\includegraphics[width=0.24\linewidth]{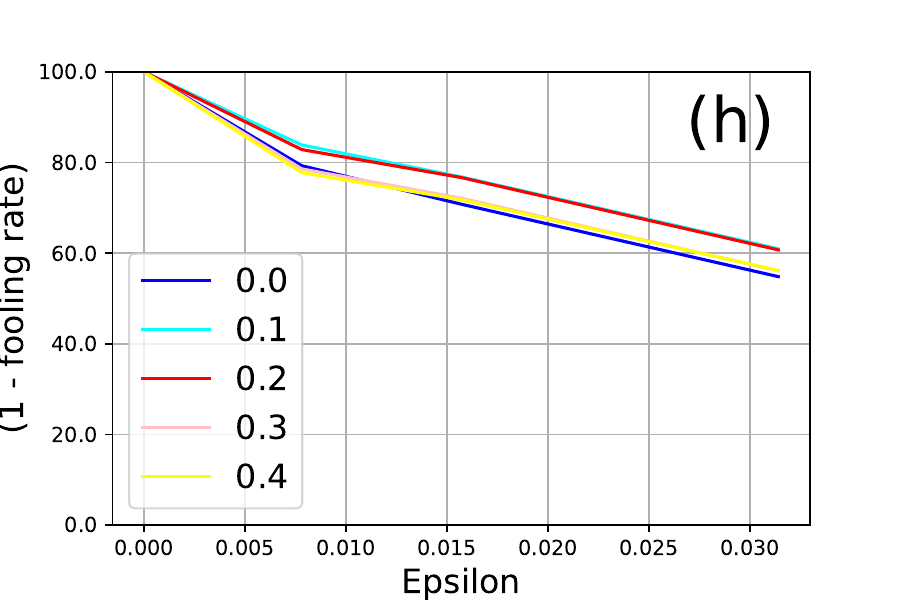}\\
\includegraphics[width=0.24\linewidth]{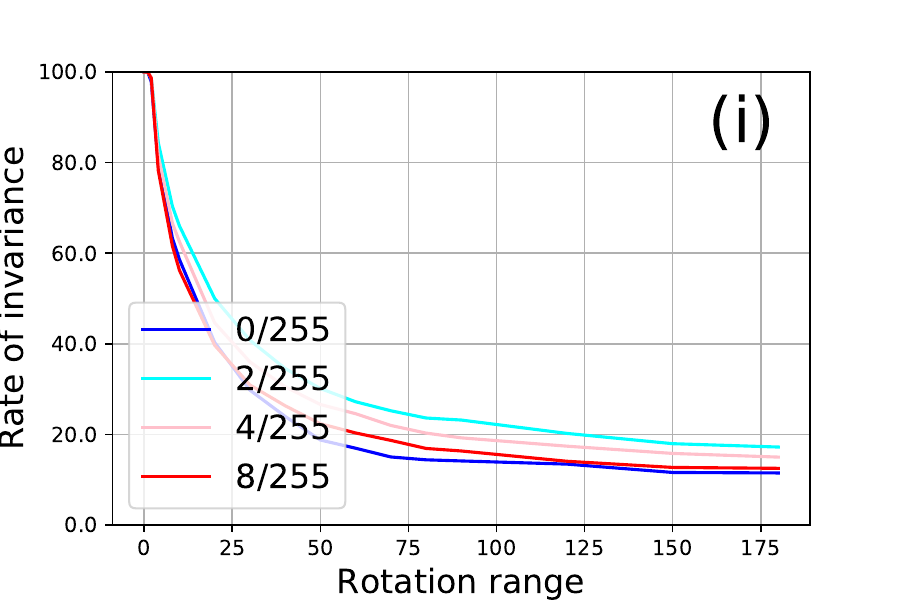}
\includegraphics[width=0.24\linewidth]{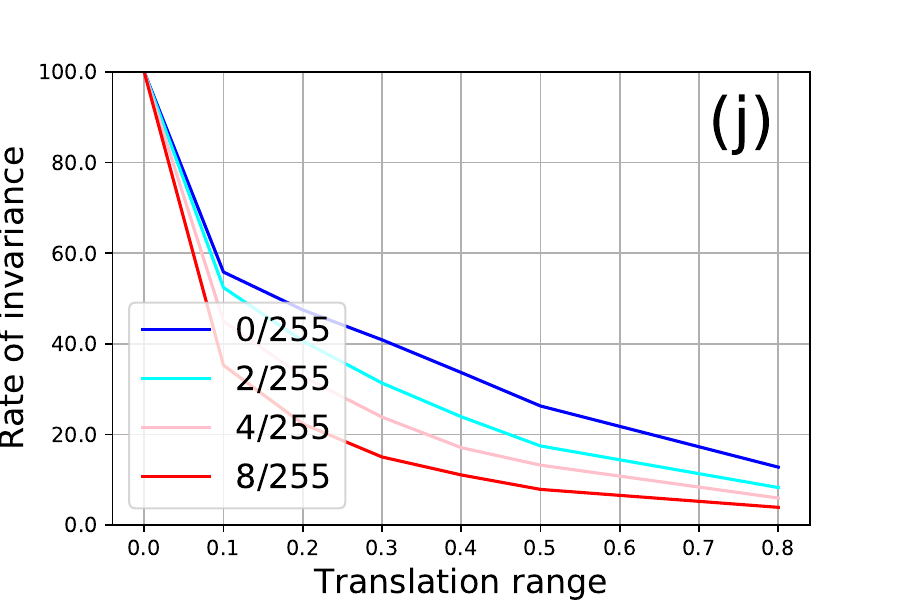}
\includegraphics[width=0.24\linewidth]{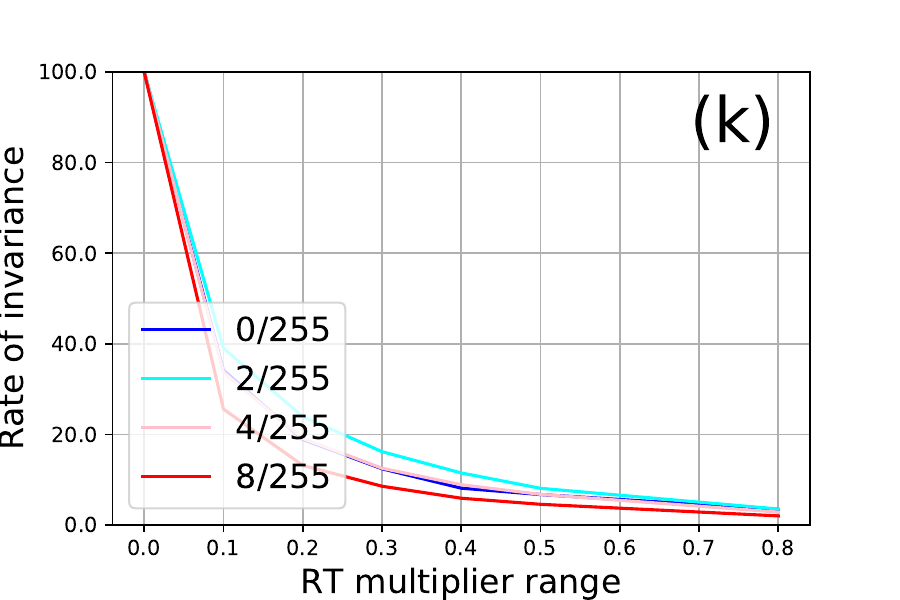}
\end{center}
\vspace{-10pt}
\caption{\textit{(Best viewed in color, zoomed in)} On Tiny-ImageNet for Densenet121 model: (a-b) \textbf{Aug - R} StdCNN, (c-d) \textbf{Aug - T} StdCNN, (e-h) \textbf{Aug - RT} StdCNN, (i-k) \textbf{Adv - PGD} StdCNN, invariance profiles of StdCNN models and corresponding robustness profiles.}
\label{tiny-stdcnn-densenet-full}
\end{figure*}

\vspace{-7pt}
In the main paper, we presented results for experiments carried out using VGG16 on CIFAR10. Please see Figs \ref{tiny-stdcnn-gcnn-resnet-full} and \ref{tiny-stdcnn-densenet-full} for additional experiments on Tiny-ImageNet using the ResNet18 and Densenet121 models, respectively. These complement the experiments in the main paper and show that our results continue to hold across different models and datasets.

\vspace{-7pt}
\section{Results with different PGD hyperparameter settings} \label{suppl:exp-pgd-hyperparameters}
\vspace{-7pt}
The results in the main paper using PGD attack were with the well known setting of $k = 10$ and step size = $2/255$. We checked the adversarial accuracy for the following 4 variants of PGD attack with $\epsilon=8/255$ but different number of steps $k$ and step sizes:
\begin{enumerate}[label=(\alph*)]
\item $k$ = 10, step size = $2/255$ or $0.0078$,
\item $k$ = 20, step size = $2/255$ or $0.0078$, 
\item $k$ = 100, step size = $2/255$ or $0.0078$, 
\item $k$ = 100, step size = $1/255$ or $0.0039$
\end{enumerate}

In Table \ref{pgd-hyperparameters}, we note the results based on the above settings of PGD attack on adversarially trained StdCNN/VGG16 network on CIFAR10 with $0^{\circ}$ and $\pm180^{\circ}$ augmented data. We observe that our conclusions about spatial and adversarial robustness trade-off continue to hold for (a-b-c-d) above.

\begin{table}
\caption{\footnotesize Results similar to Table \ref{tradeoff-table-cifar10-pgd-vgg16} row indexed 3 (PGD (Aug 0)) and row indexed 4 (PGD (Aug 180)) with different PGD hyperparameter settings on StdCNN/VGG16 with CIFAR10.\vspace{0pt}}\vspace{-18pt}
\footnotesize
    \centering
    \begin{tabular}{ l | c | c | c }
    \hline
    \bf Training Method & \bf Adv (PGD) Accuracy(\%) & \bf Std  Accuracy(\%) & \bf Spatial Accuracy(\%) (a-b-c-d)\\
\hline
PGD (Aug 0) & 79.60 & 32.51 & 45.05 - 44.75 - 43.26 - 43.27 \\
PGD (Aug 180) & 54.29 & 53.92 & 33.11 - 32.99 - 32.79 - 32.71 \\
\hline
\end{tabular}
    \label{pgd-hyperparameters}
\end{table}

\vspace{-4pt}
\section{Combining Spatial and Adversarial Training} 
\label{suppl:other-exp}
\begin{figure}[h!]
\begin{center}
\includegraphics[width=0.49\linewidth]{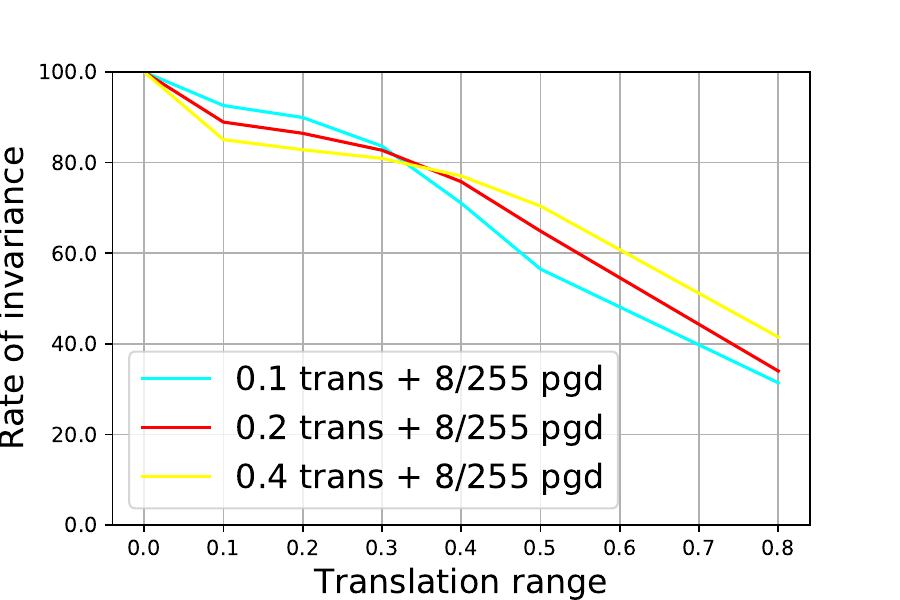}
\includegraphics[width=0.49\linewidth]{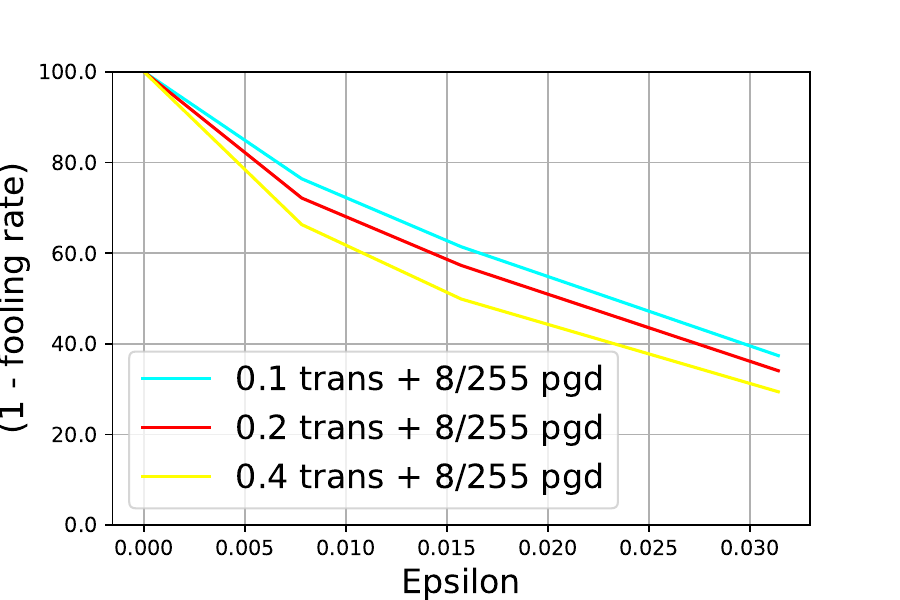}
\end{center}
\vspace{-10pt}
\caption{On CIFAR10, \textbf{Aug - T + Adv - PGD ($\epsilon=8/255$)}; 
\textit{(Left)} Translation invariance profiles and \textit{(Right)} robustness profiles of StdCNN/VGG16 models. Different colored lines represent models initially trained with different translation augmentation to make them translation invariant first. Each model made invariant to different translation ranges was further adversarially trained with $\ell_{\infty}$ budget $\epsilon=8/255$.}
\label{cifar10-stdcnn-trans-pgd}
\end{figure}

\vspace{-6pt}
We additionally performed experiments to study training using spatial augmentations as well as adversarial training. The spatial vs adversarial robustness trade-off continues to hold, e.g., the improvement in spatial robustness that we get by spatial transformation augmentation during training is lost when we do adversarial training to improve adversarial robustness. Fig \ref{cifar10-stdcnn-trans-pgd} shows translation invariance and $\ell_{\infty}$ robustness for StdCNN/VGG16 models that are first trained with progressively larger ranges of translation augmentation on CIFAR10, followed by adversarially training them with a fixed $\ell_{\infty}$ budget of $\epsilon = 8/255$. The results show the same trends as before despite the combined training.

\vspace{-4pt}
\section{Additional Empirical Evidence: Average Perturbation Distance to the Decision Boundary}
\label{suppl:dist-to-bdry}
\begin{figure}[!h]
\begin{center}
\includegraphics[width=0.49\linewidth]{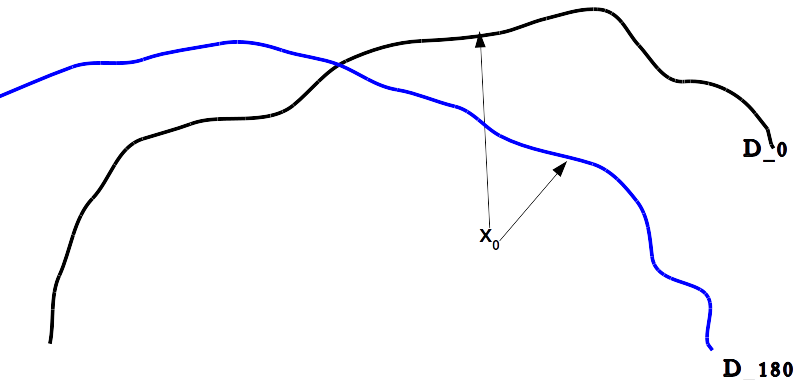}
\end{center}
\vspace{-10pt}
\caption{Distance of point $x_0$ to decision boundary $D_{180}$ obtained by augmenting training set with random rotations in range $[-180^{\circ}, 180^{\circ}]$ is different compared to the decision boundary $D_0$ obtained with no training augmentation. (This figure is only for illustrative purposes, and does not reflect actual measurements.)}
\label{dp-dist-fig}
\end{figure}

\begin{figure*}[!h]
\begin{center}
\includegraphics[width=0.24\linewidth]{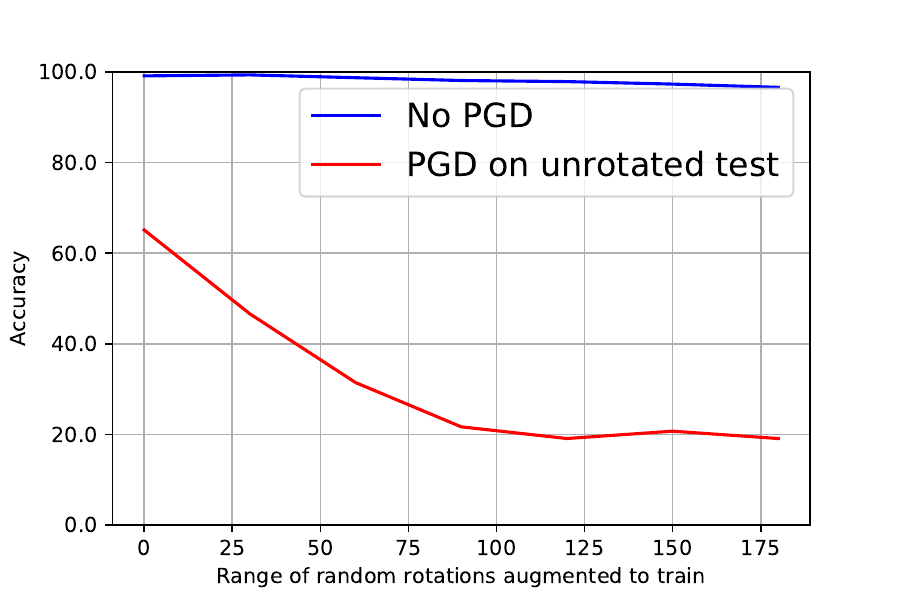}
\includegraphics[width=0.24\linewidth]{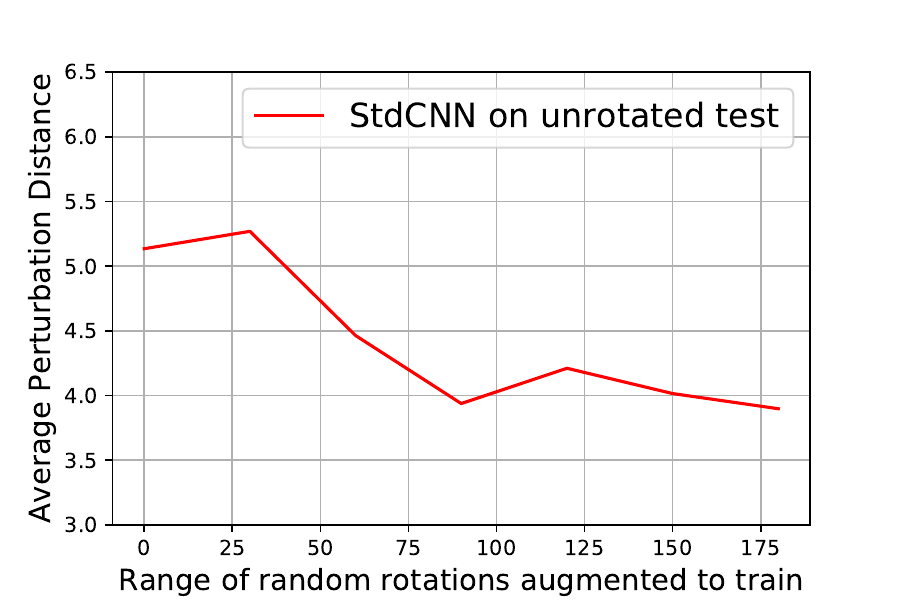}
\rulesep
\includegraphics[width=0.24\linewidth]{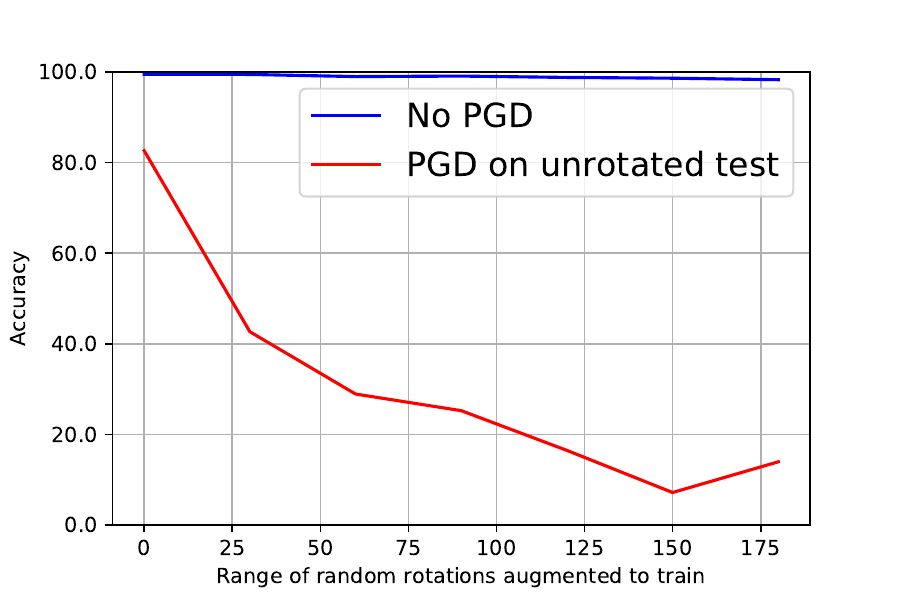}
\includegraphics[width=0.24\linewidth]{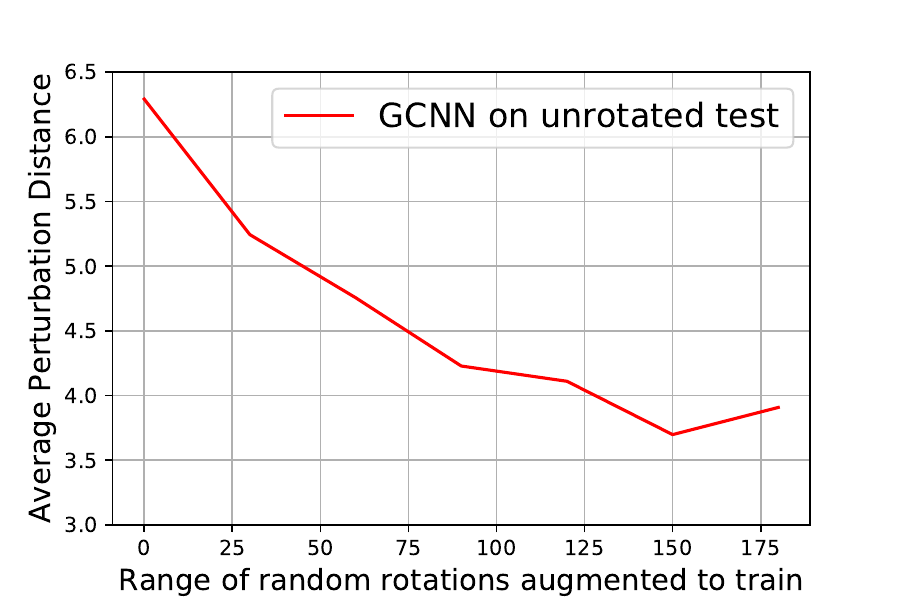}\\
\includegraphics[width=0.24\linewidth]{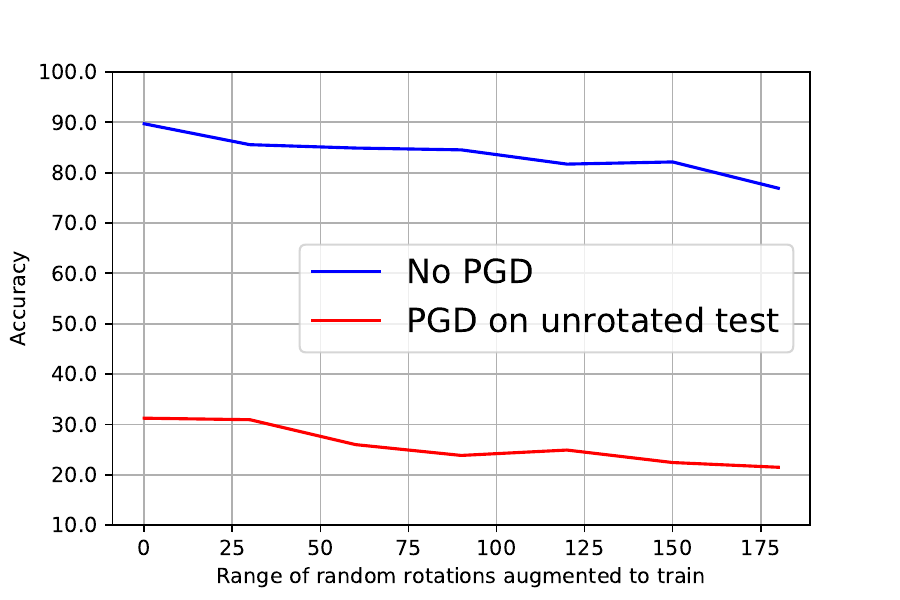}
\includegraphics[width=0.24\linewidth]{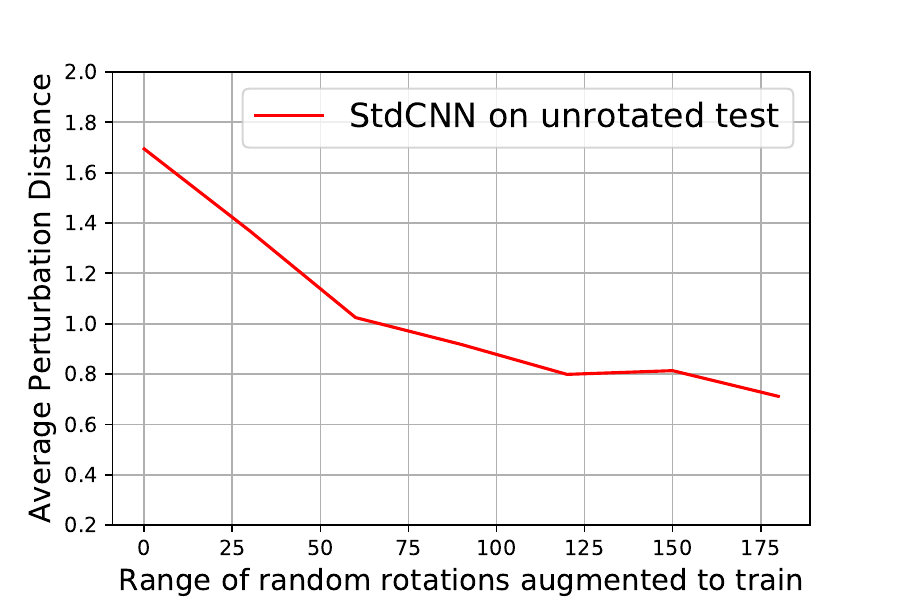}
\rulesep
\includegraphics[width=0.24\linewidth]{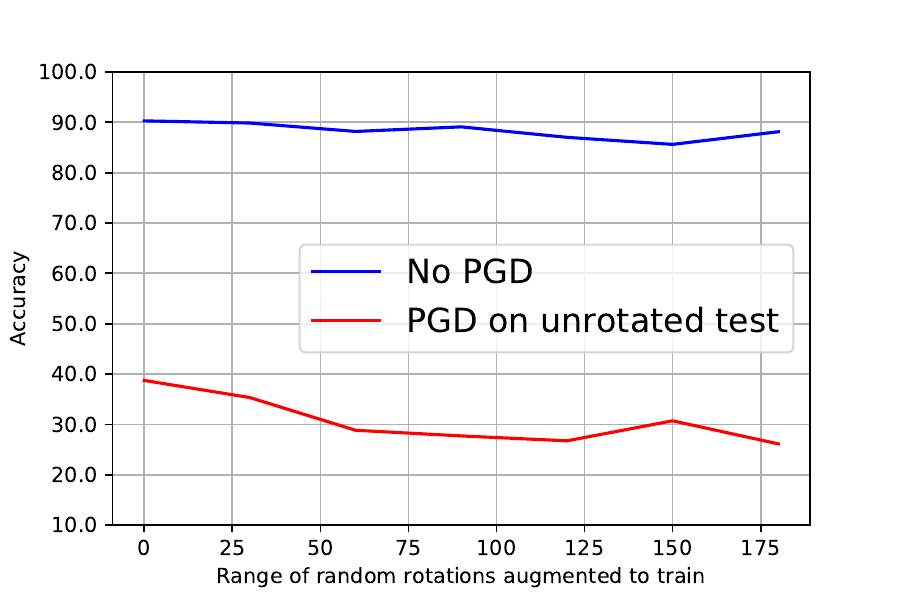}
\includegraphics[width=0.24\linewidth]{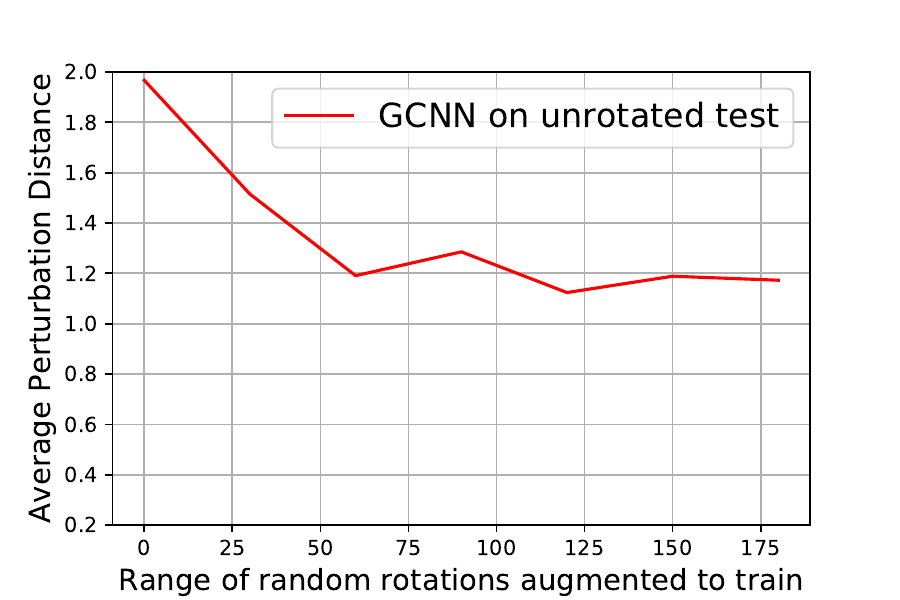}\\
\includegraphics[width=0.24\linewidth]{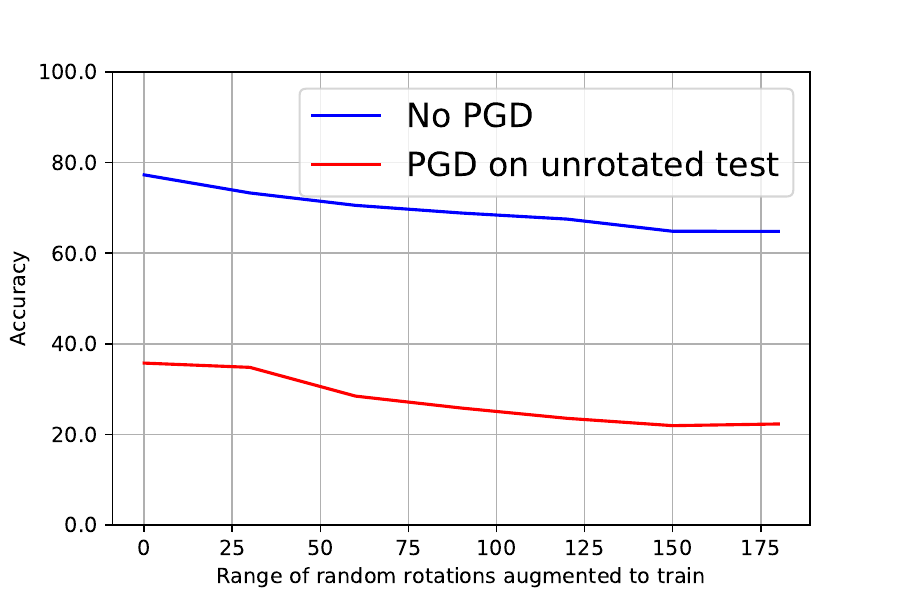}
\includegraphics[width=0.24\linewidth]{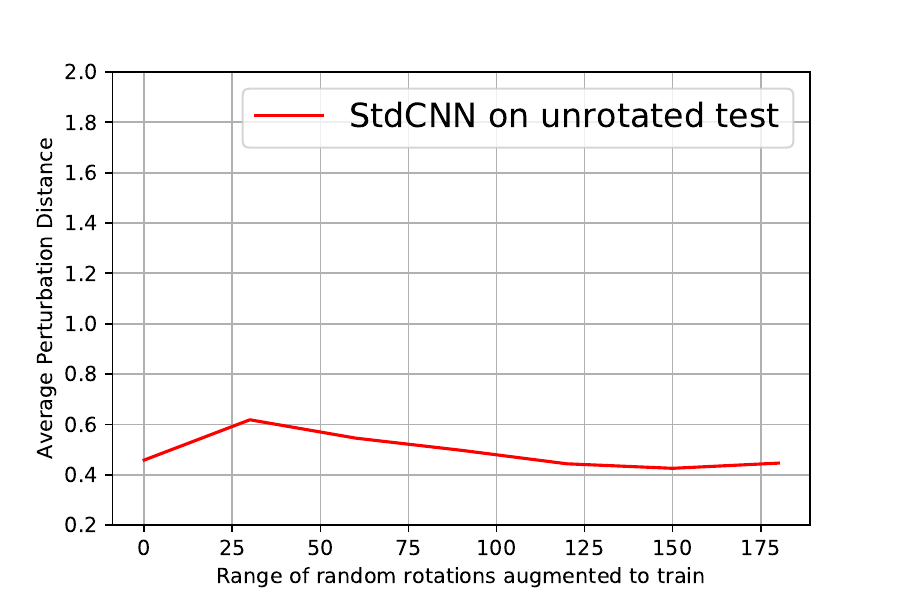}
\rulesep
\includegraphics[width=0.24\linewidth]{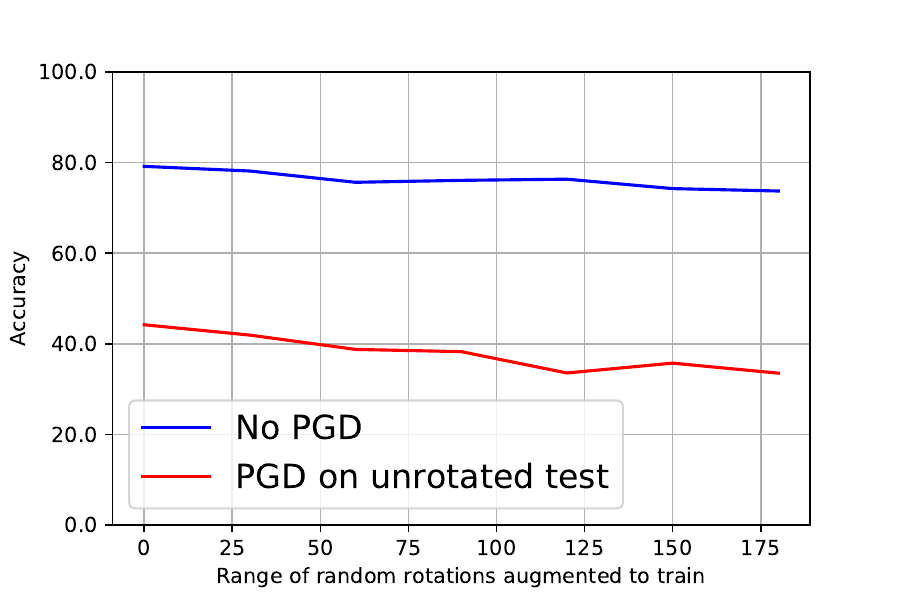}
\includegraphics[width=0.24\linewidth]{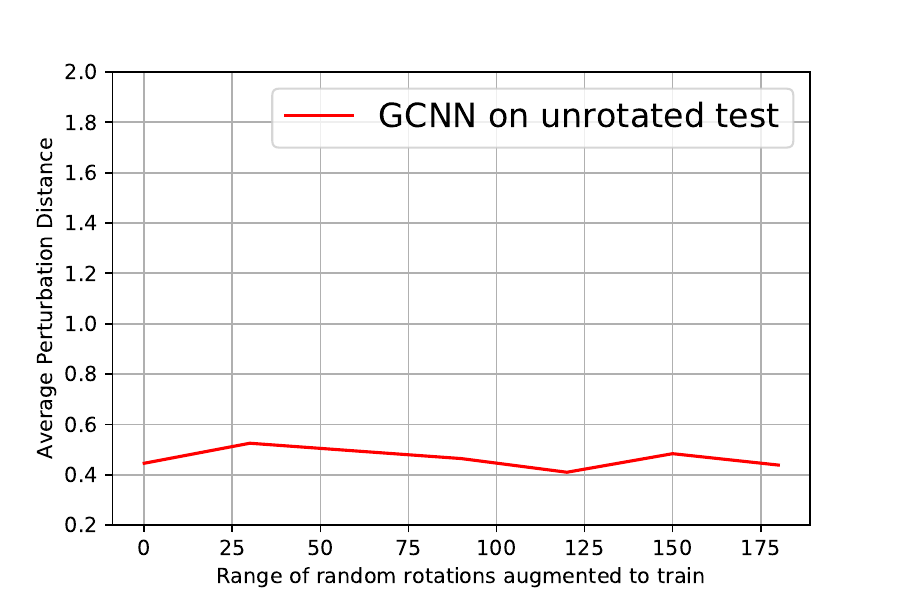}
\end{center}
\vspace{-10pt}
\caption{\textit{(Best viewed in color, zoomed in)} Accuracy of StdCNN/GCNN on MNIST(Table\ref{gcnn-table})/CIFAR10(VGG16)/CIFAR100(ResNet18) with/without PGD ($\epsilon = 0.3, \epsilon = 8/255, \epsilon = 2/255$), on unrotated test. Train/test if augmented are with random rotations in $[-\theta^{\circ}, \theta^{\circ}]$.  (a-d) MNIST, (e-h) CIFAR10, (i-l) CIFAR100, with each pair of plots eg. (a) being the accuracy while (b) being the corresponding avg. perturbation distance.}
\label{stdcnn-gcnn-pgd-norot-rot-dist-fig}
\end{figure*}

\vspace{-6pt}
For each test image, adversarial attacks find perturbations of the test point with small $\ell_{\infty}$ norm that would change the prediction of the given model. Most adversarial attacks do so by finding the directions in which the loss function of the model changes the most.  In order to explain why these networks become vulnerable to pixel-wise attacks as they learn more rotations, we see how the distance of the test points to the decision boundary changes as the networks learn larger rotations. This is illustrated in Fig \ref{dp-dist-fig} where we show the distance of a test point $x_0$ to the boundary $D_0$ (resp. $D_{180}$) when the model is trained with zero (resp. $\pm180^{\circ}$) rotations.  
We use the $L_2$ attack vectors obtained by DeepFool \cite{Dezfooli16} for the datapoints under attack. We take the norm of this attack vector as an approximation to the shortest distance of the test point to the decision boundary. For each of the test points we collect the perturbation vectors given by DeepFool attack and report the average perturbation distance. We plot this average distance as the datasets are augmented with larger rotations. 
Our experiments show that as the networks learn larger rotations with augmentation, the average perturbation distance falls. So as (symmetric) networks become invariant to rotations, they are more vulnerable to pixel-wise attacks.
The plots in Fig \ref{stdcnn-gcnn-pgd-norot-rot-dist-fig} show this for StdCNNs and GCNNs on MNIST, CIFAR10 and CIFAR100. 
We plot the accuracy of these models and also the average perturbation distance of the test points alongside in one figure, e.g. \ref{stdcnn-gcnn-pgd-norot-rot-dist-fig}(a) shows accuracy and \ref{stdcnn-gcnn-pgd-norot-rot-dist-fig}(b) shows corresponding average perturbation distance. The blue line in Fig \ref{stdcnn-gcnn-pgd-norot-rot-dist-fig}(a) shows the accuracy of a StdCNN on MNIST when the training data is augmented with $\theta$ ranging from $0$ to $180$ and the test has no augmentation. The red line shows the accuracy when the test is not augmented with rotations but is PGD perturbed with $\ell_{\infty}$ norm $0.3$.
The red line in Fig \ref{stdcnn-gcnn-pgd-norot-rot-dist-fig}(b) shows the average perturbation distance of the unrotated test points when the network is trained with rotations upto $\theta$ - this is about 5 when $\theta$ is $0^{\circ}$ (the point on $y$-axis where curves begin). As the network is trained with random rotations up to $180^{\circ}$, the average perturbation distance of the augmented test drops below 4.0. Fig \ref{stdcnn-gcnn-pgd-norot-rot-dist-fig}(a) shows that that the PGD accuracy has dropped from around 65\% for the network at $0^{\circ}$ to 20\% at $180^{\circ}$.

\vspace{-4pt}
\section{Curriculum Learning-based Approach for Pareto-Optimality: Additional Results}
\label{suppl:pareto-optimal}
\vspace{-6pt}

\begin{figure*}[h]
    \centering
    \includegraphics[scale=0.45]{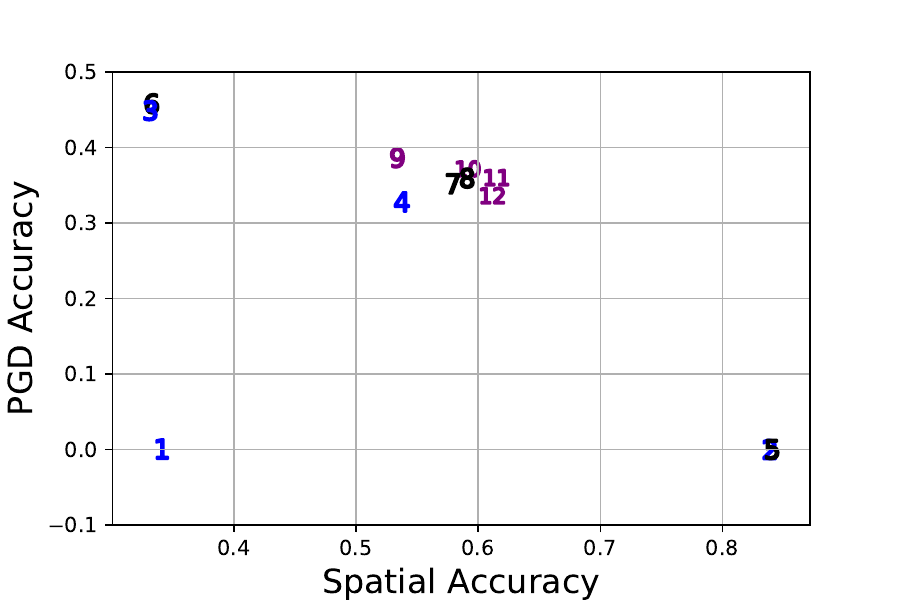}
    \includegraphics[scale=0.45]{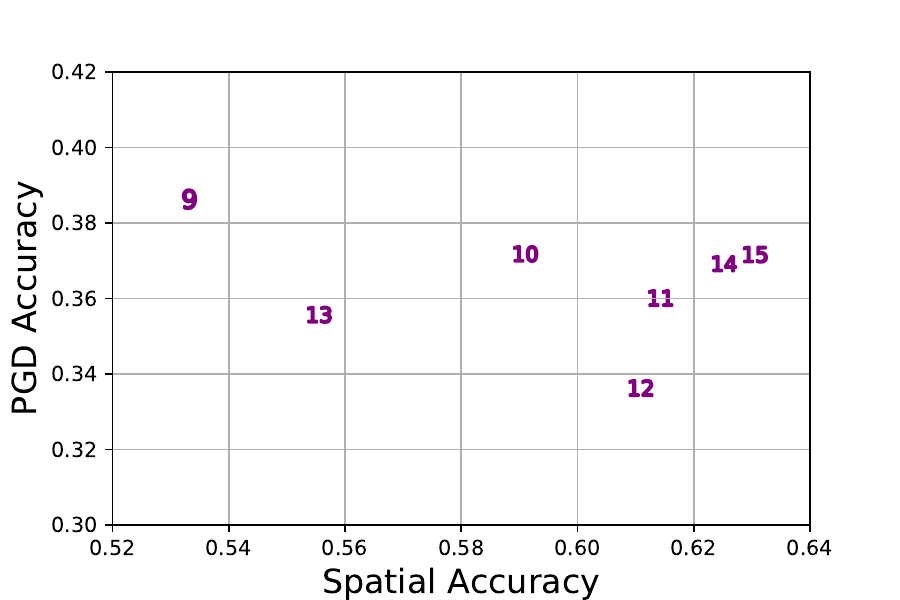}
    \caption{Visualization of performance of CuSP based on PGD against other baseline strategies on CIFAR10 for StdCNN/VGG16 model (each index corresponds to a row in Table \ref{tradeoff-table-cifar10-pgd-trades-vgg16}). On the left, we compare CuSP against various baselines. On the right, we zoom in to compare different variants of CuSP.}
    \label{sample-trade-off-cifar10-vgg16-pgd-only}
\end{figure*}

\begin{figure*}[h]
    \centering
    \includegraphics[scale=0.45]{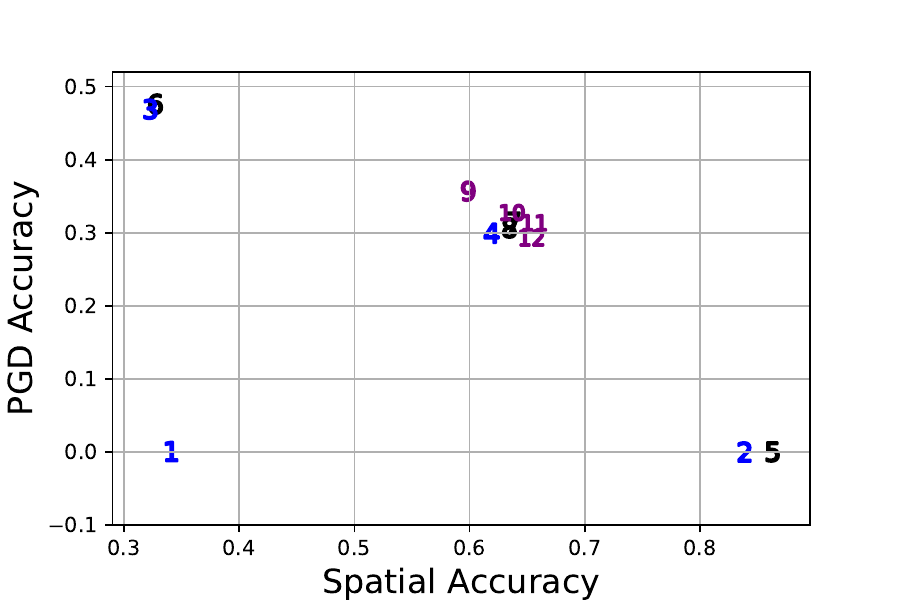}
    \includegraphics[scale=0.45]{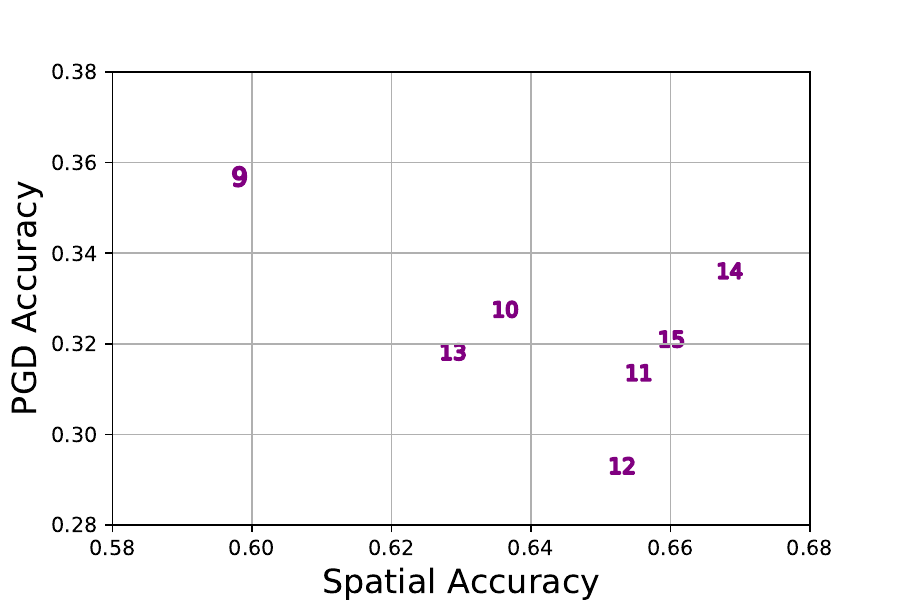}
    \caption{Visualization of performance of CuSP based on TRADES against other baseline strategies on CIFAR10 for StdCNN/VGG16 model (each index corresponds to a row in Table \ref{tradeoff-table-cifar10-pgd-trades-vgg16}). On the left, we compare CuSP against various baselines. On the right, we zoom in to compare different variants of CuSP.}
    \label{sample-trade-off-cifar10-vgg16-trades-only}
\end{figure*}

\begin{figure}[h!]
\begin{center}
    \includegraphics[scale=0.45]{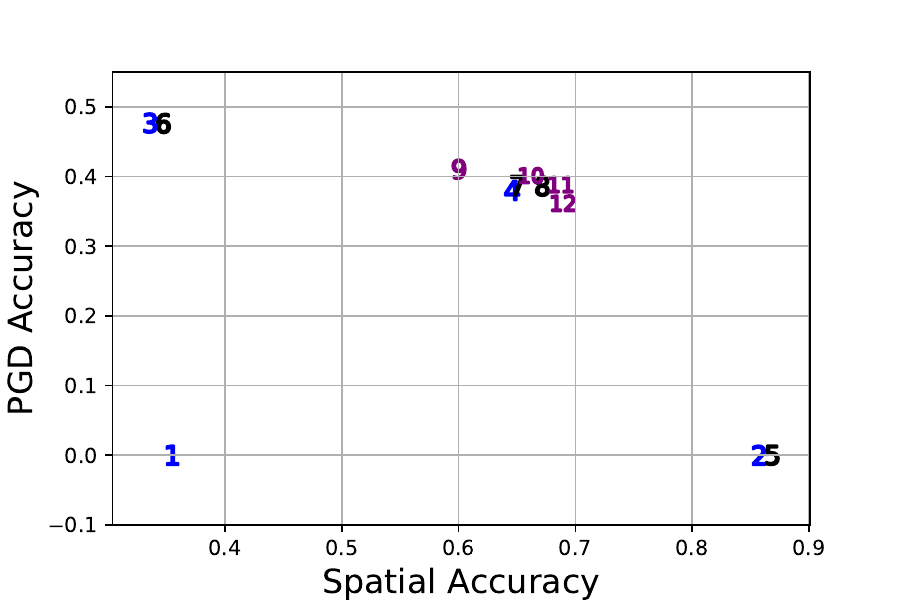}
    \includegraphics[scale=0.45]{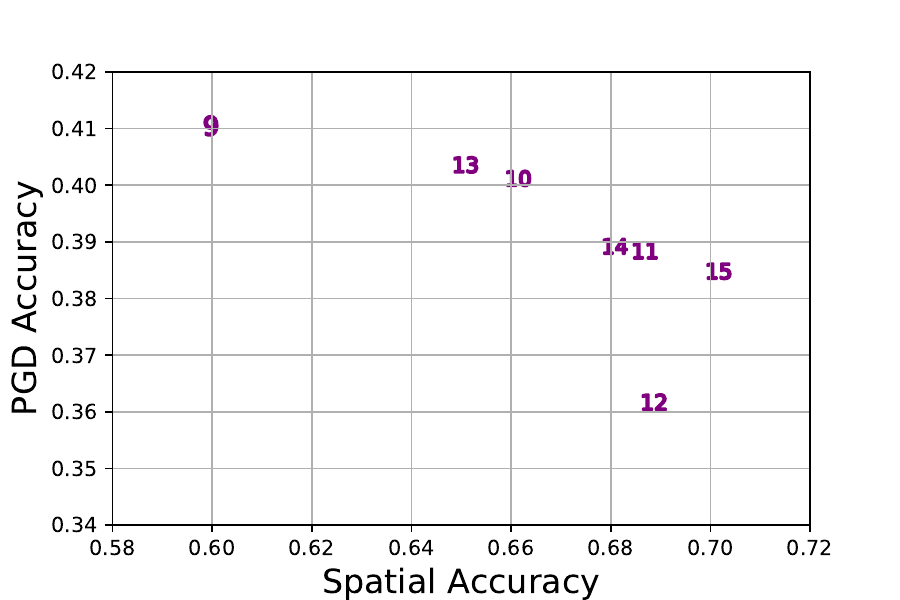}
\end{center}
\vspace{-10pt}
    \caption{Visualization of performance of CuSP based on PGD against other baseline strategies on CIFAR10 for StdCNN/ResNet18 model (each index corresponds to a row in Table \ref{tradeoff-table-cifar10-pgd-trades-resnet18}). On the left, we compare CuSP against various baselines. On the right, we zoom in to compare different variants of CuSP.}
    \label{sample-trade-off-cifar10-resnet18-pgd-only}
\end{figure}

\begin{figure}[h!]
\begin{center}
    \includegraphics[scale=0.45]{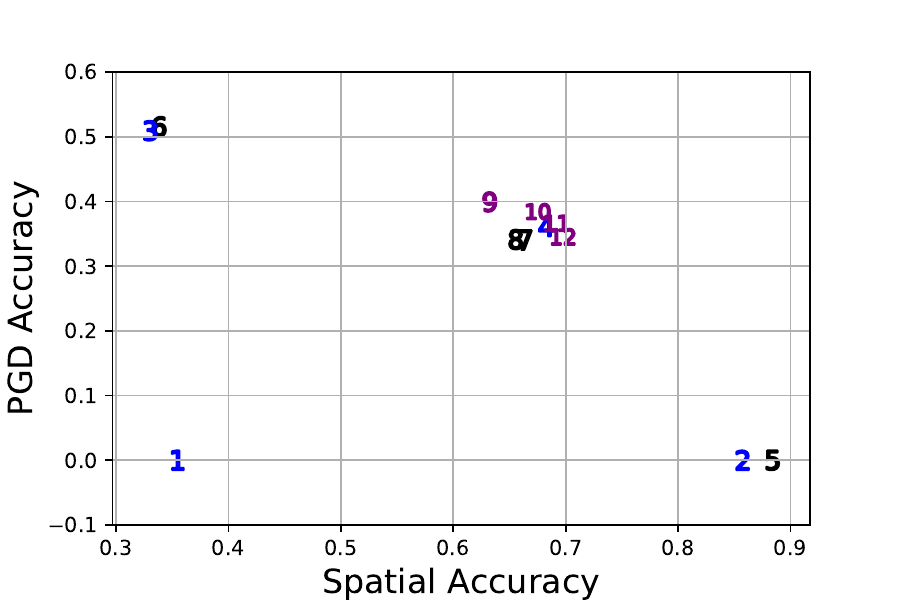}
    \includegraphics[scale=0.45]{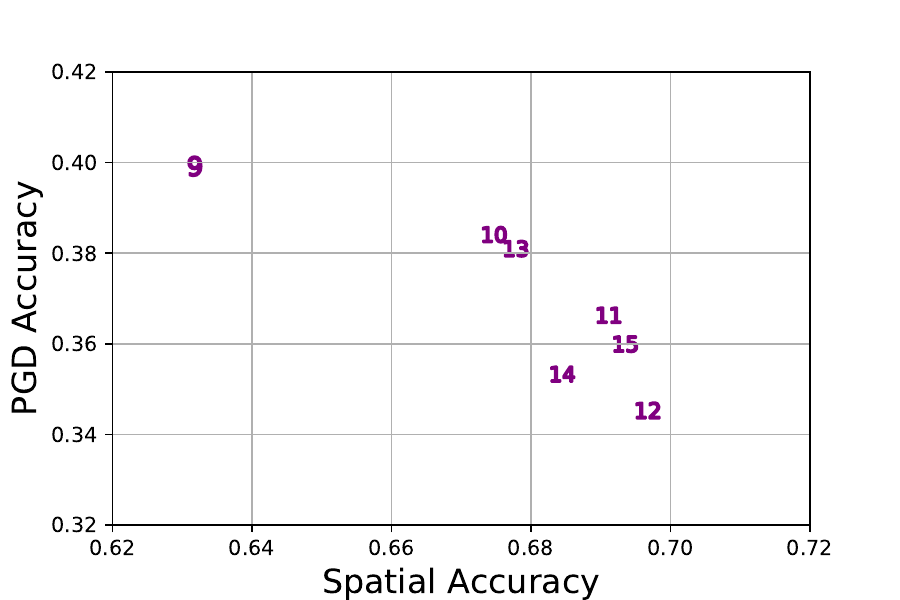}
\end{center}
\vspace{-10pt}
    \caption{Visualization of performance of CuSP based on TRADES against other baseline strategies on CIFAR10 for StdCNN/ResNet18 model (each index corresponds to a row in Table \ref{tradeoff-table-cifar10-pgd-trades-resnet18}). On the left, we compare CuSP against various baselines. On the right, we zoom in to compare different variants of CuSP.}
    \label{sample-trade-off-cifar10-resnet18-trades-only}
\end{figure}

\begin{figure}[h!]
    \centering
    \includegraphics[scale=0.45]{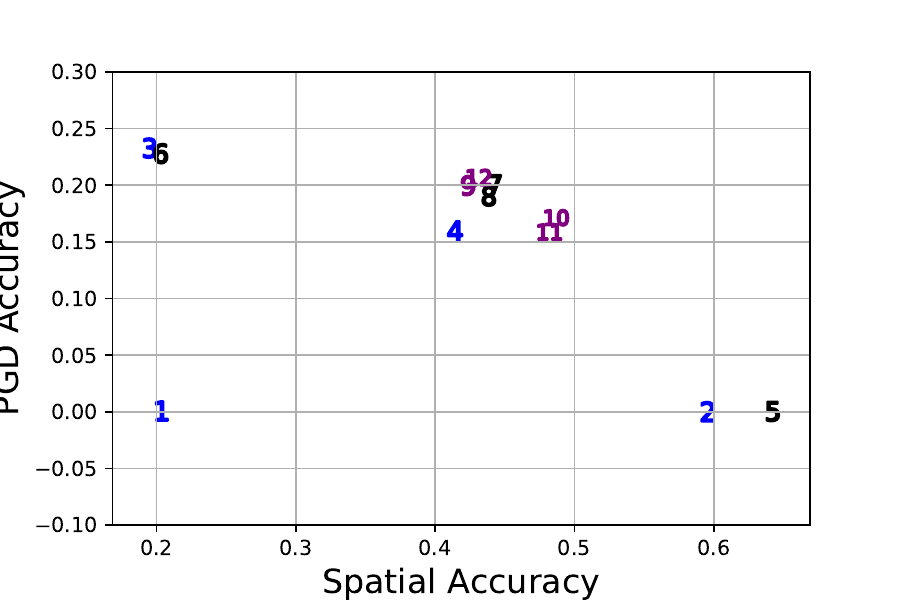}
    \includegraphics[scale=0.45]{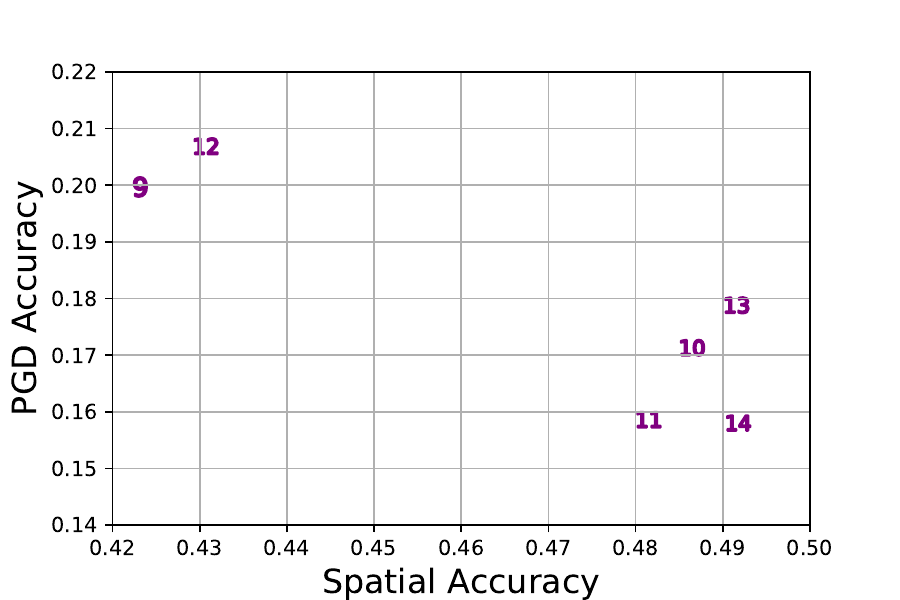}
    \caption{Visualization of performance of CuSP based on PGD against other baseline strategies on CIFAR100 for StdCNN/ResNet18 model (each index corresponds to a row in Table \ref{tradeoff-table-cifar100-pgd-trades-resnet18}). On the left, we compare CuSP against various baselines. On the right, we zoom in to compare different variants of CuSP.}
    \label{sample-trade-off-cifar100-resnet18-pgd-only}
\end{figure}

\begin{figure}[h!]
    \centering
    \includegraphics[scale=0.45]{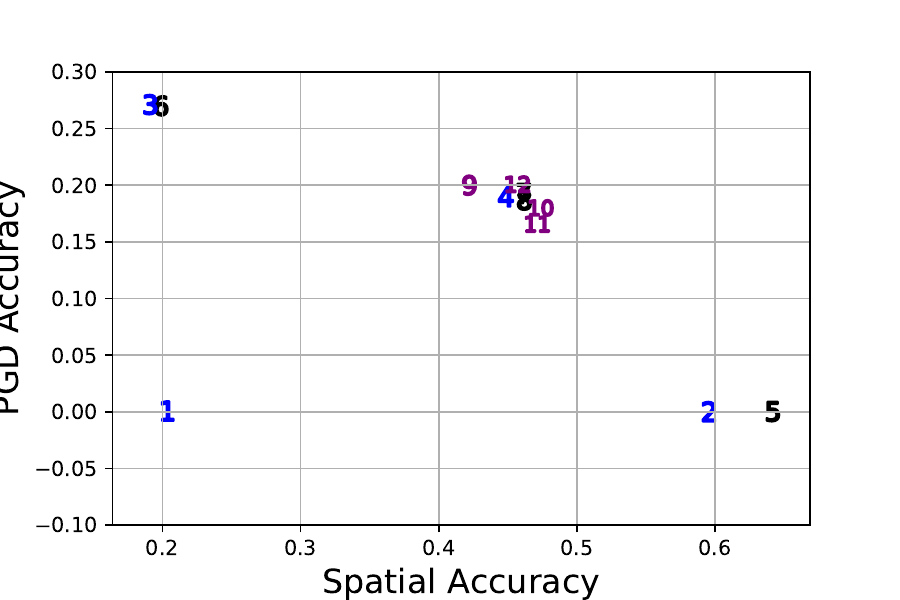}
    \includegraphics[scale=0.45]{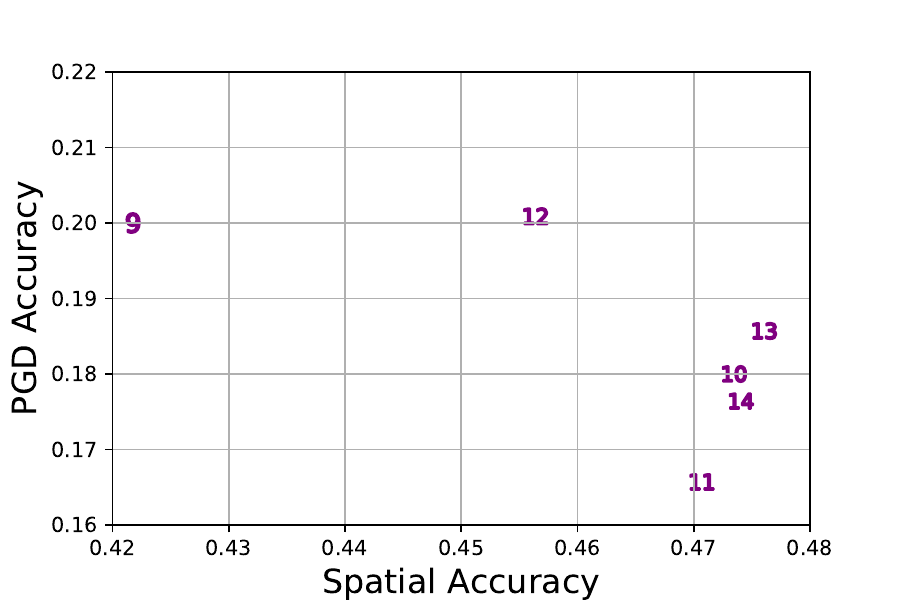}
    \caption{Visualization of performance of CuSP based on TRADES against other baseline strategies on CIFAR100 for StdCNN/ResNet18 model (each index corresponds to a row in Table \ref{tradeoff-table-cifar100-pgd-trades-resnet18}). On the left, we compare CuSP against various baselines. On the right, we zoom in to compare different variants of CuSP.}
    \label{sample-trade-off-cifar100-resnet18-trades-only}
\end{figure}

In the main paper, we presented results of CuSP based on PGD for the CIFAR10 dataset trained with StdCNN/VGG16 architecture. 
In this section, we provide additional experimental results and visualizations based on CuSP with PGD and TRADES on StdCNN/ResNet18 for the CIFAR10 and CIFAR100 datasets. We also include additional results using TRADES on StdCNN/VGG16 for the CIFAR10 dataset. Tables \ref{tradeoff-table-cifar10-pgd-trades-vgg16} and \ref{tradeoff-table-cifar10-pgd-trades-resnet18} contain the results for StdCNN/VGG16 and StdCNN/ResNet18 on CIFAR10, respectively. (Note that Table \ref{tradeoff-table-cifar10-pgd-trades-vgg16} is a more detailed version of Table 1 in the main paper). Table \ref{tradeoff-table-cifar100-pgd-trades-resnet18} presents the results for StdCNN/ResNet18 on CIFAR100. Similar to Fig 6 in the main paper, Figs \ref{sample-trade-off-cifar10-vgg16-pgd-only}, \ref{sample-trade-off-cifar10-vgg16-trades-only}, \ref{sample-trade-off-cifar10-resnet18-pgd-only}, \ref{sample-trade-off-cifar10-resnet18-trades-only}, \ref{sample-trade-off-cifar100-resnet18-pgd-only} and \ref{sample-trade-off-cifar100-resnet18-trades-only} present a visualization of the results to understand their relevance to the Pareto frontier.

\uline{\textit{Observations:}} In Table \ref{tradeoff-table-cifar10-pgd-trades-vgg16} and \ref{tradeoff-table-cifar10-pgd-trades-resnet18}, the rows indexed 5 and 6 indicate the training strategy of sequentially combining adversarial training and augmentation by spatial transformations (as in Table 1 of the main paper), i.e., take an adversarially trained network and re-train it with spatial (e.g., rotation) augmentation, or vice versa. It is clear from the visualizations in Figs \ref{sample-trade-off-cifar10-vgg16-pgd-only}, \ref{sample-trade-off-cifar10-vgg16-trades-only}, \ref{sample-trade-off-cifar10-resnet18-pgd-only} and \ref{sample-trade-off-cifar10-resnet18-trades-only} that they produce models which lie either bottom-right of the plot (high spatial accuracy but low adversarial robustness) or top-left of the plot (high adversarial robustness but low spatial accuracy). These are examples of the version of catastrophic forgetting displayed by these networks, as discussed earlier. Another viable baseline for our study is adversarial training with data augmented with rotations in the range $[-180^{\circ}, +180^{\circ}]$. 
We observe from these results that CuSP with various configurations produces a model with both spatial invariance and adversarial robustness, in comparison to the baselines. While rows indexed 4 and 6 across the tables seem to show promise, \textit{PGD(Aug 180)} expectedly suffers from relatively poorer adversarial accuracy, while \textit{Aug 180} $\rightarrow$ \textit{PGD (Aug 0)} expectedly suffers from poor spatial accuracy. This is clearly a case of the aforementioned catastrophic forgetting issue, as the latter has a behavior similar to row indexed 3 \textit{PGD(Aug 0)}. 
The proposed method, CuSP, consistently outperforms these methods across the datasets and models -- supporting the role of curriculum learning for simultaneous robustness on multiple fronts.  

\uline{\textit{Other Settings and Extensions:}} While the main set of experiments were based on TRADES learning rate schedule 75-90-100, we also tried other schedules like 40-80 (40-th and 80-th epochs) and 50-100. Table \ref{tradeoff-table-cifar10-vgg16-other} shows that these also obtain competent results, hence indicating that CuSP can be further finetuned to obtain better Pareto-optimal solutions, which is a promising direction of future work.

\begin{table}[!h]
\caption{\footnotesize Comparison of performance of proposed CuSP using StdCNN/VGG16 with other baseline strategies on CIFAR10. (Aug $\theta$): denotes training data augmented with random rotations in the range $[-\theta, +\theta]$; $a \rightarrow b$: denotes $a$ sequentially followed by $b$ during training.}
\footnotesize
    \centering
    \begin{tabular}{c | l | c  c  c }
    \hline
    &\bf Training Method&\bf Adv (PGD)&\bf Std &\bf Spatial\\
    &&\bf Accuracy(\%)&\bf Accuracy(\%)&\bf Accuracy(\%)\\
\hline
1 & Natural (Aug 0) & 00.05 & 92.89 & 33.99 \\
2 & Natural (Aug 180) & 00.00 & 85.00 & 83.78 \\
3 & PGD (Aug 0) & 44.88 & 80.08 & 33.07 \\
4 & PGD (Aug 180) & 32.79 & 54.60 & 53.69 \\
\hline
5 & PGD (Aug 0) $\rightarrow$ Aug 180 & 00.00 & 85.46 & 83.99 \\
6 & Aug 180 $\rightarrow$ PGD (Aug 0) & 45.80 & 81.03 & 33.19 \\
7 & PGD (Aug 0) $\rightarrow$ PGD (Aug 180) & 35.19 & 58.96 & 57.84 \\
8 & Aug 180 $\rightarrow$ PGD (Aug 180) & 35.94 & 60.27 & 59.02 \\
\hline
9 & CuSP (PGD, 30-60-180, $\{\frac{2}{255}, \frac{4}{255}, \frac{8}{255}\}$) & 38.63 & 65.92 & 53.31 \\
10 & CuSP (PGD, 60-120-180, $\{\frac{2}{255}, \frac{4}{255}, \frac{8}{255}\}$) & 37.18 & 65.88 & 59.09 \\
11 & CuSP (PGD, 120-150-180,  $\{\frac{2}{255}, \frac{4}{255}, \frac{8}{255}\}$) & {36.01} & {64.96} & {61.41} \\
12 & CuSP (PGD, 180, $\{\frac{2}{255}, \frac{4}{255}, \frac{8}{255}\}$) & 33.63 & 62.17 & 61.07 \\
13 & CuSP (PGD, 120-150-180,$\{\frac{8}{255}\}$) & 35.57 & 58.08 & 55.54 \\
14 & PGD (Aug 0) $\rightarrow$ & & & \\
 & CuSP (PGD, 120-150-180,  $\{\frac{2}{255}, \frac{4}{255}, \frac{8}{255}\}$) & 36.92 & 66.25 & 62.51 \\
15 & Aug 180 $\rightarrow$ & & & \\
 & Aug 180 $\rightarrow$ CuSP (PGD, 120-150-180,  $\{\frac{2}{255}, \frac{4}{255}, \frac{8}{255}\}$) & 37.16 & 66.58 & 63.04 \\
\hline
\hline
1 & Natural (Aug 0) & 00.05 & 92.89 & 33.99 \\
2 & Natural (Aug 180) & 00.00 & 85.00 & 83.78 \\
3 & TRADES (Aug 0) & 46.90 & 79.35 & 32.20 \\
4 & TRADES (Aug 180) & 29.93 & 62.38 & 61.85 \\
\hline
5 & TRADES (Aug 0) $\rightarrow$ Aug 180 & 00.00 & 87.82 & 86.22 \\
6 & Aug 180 $\rightarrow$ TRADES (Aug 0) & 47.61 & 79.90 & 32.65 \\
7 & TRADES (Aug 0) $\rightarrow$ TRADES (Aug 180) & 31.43 & 63.77 & 63.70 \\
8 & Aug 180 $\rightarrow$ TRADES (Aug 180) & 30.65 & 63.37 & 63.38 \\
\hline
9 & CuSP (TRADES, 30-60-180, $\{\frac{2}{255}, \frac{4}{255}, \frac{8}{255}\}$) & 35.69 & 71.47 & 59.81 \\
10 & CuSP (TRADES, 60-120-180, $\{\frac{2}{255}, \frac{4}{255}, \frac{8}{255}\}$) & 32.76 & 69.80 & 63.62 \\
11 & CuSP (TRADES, 120-150-180,  $\{\frac{2}{255}, \frac{4}{255}, \frac{8}{255}\}$) & {31.36} & {68.04} & { 65.53} \\
12 & CuSP (TRADES, 180, $\{\frac{2}{255}, \frac{4}{255}, \frac{8}{255}\}$) & 29.31 & 66.14 & 65.29 \\
13 & CuSP (TRADES, 120-150-180,$\{\frac{8}{255}\}$) & 31.82 & 65.17 & 62.87 \\
14 & TRADES (Aug 0) $\rightarrow$ & & & \\
 & CuSP (TRADES, 120-150-180,  $\{\frac{2}{255}, \frac{4}{255}, \frac{8}{255}\}$) & 33.61 & 69.56 & 66.84 \\
15 & Aug 180 $\rightarrow$ & & & \\
 & Aug 180 $\rightarrow$ CuSP (TRADES, 120-150-180,  $\{\frac{2}{255}, \frac{4}{255}, \frac{8}{255}\}$) & 32.10 & 69.29 & 66.00 \\
\hline
\end{tabular}
    \label{tradeoff-table-cifar10-pgd-trades-vgg16}
\end{table}

\begin{table}[!h]
\caption{\footnotesize Comparison of performance of proposed CuSP using StdCNN/ResNet18 with other baseline strategies on CIFAR10. (Aug $\theta$): denotes training data augmented with random rotations in the range $[-\theta, +\theta]$; $a \rightarrow b$: denotes $a$ sequentially followed by $b$ during training.}
\footnotesize
    \centering
    \begin{tabular}{c | l | c  c  c }
    \hline
    &\bf Training Method&\bf Adv (PGD)&\bf Std &\bf Spatial\\
    &&\bf Accuracy(\%)&\bf Accuracy(\%)&\bf Accuracy(\%)\\
\hline
1 & Natural (Aug 0) & 00.00 & 94.44 & 35.35 \\
2 & Natural (Aug 180) & 00.00 & 86.99 & 85.65 \\
3 & PGD (Aug 0) & 47.67 & 83.72 & 33.50 \\
4 & PGD (Aug 180) & 38.00 & 65.30 & 64.53 \\
\hline
5 & PGD (Aug 0) $\rightarrow$ Aug 180 & 00.00 & 87.90 & 86.80 \\
6 & Aug 180 $\rightarrow$ PGD (Aug 0) & 47.62 & 83.73 & 34.63 \\
7 & PGD (Aug 0) $\rightarrow$ PGD (Aug 180) & 38.70 & 65.88 & 64.94 \\
8 & Aug 180 $\rightarrow$ PGD (Aug 180) & 38.62 & 67.79 & 67.10  \\
\hline
9 & CuSP (PGD, 30-60-180, $\{\frac{2}{255}, \frac{4}{255}, \frac{8}{255}\}$) & 41.05 & 71.66 & 59.96 \\
10 & CuSP (PGD, 60-120-180, $\{\frac{2}{255}, \frac{4}{255}, \frac{8}{255}\}$) & 40.12 & 71.57 & 66.13 \\
11 & CuSP (PGD, 120-150-180,  $\{\frac{2}{255}, \frac{4}{255}, \frac{8}{255}\}$) & {38.84} & {71.31} & {68.67} \\
12 & CuSP (PGD, 180, $\{\frac{2}{255}, \frac{4}{255}, \frac{8}{255}\}$) & 36.17 & 69.65 & 68.85 \\
13 & CuSP (PGD, 120-150-180,$\{\frac{8}{255}\}$) & 40.36 & 67.32 & 65.07 \\
14 & PGD (Aug 0) $\rightarrow$ & & & \\
 & PGD (Aug 0) $\rightarrow$ CuSP (PGD, 120-150-180,  $\{\frac{2}{255}, \frac{4}{255}, \frac{8}{255}\}$) & 38.92 & 70.65 & 68.07 \\
15 & Aug 180 $\rightarrow$ & & & \\
 & CuSP (PGD, 120-150-180,  $\{\frac{2}{255}, \frac{4}{255}, \frac{8}{255}\}$) & 38.48 & 72.27 & 70.15 \\
\hline
\hline
1 & Natural (Aug 0) & 00.00 & 94.44 & 35.35 \\
2 & Natural (Aug 180) & 00.00 & 86.99 & 85.65 \\
3 & TRADES (Aug 0) & 50.92 & 81.83 & 32.96 \\
4 & TRADES (Aug 180) & 36.13 & 68.15 & 68.22 \\
\hline
5 & TRADES (Aug 0) $\rightarrow$ Aug 180 & 00.00 & 89.46 & 88.33 \\
6 & Aug 180 $\rightarrow$ TRADES (Aug 0) & 51.55 & 82.39 & 33.75 \\
7 & TRADES (Aug 0) $\rightarrow$ TRADES (Aug 180) & 34.04 & 66.13 & 66.29 \\
8 & Aug 180 $\rightarrow$ TRADES (Aug 180) & 34.11 & 65.67 & 65.48 \\
\hline
9 & CuSP (TRADES, 30-60-180, $\{\frac{2}{255}, \frac{4}{255}, \frac{8}{255}\}$) & 39.92 & 73.79 & 63.17 \\
10 & CuSP (TRADES, 60-120-180, $\{\frac{2}{255}, \frac{4}{255}, \frac{8}{255}\}$) & 38.41 & 73.16 & 67.46 \\
11 & CuSP (TRADES, 120-150-180,  $\{\frac{2}{255}, \frac{4}{255}, \frac{8}{255}\}$) & {36.63} & {71.76} & {69.10} \\
12 & CuSP (TRADES, 180, $\{\frac{2}{255}, \frac{4}{255}, \frac{8}{255}\}$) & 34.53 & 70.19 & 69.66 \\
13 & CuSP (TRADES, 120-150-180,$\{\frac{8}{255}\}$) & 38.10 & 69.97 & 67.77 \\
14 & TRADES (Aug 0) $\rightarrow$ & & & \\
 & CuSP (TRADES, 120-150-180,  $\{\frac{2}{255}, \frac{4}{255}, \frac{8}{255}\}$) & 35.33 & 70.64 & 68.44 \\
15 & Aug 180 $\rightarrow$ & & & \\
 & CuSP (TRADES, 120-150-180,  $\{\frac{2}{255}, \frac{4}{255}, \frac{8}{255}\}$) & 35.99 & 71.37 & 69.34 \\
\hline
\end{tabular}
    \label{tradeoff-table-cifar10-pgd-trades-resnet18}
\end{table}

\begin{table}[!h]
\caption{\footnotesize Comparison of performance of proposed CuSP using StdCNN/ResNet18 with other baseline strategies on CIFAR100. (Aug $\theta$): denotes training data augmented with random rotations in the range $[-\theta, +\theta]$; $a \rightarrow b$: denotes $a$ sequentially followed by $b$ during training.}
\footnotesize
    \centering
    \begin{tabular}{c | l | c  c  c }
    \hline
    &\bf Training Method&\bf Adv (PGD)&\bf Std &\bf Spatial\\
    &&\bf Accuracy(\%)&\bf Accuracy(\%)&\bf Accuracy(\%)\\
\hline
1 & Natural (Aug 0) & 00.06 & 74.93 & 20.27 \\
2 & Natural (Aug 180) & 00.01 & 58.96 & 59.49 \\
3 & PGD (Aug 0) & 23.28 & 56.27 & 19.44 \\
4 & PGD (Aug 180) & 15.96 & 41.48 & 41.39 \\
\hline
5 & PGD (Aug 0) $\rightarrow$ Aug 180 & 00.02 & 64.24 & 64.17 \\
6 & Aug 180 $\rightarrow$ PGD (Aug 0) & 22.79 & 56.48 & 20.24 \\
7 & PGD (Aug 0) $\rightarrow$ PGD (Aug 180) & 20.03 & 44.24 & 44.19 \\
8 & Aug 180 $\rightarrow$ PGD (Aug 180) & 19.04 & 43.82 & 43.79 \\
\hline
9 & CuSP (PGD, 30-60-180, $\{\frac{2}{255}, \frac{4}{255}, \frac{8}{255}\}$) & 19.97 & 51.35 & 42.31 \\
10 & CuSP (PGD, 120-150-180,  $\{\frac{2}{255}, \frac{4}{255}, \frac{8}{255}\}$) & {17.13} & {49.80} & {48.64} \\
11 & CuSP (PGD, 180, $\{\frac{2}{255}, \frac{4}{255}, \frac{8}{255}\}$) & 15.86 & 48.34 & 48.14 \\
12 & CuSP (PGD, 120-150-180,$\{\frac{8}{255}\}$) & 20.69 & 43.95 & 43.06 \\
13 & PGD (Aug 0) $\rightarrow$ & & & \\
 & CuSP (PGD, 120-150-180,  $\{\frac{2}{255}, \frac{4}{255}, \frac{8}{255}\}$) & 17.88 & 50.10 & 49.15 \\
14 & Aug 180 $\rightarrow$ & & & \\
 & CuSP (PGD, 120-150-180,  $\{\frac{2}{255}, \frac{4}{255}, \frac{8}{255}\}$) & 15.80 & 49.85 & 49.17 \\
\hline
\hline
1 & Natural (Aug 0) & 00.06 & 74.93 & 20.27 \\
2 & Natural (Aug 180) & 00.01 & 58.96 & 59.49 \\
3 & TRADES (Aug 0) & 27.12 & 55.55 & 19.05 \\
4 & TRADES (Aug 180) & 18.98 & 44.75 & 44.82 \\
\hline
5 & TRADES (Aug 0) $\rightarrow$ Aug 180 & 00.01 & 64.84 & 64.13 \\
6 & Aug 180 $\rightarrow$ TRADES (Aug 0) & 27.01 & 55.10 & 19.83 \\
7 & TRADES (Aug 0) $\rightarrow$ TRADES (Aug 180) & 19.27 & 46.46 & 46.08 \\
8 & Aug 180 $\rightarrow$ TRADES (Aug 180) & 18.69 & 45.74 & 46.13 \\
\hline
9 & CuSP (TRADES, 30-60-180, $\{\frac{2}{255}, \frac{4}{255}, \frac{8}{255}\}$) & 20.00 & 50.43 & 42.17 \\
10 & CuSP (TRADES, 120-150-180,  $\{\frac{2}{255}, \frac{4}{255}, \frac{8}{255}\}$) & {18.00} & {48.83} & {47.34} \\
11 & CuSP (TRADES, 180, $\{\frac{2}{255}, \frac{4}{255}, \frac{8}{255}\}$) & 16.57 & 46.69 & 47.06 \\
12 & CuSP (TRADES, 120-150-180,$\{\frac{8}{255}\}$) & 20.09 & 46.61 & 45.63 \\
13 & TRADES (Aug 0) $\rightarrow$ & & & \\
 & CuSP (TRADES, 120-150-180,  $\{\frac{2}{255}, \frac{4}{255}, \frac{8}{255}\}$) & 18.57 & 48.90 & 47.60 \\
14 & Aug 180 $\rightarrow$ & & &  \\
 & CuSP (TRADES, 120-150-180,  $\{\frac{2}{255}, \frac{4}{255}, \frac{8}{255}\}$) & 17.64 & 48.97 & 47.40  \\
\hline
\end{tabular}
    \label{tradeoff-table-cifar100-pgd-trades-resnet18}
\end{table}

\begin{table}[!h]
\caption{\footnotesize Performance of proposed CuSP vs PGD (Aug 180) for CIFAR10 on StdCNN/VGG16 with various learning rate schedules (Aug $\theta$): denotes training data augmented with random rotations in the range $[-\theta, +\theta]$; $a \rightarrow b$: denotes $a$ sequentially followed by $b$ during training; $[E_{1}-E_{2}-...-E_{h}]$ : denotes the learning rate changes at epoch $E_{i}$.}
\footnotesize
    \centering
    \begin{tabular}{l | l | c  c  c }
    \hline
\bf Learning Rate Schedule &\bf Training Method&\bf Adv (PGD)&\bf Std &\bf Spatial\\
    &&\bf Accuracy(\%)&\bf Accuracy(\%)&\bf Accuracy(\%)\\
\hline
$[75-90-100]$ & PGD (Aug 180) & 32.79 & 54.60 & 53.69 \\
$[50-100]$ & PGD (Aug 180) & 34.17 & 57.48 & 56.97 \\
$[40-80]$ & PGD (Aug 180) & 34.00 & 57.40 & 56.40 \\
\hline
\hline
$[75-90-100]$ & CuSP (PGD, 180, $\{\frac{2}{255}, \frac{4}{255}, \frac{8}{255}\}$) & 33.63 & 62.17 & 61.07 \\
$[50-100]$  & CuSP (PGD, 180, $\{\frac{2}{255}, \frac{4}{255}, \frac{8}{255}\}$) & 34.86 & 65.09 & 63.87 \\
$[40-80]$ & CuSP (PGD, 180, $\{\frac{2}{255}, \frac{4}{255}, \frac{8}{255}\}$) & 34.06 & 61.97 & 61.21 \\
\hline
\end{tabular}
    \label{tradeoff-table-cifar10-vgg16-other}
\end{table}

\clearpage
\vspace{-4pt}
\section{Additional Experiments on Different Architectures}
\label{suppl:newarch}
\vspace{-6pt}
To study the trade-off further across a wider variety of architectures, we ran additional experiments using LeNet (2 Conv + 2 FullyConnected) and two MLP (4 layers) architectures and obtain the following results in Table \ref{tradeoff-table-cifar10-newarch} similar to Table \ref{tradeoff-table-cifar10-pgd-vgg16} row 3 (PGD (Aug 0)) and row 4 (PGD (Aug 180)) with CIFAR10. Note that the trade-off exists here too, supporting our observations across a gradation of architectures (MLP, LeNet, VGG16, ResNet18, WideResNet34, GCNN).

\begin{table}{!h}
\caption{\footnotesize Trade-off (PGD(Aug 0) vs PGD(Aug 180)) on simpler architectures like LeNet and MLP on CIFAR10 dataset.\vspace{-18pt}}\vspace{-18pt}
\footnotesize
    \centering
    \begin{tabular}{c | l | c  c  c }
    \hline
    \bf Network Architecture &\bf Training Method&\bf Adv (PGD)&\bf Std &\bf Spatial\\
    &&\bf Accuracy(\%)&\bf Accuracy(\%)&\bf Accuracy(\%)\\
\hline
LeNet (6-16-120-84-10)     & PGD(Aug 0)   & 27.38 & 40.52 & 23.36 \\ 
LeNet (6-16-120-84-10)     & PGD(Aug 180) & 20.80 & 26.94 & 27.18 \\ 
MLP (2048-2048-240-168-10) & PGD(Aug 0)   & 28.76 & 44.64 & 24.66 \\ 
MLP (2048-2048-240-168-10) & PGD(Aug 180) & 19.51 & 25.66 & 25.89 \\ 
MLP (4096-4096-480-336-10) & PGD(Aug 0)   & 28.89 & 45.34 & 25.31 \\ 
MLP (4096-4096-480-336-10) & PGD(Aug 180) & 19.87 & 26.77 & 26.75 \\ 
\hline
\end{tabular}
    \label{tradeoff-table-cifar10-newarch}
\end{table}

\end{document}